\documentclass{article}

\usepackage{arxiv}

\usepackage[utf8]{inputenc} 
\usepackage[T1]{fontenc}    
\usepackage{hyperref}       
\usepackage{url}            
\usepackage{booktabs}
\usepackage{comment}    
\usepackage{amsfonts}       
\usepackage{nicefrac}       
\usepackage{microtype}      
\usepackage{lipsum}		
\usepackage{graphicx}
\usepackage[numbers]{natbib}
\usepackage{doi}
\usepackage{adjustbox}
\usepackage{amsmath}
\usepackage[table,xcdraw]{xcolor}
\usepackage{multirow}
\usepackage{subcaption}

\title{Proto-LeakNet: Towards Signal-Leak Aware Attribution in Synthetic Human Face Imagery}

\author{
    \href{https://orcid.org/0009-0001-1284-6631}{\includegraphics[scale=0.06]{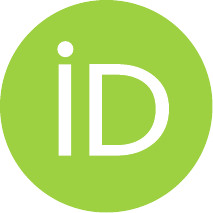}\hspace{1mm}Claudio Giusti} \\
  Department of Mathematics and Computer Science\\
  University of Catania\\
  Catania, CT 95125 \\
  \texttt{claudio.giusti@studium.unict.it} \\
   \And
  \href{https://orcid.org/0000-0001-8315-351X}{\includegraphics[scale=0.06]{orcid.pdf}\hspace{1mm}Luca Guarnera} \\
  Department of Mathematics and Computer Science\\
  University of Catania\\
  Catania, CT 95125 \\
  \texttt{luca.guarnera@unict.it} \\
  \And
  \href{https://orcid.org/0000-0001-6127-2470}{\includegraphics[scale=0.06]{orcid.pdf}\hspace{1mm}Sebastiano Battiato} \\
  Department of Mathematics and Computer Science\\
  University of Catania\\
  Catania, CT 95125 \\
  \texttt{sebastiano.battiato@unict.it} \\
}

\begin{document}
\maketitle

\begin{abstract}
The growing sophistication of synthetic image and deepfake generation models has turned source attribution and authenticity verification into a critical challenge for modern computer vision systems.
Recent studies suggest that diffusion pipelines unintentionally imprint persistent statistical traces, known as signal-leaks, within their outputs, particularly in latent representations.  
Building on this observation, we propose Proto-LeakNet, a signal-leak-aware and interpretable attribution framework that integrates Closed-set classification with a density-based Open-set
evaluation on the learned embeddings, enabling analysis of unseen
generators without retraining.  
Acting in the latent domain of diffusion models, our method re-simulates partial forward diffusion to expose residual generator-specific cues.  
A temporal attention encoder aggregates multi-step latent features, while a feature-weighted prototype head structures the embedding space and enables transparent attribution.  
Trained solely on closed data and achieving a Macro AUC of 98.13\%, Proto-LeakNet learns a latent geometry that remains robust under post-processing, surpassing state-of-the-art methods, and achieves strong separability both between real images and known generators, and between known and unseen ones. 
The codebase is available at the following link:~\url{https://github.com/claudiunderthehood/Proto-LeakNet}.
\end{abstract}

\maketitle

\section{Introduction}
\label{sec:intro}
The rapid progress of generative models has transformed digital content creation, enabling the synthesis of highly realistic images and videos, also called deepfakes, that are often indistinguishable from authentic ones~\citep{goodfellow2014generative,ho2020denoising,rombach2022high}. 
The increasing realism of such content has given rise to societal consequences beyond mere technical challenges: as synthetic media proliferates, individuals tend to systematically question the authenticity of even genuine content, a phenomenon recently formalized as the \textit{Impostor Bias}~\citep{CASU2024301795}. 
From early Generative Adversarial Networks~\citep{goodfellow2014generative} to modern diffusion-based architectures~\citep{ho2020denoising,rombach2022high}, the quality and accessibility of synthetic content have grown at an unprecedented pace, lowering the technical barrier for malicious use. In fact, deepfakes have already been exploited in real-world scenarios ranging from political disinformation campaigns and non-consensual intimate imagery to financial fraud and identity theft~\citep{verdoliva2020media,xie2025deepfake}. 
While these advances have fostered creativity and accessibility, they have also blurred the boundary between real and artificial content, posing serious challenges to media forensics, public trust, and legal accountability. As synthetic media proliferates across social, political, and creative domains, the ability to not only detect but also attribute its origin has become critical for security, digital evidence validation, and the assignment of responsibility~\citep{xie2025deepfake,verdoliva2020media}. 
Early research in multimedia forensics focused primarily on classifying whether an image is real or generated~\citep{guarnera2020deepfake,wang2021fakespotter}. 
However, a deeper forensic question lies in attribution, namely identifying which generative model produced a given image. 
Such a task, essential for tracing provenance and assessing responsibility, remains extremely challenging, especially in Open-set conditions where unknown generators appear at test time~\citep{khoo2022deepfake,bindini2024tiny}. 
Recent studies have shown that even advanced detectors struggle to generalize beyond the closed domain or to maintain interpretability when facing unseen architectures~\cite{corvi2023detection,wang2025openset}.
Diffusion models have recently reached state-of-the-art image quality, 
yet they introduce subtle statistical artifacts in their latent representations, known as \emph{signal-leaks}, caused by residual low-frequency information that survives the noising process~\cite{Everaert2024SignalLeak}. 
These traces, although imperceptible, encode model-specific biases and can serve as reliable forensic cues 
for deepfake source attribution. However, existing attribution methods typically operate in pixel or frequency domains, or extract static latent embeddings, without explicitly modeling the generative process dynamics. As a result, they lack robustness to domain shifts and offer limited interpretability, especially under Open-set conditions or when images undergo heavy post-processing. These limitations motivate the need for a method that directly exploits the temporal evolution of diffusion latents as stable, model-specific signatures for attribution.

To address these challenges, we propose Proto-LeakNet, which encodes image $x$ through Stable Diffusion latents and temporal attention, aggregates timestep embeddings via ResNet18, and performs attribution using prototype-based distances modulated by per-feature attention and gating. The pipeline is synthetically sketched in Fig.~\ref{fig:pipeline}. 
\begin{center}
  \includegraphics[width=1\linewidth]{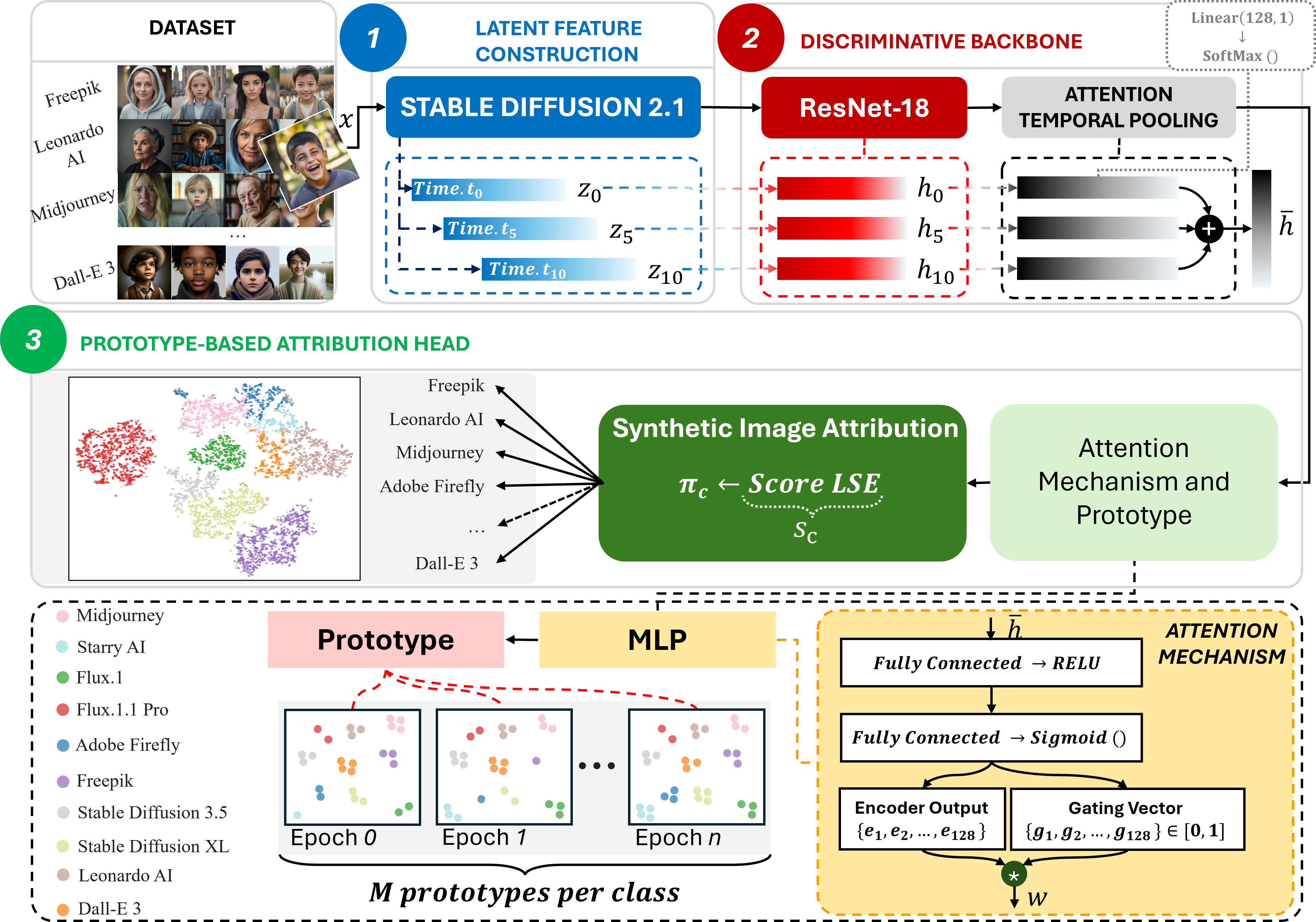}
  \captionof{figure}{Proto-LeakNet: given an input image $x$, latent features are extracted from the pretrained Stable Diffusion~2.1 Variational Autoencoder (VAE) in \textbf{Block~1 (Latent Feature Construction)}.
  For each diffusion step $t\!\in\!\{0,5,10\}$, we extract a latent $z_t$ which is normalized and has dimension $(4, 32, 32)$.
  In \textbf{Block~2 (Discriminative Backbone)}, each $z_t$ is encoded by a ResNet18, producing embeddings $\{h_t\}$ that are temporally aggregated through the \textbf{Attention Temporal Pooling} module to yield a single representation $\bar{h}$.
  \textbf{Block~3 (Prototype-Based Attribution)} computes distances between $\bar{h}$ and class prototypes $p_{c,m}$, modulated by a feature-wise gating vector $w$ obtained from a small MLP.
  The resulting attention-weighted distances are aggregated via a \textit{LogSumExp} scoring function to produce class probabilities $\pi_c$.
  Symbols: ``\textbf{+}'' denotes the weighted sum over attention coefficients across timesteps, and ``\textbf{*}'' indicates the element-wise product between the encoder output and the gating vector.}
  \label{fig:pipeline}
\end{center}

Our main contributions are the following:
\begin{itemize}
    \item We introduce \textbf{Proto-LeakNet}, a signal-leak-aware and interpretable attribution framework that operates entirely in the latent domain of diffusion models, learning generator-specific biases as stable forensic cues which demonstrates that modeling \textbf{signal-leak bias in latent space} leads to robust attribution

    \item We design a \textbf{temporal attention pooling mechanism} that aggregates latent representations across diffusion timesteps, enhancing discriminative power and interpretability by revealing which steps contribute most to attribution.
    
    \item We propose a \textbf{prototype-based attribution head} that shapes the latent geometry through learnable class prototypes and per-feature attention, enabling both compact cluster formation and feature-level interpretability.
    
    \item We develop a \textbf{density-based Open-set evaluation} using kernel density estimation on the learned embeddings to assess separability between real images and known generators, and between known and unseen ones without retraining.
    
    \item We demonstrate that modeling \textbf{signal-leak bias in latent space} leads to robust attribution under heavy post-processing and provides transparent prototype-level explanations.
\end{itemize}

This work is organized as follows. Section~\ref{sec:related} reviews the relevant state of the art. Section~\ref{sec:signal_leak_theory} presents signal-leak bias and provides a formal definition and theoretical justification. Section~\ref{sec:method} introduces our proposed framework and explores how the signal-leak bias is involved and exploited in the presented methodology. Section~\ref{sec:dataset} describes the datasets and metrics used in our study. Section~\ref{sec:experiments} reports the experimental results together with additional experiments and ablations supporting our framework. Section~\ref{sec:eval_part_dataset} further evaluates Proto-LeakNet on a challenging benchmark of partially manipulated real images. Section~\ref{sec:discussion} examines the implications and limitations of our approach. Finally, Section~\ref{sec:conclusions} summarizes the key findings and outlines future research directions.

\section{Related Work}
\label{sec:related}

The rapid evolution of generative models has fundamentally reshaped the landscape of digital media forensics. 
Early research primarily focused on binary deepfake detection, aiming to distinguish authentic images from manipulated or fully synthetic content. 
However, as generative models have become increasingly photorealistic and widely accessible, the forensic challenge has shifted from mere detection toward identifying the specific source model responsible for generating an image. 
This transition from Closed-set detection to source attribution and open-world recognition introduces new requirements, including robustness to unseen generators, generalization across datasets, and interpretability of learned representations.
In parallel, the emergence of diffusion-based generative models has altered the forensic signal landscape. 
Unlike GAN-based pipelines, diffusion models generate content through iterative denoising processes operating in structured latent spaces, introducing distinct temporal and frequency-domain characteristics. 
Recent works have begun to explore latent-space reasoning and model-specific artifacts in diffusion pipelines, yet a comprehensive understanding of how generator-specific traces evolve across denoising steps remains limited.
In the following, we review prior work in deepfake detection, source attribution, open-world recognition, frequency-based forensics, and diffusion-latent analysis, highlighting the methodological gaps that motivate our proposed approach.

\subsection{Deepfake Detection}
Early research in synthetic media analysis relied on handcrafted features targeting compression inconsistencies, noise patterns, and color statistics~\citep{verdoliva2020media}. With the advent of deep learning, CNN-based detectors became dominant, learning discriminative features directly from data. Wang et al.~\citep{wang2020cnn} demonstrated that a classifier trained on ProGAN images generalizes surprisingly well to other generators, suggesting the existence of common forensic traces across synthesis pipelines, while Chai et al.~\citep{chai2020makes} further analyzed which image properties enable cross-generator detection. 

More recent approaches have extensively focused on improving the generalization of detectors toward unseen generators. $D^3$~\citep{yang2025d} proposed scaling up the training set by including multiple generators to better capture such traces. Other methods leveraged the generalization capabilities of vision-language models such as CLIP, through prompt learning~\citep{tan2025c2p} or fine-tuning~\citep{CLIP2}. UFD~\citep{UFD} was the first to show that a feature space not explicitly learned for generated image detection improves robustness due to its unbiased decision boundaries. Wang et al.~\citep{wang2026penny} identified a "Penny-Wise and Pound-Foolish" phenomenon in CLIP-based detectors, showing that fine-tuning with a binary objective degrades upstream knowledge and impairs generalization; their PoundNet framework addresses this through a balanced objective that jointly preserves semantic knowledge and improves cross-generator detection. Willi et al.~\citep{willi2026synthetic} further investigated what CLIP-based detectors actually learn, showing that they rely primarily on high-level photographic attributes rather than generator-specific fingerprints, and that generalization across generator families drops sharply, as low as 0.37 mAP, highlighting the need for broader training exposure and continual model updates.

The emergence of diffusion models introduced additional challenges: Wang et al.~\citep{wang2023dire} proposed using reconstruction error from pre-trained diffusion models as a discriminative signal, while Corvi et al.~\citep{corvi2023detection} showed that detectors trained on GAN-generated images often fail on diffusion-based content. This challenge was explicitly addressed in~\cite{guarnera2024mastering}, where a unified framework was proposed to distinguish both GAN- and diffusion-generated imagery, highlighting the growing complexity of the detection landscape.  To address both GAN and diffusion detection jointly, Pontorno et al.~\citep{pontorno2025deepfeaturex} proposed DeepFeatureX-SN, a tripartite architecture of specialized feature extractors trained via contrastive learning, each dedicated to one image class (real, GAN-generated, or diffusion-generated), achieving strong generalization to unseen architectures.

Several methods exploit frequency-domain artifacts left by generative pipelines. 
Tan et al.~\citep{freqnet2023} introduced FreqNet, a detector operating partially in the Fourier domain, enforcing the learning of high-frequency features via an explicit FFT branch. 
Similarly, Tan et al.~\citep{npr2024} proposed NPR, targeting upsampling artifacts through pixel-wise residual analysis in local patches.
Patch-level forensic approaches also focus on localized inconsistencies. 
Bernabeu-Pérez et al.~\citep{susy2024} introduced SuSy, a CNN-based patch classifier that extracts high-contrast patches via GLCM texture analysis to detect and attribute AI-generated imagery.
These methods demonstrate that spectral and local residual cues remain informative; however, they often operate in static feature spaces and do not explicitly model the generative process dynamics.

However, binary detection does not provide information about the source model and often fails under distribution shifts, especially when confronted with unseen generators. This motivates the shift toward source attribution.

\subsection{Deepfake Source Attribution}

Beyond binary detection of synthetic media, recent research has increasingly focused on \emph{source attribution}, where the goal is to identify the specific generative model or manipulation pipeline responsible for producing a synthetic image. 
Early work highlighted that different deepfake generation techniques introduce distinct artifacts and manipulation traces, motivating attribution as a more informative forensic task than simple real–fake classification. In this direction, Guarnera et al. \cite{guarnera2022exploitation} proposed one of the first pipelines targeting model-level attribution, going beyond architecture recognition to identify which specific instance of a StyleGAN2-ADA \cite{karras2020training} model generated a given image. By exploiting the latent space of 50 slightly different models and training a ResNet18-based encoder, they achieved over 96\% classification accuracy and introduced a generalizable metric for unseen models, drawing an explicit analogy with camera source identification in traditional multimedia forensics.

Khoo et al.~\citep{khoo2022deepfake} provided one of the first systematic discussions of deepfake attribution, categorizing manipulation pipelines into identity swap, attribute manipulation, expression manipulation, and full-face synthesis. 
Their analysis emphasized that attribution can provide stronger interpretability and forensic value than binary detection. 
Similarly, Jain et al.~\citep{jain2021improving} demonstrated that training models directly for attribution improves cross-dataset generalization compared to binary detection models. 
Using XceptionNet and EfficientNet backbones across multiple deepfake datasets, they showed that multi-class attribution objectives, especially when combined with triplet loss, produce embeddings that transfer more effectively to unseen manipulations.

Another line of research investigates how attribution models behave under dataset shifts. 
Baxevanakis et al.~\citep{baxevanakis2025deepfake} conducted a large-scale comparison of binary and multi-class models across several deepfake datasets, revealing that attribution models often struggle with cross-dataset generalization due to distribution shifts between manipulation pipelines. 
Their study also showed that contrastive learning objectives can partially mitigate these issues by encouraging manipulation-aware embeddings.

\noindent
Closed-set attribution models typically assume that all generator classes are known during training, which limits their applicability in real-world scenarios where new generators appear frequently. To address this limitation, Sun et al.~\citep{sun2023contrastive} introduced the Open-World DeepFake Attribution (OW-DFA) benchmark and proposed Contrastive Pseudo Learning, which combines global–local feature voting with confidence-based pseudo-labeling to improve attribution robustness under Open-set conditions. Few-shot and one-class approaches further aim to recognize unseen generators with limited supervision. For instance, Liu et al.~\citep{occlip2024} proposed OCC-CLIP, which leverages CLIP embeddings to perform one-class attribution and detect unseen generators with minimal labeled examples. \color{black} To extend these capabilities to unseen models, OWDFA-CAL~\citep{zheng2026open} shifts the focus toward open-world scenarios. Instead of relying solely on Closed-set artifacts, it employs a confidence-aware asymmetric learning framework that dynamically balances model confidence, allowing it to discover and attribute novel forgery types without knowing the number of unknown generators a priori. Expanding on this need to handle unseen generators, Fang et al.~\citep{fang2023open} propose a metric learning-based approach for open-set attribution that utilizes transferable embeddings, employing a normalized distance-based rejection criterion to reliably identify whether an image originates from a known or unknown source. Advancing this concept into a few-shot paradigm, OmniDFA~\citep{wu2025omnidfa} utilizes a dual-path architecture and supervised contrastive learning to explicitly isolate model-specific biases, enabling the attribution of novel models using only minimal support samples. Taking this evolution further into semi-supervised open-world environments, the DATA framework~\citep{liu2025data} introduces multi-disentanglement contrastive learning. To suppress forgery-irrelevant dataset biases, DATA projects features onto learned "Orthonormal Deepfake Bases" to disentangle method-specific clues, while an augmented-memory clustering mechanism dynamically discovers and categorizes novel generator prototypes. While these approaches demonstrate that spectral and local artifacts contain useful forensic information, they generally operate in static feature spaces and do not explicitly model the generative process that produced the image.

\color{black}
\noindent
\textbf{Artifact- and Frequency-Based Detection and Attribution.} 
Many attribution approaches rely on artifacts introduced by generative pipelines. 
Frequency-domain analysis has proven particularly effective in revealing generator-specific signatures. 
FreqNet~\citep{freqnet2023} introduces a detection architecture that explicitly processes images in the Fourier domain, enforcing the learning of high-frequency cues through a dedicated FFT branch. 
Similarly, NPR~\citep{npr2024} targets upsampling artifacts produced by convolutional generators, extracting pixel-wise residual correlations across image patches. 
Patch-based forensic approaches also exploit localized inconsistencies. 
SuSy~\citep{susy2024}, for example, identifies high-contrast image regions using GLCM texture statistics and trains a CNN-based classifier on these patches to perform generator attribution. 
While these approaches demonstrate that spectral and local artifacts contain useful forensic information, they generally operate in static feature spaces and do not explicitly model the generative process that produced the image.

\color{black}
\noindent
\textbf{Latent-Space and Diffusion-Based Forensics.} 
More recent work has shifted toward analyzing generative models directly in their latent representations. 
LatentTracer~\citep{latenttracer2024} adopts a reverse-engineering strategy, projecting images into candidate generative models and comparing reconstruction errors to identify the most likely source generator. 
Similarly, LATTE~\citep{vasilcoiu2025latte} operates directly in diffusion latent space using a transformer architecture to model long-range dependencies among latent tokens. 
These methods highlight that diffusion models encode distinctive signatures in their latent trajectories. 
However, most existing approaches rely on static embeddings extracted at a single timestep and do not explicitly analyze how generator-specific traces evolve temporally during the diffusion process.

\color{black}
\subsection{Positioning of Our Work}
Existing attribution methods either operate in pixel/frequency domains or extract static latent embeddings, often lacking temporal modeling of diffusion dynamics and providing limited interpretability.
A particularly relevant line of research identifies structural biases in diffusion training dynamics that motivate a different approach. Everaert et al.~\citep{Everaert2024SignalLeak} showed that diffusion models exhibit a \emph{signal-leak bias}, whereby low-frequency mismatches between data and noise distributions encode generator-specific traces. Their analysis demonstrates that signal leakage persists at high timesteps and influences style and brightness characteristics. While originally studied for controllability and style adaptation, this phenomenon suggests that diffusion latents inherently retain structured residual information tied to training distributions.
Building on these insights, we introduce \textbf{Proto-LeakNet}, a framework that explicitly aggregates diffusion latent residuals across timesteps through prototype supervision and temporal attention. By modeling the temporal evolution of generator-specific cues, our approach enables robust and interpretable attribution under both closed- and Open-set conditions.

Looking at the surveyed methods as a whole, some clear limitations emerge. Frequency-domain approaches such as FreqNet and NPR work well on clean images, but they rely on high-frequency artifacts that are easily destroyed by compression or smoothing, making them fragile under realistic post-processing conditions. Pixel-domain and patch-based methods suffer from the same problem: the features they extract are tied to surface-level textures that do not survive aggressive image transformations. Latent-space methods such as LatentTracer and LATTE are more robust, since they operate on representations that are less sensitive to pixel-level noise. However, most of them extract a single static embedding at one diffusion timestep, ignoring the temporal structure of the generative process and losing potentially useful information. Prototype-based and contrastive methods such as OCC-CLIP and OW-DFA tackle the Open-set problem, but they do not model the physical mechanism that produces generator-specific traces in the first place. Proto-LeakNet is designed to address all these limitations at once: by combining latent residuals from multiple diffusion timesteps, it captures generator-specific cues that are robust to post-processing, while the prototype-based supervision and density estimation provide a principled solution for both Closed-set attribution and Open-set rejection.

\section{Theoretical Analysis of Signal Leakage}
\label{sec:signal_leak_theory}
To clarify what signal-leak represents in our setting, we introduce a theoretical analysis independent of the main pipeline. This allows us to formalize the phenomenon and rigorously justify its existence. \\ Diffusion models are trained to approximate the reverse of a fixed forward noising process.  
Let $x_0 \!\sim\! q(x_0)$ denote a clean training image drawn from the empirical data distribution,
and let $p_{\text{noise}} = \mathcal{N}(0,I)$ (where $I$ is the identity matrix) denote the standard isotropic Gaussian noise prior in $\mathbb{R}^d$,
where $d$ is the latent dimensionality.
At each discrete timestep $t\!\in\![1,T]$, the forward noising process adds Gaussian noise according to
\begin{align}
    q(x_t|x_{t-1}) 
    &= \mathcal{N}\!\left(\sqrt{\alpha_t}\,x_{t-1},\, (1-\alpha_t)I\right),
    \label{eq:markov}
\end{align}
where $\alpha_t\!\in\!(0,1)$ controls the noise level at step $t$.
Equation~\eqref{eq:markov} defines a first–order Markov chain that progressively corrupts the image.

Let $\bar{\alpha}_t = \prod_{s=1}^t \alpha_s$ denote the cumulative product of noise coefficients.
By iterating~\eqref{eq:markov}, one obtains a closed–form expression for $x_t$ as a function of $x_0$:
\begin{align}
    x_t 
    &= \sqrt{\bar{\alpha}_t}\,x_0 
     + \sqrt{1-\bar{\alpha}_t}\,\varepsilon, 
     \quad \varepsilon \!\sim\! \mathcal{N}(0,I).
     \label{eq:xt}
\end{align}
Here, $\varepsilon$ represents the injected Gaussian noise independent of $x_0$.
Equation~\eqref{eq:xt} indicates that $x_t$ is a convex combination of the original signal $x_0$
and white noise $\varepsilon$, weighted respectively by $\sqrt{\bar{\alpha}_t}$ and $\sqrt{1-\bar{\alpha}_t}$.
The denoising network $\varepsilon_\theta(x_t,t)$ is trained to recover the
Gaussian noise that was injected during the forward process.  
During training, the expectation $\mathbb{E}_{x_0,\varepsilon_0,t}$ denotes a joint average over: 
\begin{itemize}
    \item data samples $x_0 \sim q(x_0)$,
    \item independent Gaussian noise samples $\varepsilon_0 \sim \mathcal{N}(0,I)$,
    \item noise levels $t$ sampled uniformly from $\{1,\ldots,T\}$.
\end{itemize}
Under this sampling procedure, the objective becomes
\begin{align}
    \mathcal{L}_\theta
    &= \mathbb{E}_{x_0,\varepsilon_0,t}
       \Big[\|\varepsilon_0 - \varepsilon_\theta(x_t,t)\|^2\Big].
       \label{eq:loss}
\end{align}
This joint expectation ensures that the model learns to denoise across both
the full data distribution and the entire range of noise levels encountered
during training.
Here, $t$ is uniformly sampled from $\{1,\ldots,T\}$ at each iteration,
ensuring that the model learns to denoise across a range of noise levels
and effectively approximates the reverse diffusion process.

\subsection{Residual signal at the terminal step}  
The marginal distribution of $x_T$, the last noised state seen during training is
\begin{align}
    q(x_T)
    &= \mathcal{N}\!\big(
       \sqrt{\bar{\alpha}_T}\,\mathbb{E}[x_0],\;
       (1-\bar{\alpha}_T)I 
       + \bar{\alpha}_T \operatorname{Cov}[x_0]
       \big),
       \label{eq:qxt}
\end{align}
where $\mathbb{E}[x_0]$ and $\operatorname{Cov}[x_0]$ are the empirical mean and covariance of the data distribution.
Unless $\bar{\alpha}_T\!=\!0$, this Gaussian retains a nonzero mean component 
proportional to $\sqrt{\bar{\alpha}_T}\,\mathbb{E}[x_0]$,  
implying that $x_T$ still carries residual information about the original data.  
We rewrite~\eqref{eq:xt} at $t\!=\!T$ to isolate this effect:
\begin{align}
    x_T 
    &= \underbrace{\sqrt{\bar{\alpha}_T}\,x_0}_{\text{signal-leak}}
     + \underbrace{\sqrt{1-\bar{\alpha}_T}\,\varepsilon}_{\text{noise term}}.
     \label{eq:leak}
\end{align}
The first term, $\sqrt{\bar{\alpha}_T}x_0$, is referred to as the \emph{signal-leak}, %
a residual projection of $x_0$ that survives the forward diffusion process.

At inference time, however, the standard procedure initializes from pure Gaussian noise:
\begin{align}
    \hat{x}_T &\sim p_{\text{noise}} = \mathcal{N}(0,I),
    \label{eq:inference_latent}
\end{align}
implicitly assuming that $q(x_T)\!=\!p_{\text{noise}}$.
This assumption introduces a \emph{distributional mismatch} between the statistics of 
training inputs $x_T$ and inference inputs $\hat{x}_T$,
measurable by the Kullback–Leibler divergence ($\mathcal{D}_{\text{KL}}$):
\begin{align}
    \mathcal{D}_{\text{KL}}\!\left(q(x_T)\,\|\,p_{\text{noise}}\right) > 0.
    \label{eq:kl_gap}
\end{align}
The nonzero divergence in~\eqref{eq:kl_gap} defines the signal-leak bias.
Because the model’s learned reverse process depends on the distribution of $x_T$,
starting inference from the incorrect prior $p_{\text{noise}}$ perturbs 
the score field $\nabla_x \log q_\theta(x)$ and alters the reverse-diffusion trajectory.
Here, $\nabla_x \log q_\theta(x)$ denotes the model’s estimate of the 
score function, the gradient of the log-density with respect to the input.
This vector field indicates the direction in which the model believes the data
distribution becomes more likely, and it governs each step of the learned
reverse diffusion dynamics. To support this claim Figure~\ref{fig:kl_vs_alpha} helps us visualize that although diffusion training assumes that the final noised samples $x_T$ follow $p_{\text{noise}} = \mathcal{N}(0,I)$, in practice there remains a residual signal component whose strength grows with $\bar{\alpha}_T$. 

\begin{center}
  \includegraphics[width=.7\linewidth]{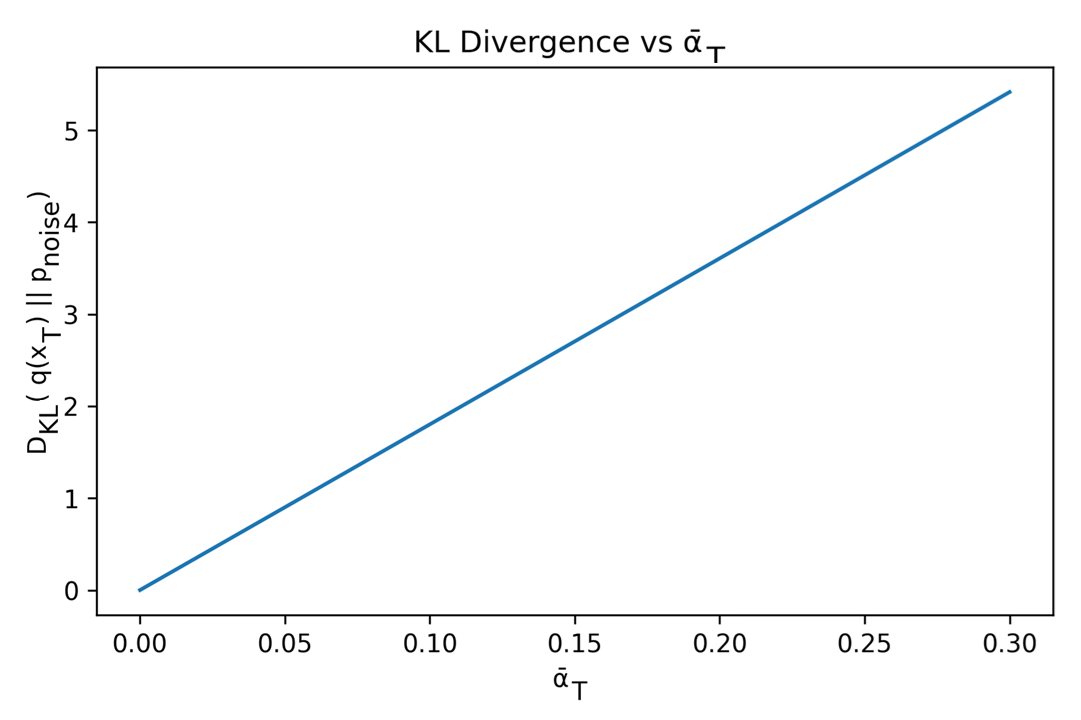}
  \captionof{figure}{Plot that supports that the residual signal magnitude at the terminal step grows proportionally to the retained signal coefficient $\bar{\alpha}_T$.}
  \label{fig:kl_vs_alpha}
\end{center}

\subsection{Spectral characterization}  
Let $X^{(0)}_{u,v}$ denote the $(u,v)$–th coefficient of the two–dimensional Discrete Cosine Transform (DCT)
of the clean image $x_0$, and let $E_{u,v}$ denote the corresponding DCT coefficient of the Gaussian noise $\varepsilon$.
We adopt the DCT representation because it provides a compact, orthogonal basis aligned with the spatial frequency content of natural images.
Unlike raw pixel space, the DCT separates the signal into interpretable frequency bands,
allowing a direct analysis of how diffusion noise interacts with each component of the spectrum.
This frequency–domain formulation makes the residual signal energy quantitatively measurable
and facilitates a closed–form expression for the per–frequency signal–to–noise ratio (SNR).
Natural images follow a well–known power–spectrum law $P(f)\!\propto\!1/f^2$~\citep{field1987relations},
indicating that most of their energy is concentrated in the lowest spatial frequencies,
which are also the most affected by residual leakage in diffusion models.
The signal–to–noise ratio (SNR) of the residual component at each frequency can be derived from~\eqref{eq:leak} as
\begin{align}
    \mathrm{SNR}(u,v)
    &= 
    \frac{\bar{\alpha}_T
          \,\mathbb{E}\!\left[(X^{(0)}_{u,v})^2\right]}
         {(1-\bar{\alpha}_T)\,\mathbb{E}\!\left[(E_{u,v})^2\right]}
     = 
     \frac{\bar{\alpha}_T}{1-\bar{\alpha}_T}
     \,\mathbb{E}\!\left[(X^{(0)}_{u,v})^2\right].
     \label{eq:snr}
\end{align}
Since $E_{u,v}\!\sim\!\mathcal{N}(0,1)$, its variance equals $1$,  
and thus the ratio in~\eqref{eq:snr} directly reflects the energy of each image frequency.
For low–frequency indices $(u,v)\!\approx\!0$, 
$\mathbb{E}[(X^{(0)}_{u,v})^2]$ is large, leading to $\mathrm{SNR}(u,v)\!\gg\!0$.
Consequently, low–frequency structures (brightness and color) remain partially intact even at $t=T$, 
while high–frequency details are almost completely dominated by noise.
This asymmetry implies that the terminal state $x_T$ retains a biased footprint of the original image,
consisting primarily of global luminance and coarse spatial structure.
At inference, however, the initialization $\hat{x}_T \sim \mathcal{N}(0,I)$ contains no such low-frequency bias,
so the two distributions differ most prominently in their low-frequency components.
This mismatch manifests as a systematic shift toward \emph{medium brightness} or desaturated tones
when sampling from the reverse diffusion process.

\subsection{Distributional mismatch formulation}  
The distribution $q(x_T)$ in~\eqref{eq:qxt} can be expressed as a convolution
between the data distribution and a Gaussian smoothing kernel.  
Formally, let $G_T$ denote the Gaussian density
\begin{equation}
    G_T(z) = \mathcal{N}\!\left(z; 0,\,(1-\bar{\alpha}_T)I \right),
\end{equation}
where $z$ is an integration variable and $(1-\bar{\alpha}_T)I$ is the covariance
of the noise accumulated up to timestep $T$.
The convolution of $G_T$ with the data density $q(x_0)$ is defined as
\begin{equation}
    (G_T * q)(x_T)
    = \int_{\mathbb{R}^d} G_T(x_T - x_0)\, q(x_0)\, dx_0.
\end{equation}
Under the forward diffusion process, the terminal distribution therefore satisfies
\begin{equation}
    q(x_T) = (G_T * q)(x_T),
    \label{eq:conv}
\end{equation}
showing that $q(x_T)$ is obtained by smoothing the original data distribution
with a Gaussian kernel of variance $(1-\bar{\alpha}_T)$.
This convolutional form makes explicit that $q(x_T)$ retains a residual imprint
of $q(x_0)$ whenever $\bar{\alpha}_T > 0$, and hence cannot match the isotropic
noise prior $p_{\text{noise}}=\mathcal{N}(0,I)$ used at inference.
When $q(x_0)\!\neq\!\mathcal{N}(0,I)$, this convolution produces a shifted and anisotropic density,
so $q(x_T)$ differs from the isotropic prior $p_{\text{noise}}$.  
The discrepancy can also be characterized by a nonzero cross–covariance term:
\begin{align}
    \Sigma_{x_0,\varepsilon}(T)
    &= \sqrt{\bar{\alpha}_T(1-\bar{\alpha}_T)}
       \operatorname{Cov}(x_0,\varepsilon)
       \neq 0.
       \label{eq:crosscov}
\end{align}
During training, the model implicitly observes nonzero $\Sigma_{x_0,\varepsilon}(T)$,
whereas at inference, $\operatorname{Cov}(x_0,\varepsilon)\!=\!0$ by construction. This term is illustrated in Figure~\ref{fig:crosscov_heatmap} where the bright diagonal blocks show strong intra-correlation between corresponding dimensions of $x_t$ and $\varepsilon$, supporting the existence of a measurable residual dependence, the "leak channel".
This covariate shift leads to accumulated deviations across denoising steps. To better visualize the mismatch, Figure~\ref{fig:mismatch} clearly shows us that the two distributions, although overlapping exhibit a clear offset. That occurs due to the training-time forward process distribution ($q(x_T)$) not being identical to the inference-time prior $p_{\text{noise}}$. The consequent small mean shift and slightly different variance along the principal axes are the signal-leak bias itself.

\begin{center}
  \includegraphics[width=.7\linewidth]{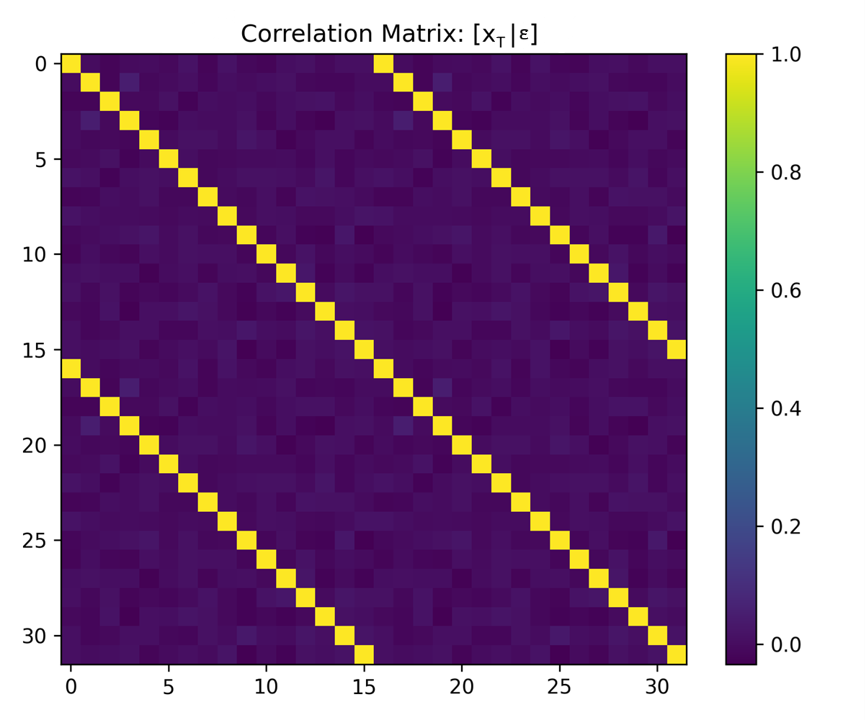}
  \captionof{figure}{Heatmap of the cross-covariance term measuring intra-correlation between $x_t$ and $\varepsilon$.}
  \label{fig:crosscov_heatmap}
\end{center}

\begin{center}
  \includegraphics[width=.7\linewidth]{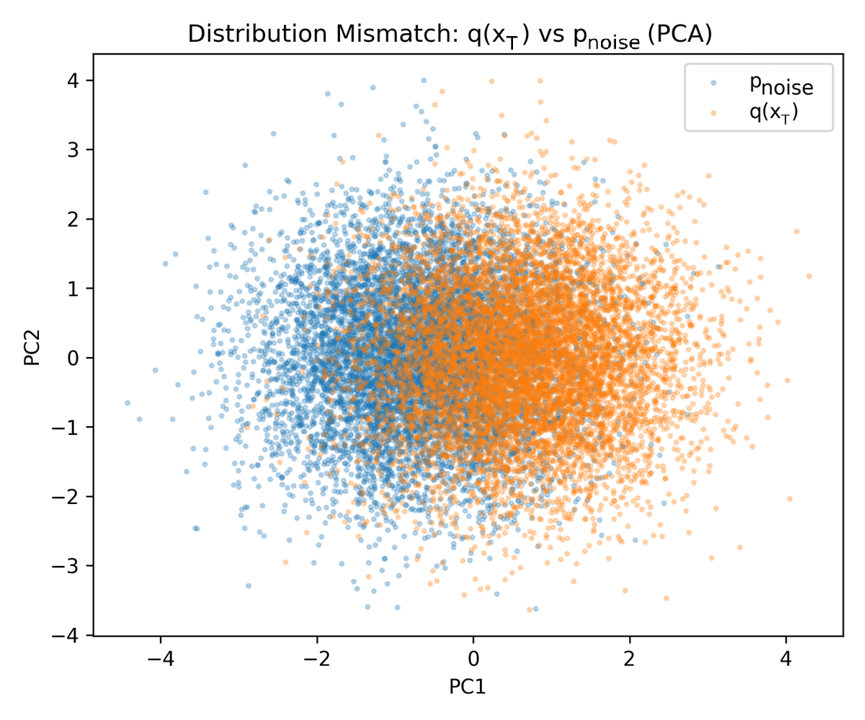}
  \captionof{figure}{PCA plot which illustrates the distribution mismatch between $q(x_T)$ and $p_{\text{noise}}$.}
  \label{fig:mismatch}
\end{center}

\subsection{Energy analysis and limit case.}  
The expected squared norm of $x_T$ follows from~\eqref{eq:xt}:
\begin{align}
    \mathbb{E}\!\left[\|x_T\|^2\right]
    &= 
    \bar{\alpha}_T\,\mathbb{E}\!\left[\|x_0\|^2\right]
     + d(1-\bar{\alpha}_T),
     \label{eq:energy}
\end{align}
where $d$ is the latent dimensionality.
If $\bar{\alpha}_T\!\to\!0$, 
then $\mathbb{E}[\|x_T\|^2]\!\to\!d$, matching the energy of pure noise and eliminating the leak.
However, for typical schedules (e.g., the $\sqrt{\beta}$ schedule in Stable Diffusion~\citep{rombach2022high}),
$\sqrt{\bar{\alpha}_T}\!\approx\!0.07$, 
so the first term in~\eqref{eq:energy} remains non-negligible,
implying that training never observes purely noisy inputs.

\subsection{Expected observable bias}  
Let $\phi(x)$ denote any measurable low–frequency observable,
for example, the mean pixel intensity.
The expected bias between training and inference distributions is
\begin{align}
    \Delta_{\text{bias}}
    &= 
    \mathbb{E}_{x_T\sim q(x_T)}[\phi(x_T)]
     - 
     \mathbb{E}_{x_T\sim p_{\text{noise}}}[\phi(x_T)].
     \label{eq:bias}
\end{align}
Empirically, $\Delta_{\text{bias}}>0$ for luminance observables,
manifesting as consistently medium–gray outputs in generated images.

\subsubsection*{Realigning Training and Inference Distributions}

To correct the mismatch, the inference process can be modified to include an explicit signal–leak component.
Let $\tilde{q}_0$ denote an empirical approximation of $q(x_0)$,
for example, a Gaussian distribution $\mathcal{N}(\mu,\operatorname{diag}(\sigma^2))$
estimated from a small set of reference or target images.
Drawing $\tilde{x}\!\sim\!\tilde{q}_0$ and $\varepsilon\!\sim\!\mathcal{N}(0,I)$,
we redefine the initialization at the last timestep as
\begin{align}
    \hat{x}_T 
    &= \sqrt{\bar{\alpha}_T}\,\tilde{x}
     + \sqrt{1-\bar{\alpha}_T}\,\varepsilon.
     \label{eq:correction}
\end{align}
All variables in~\eqref{eq:correction} mirror their training counterparts in~\eqref{eq:xt},
ensuring that $\hat{x}_T$ follows the same distributional structure as $x_T$.

Let $\hat{q}(x_T)$ denote the distribution induced by~\eqref{eq:correction}.
The Kullback–Leibler divergence between the training and inference distributions is then
\begin{equation}
    \begin{split}
    D_{\text{KL}}\!\big(q(x_T)\,\|\,\hat{q}(x_T)\big)
    &= \tfrac{1}{2}\Big(
        \|\mu_q - \mu_{\hat{q}}\|_{\Sigma^{-1}}^2
    \\ &\quad
        + \mathrm{Tr}\!\left(\Sigma_{\hat{q}}^{-1}\Sigma_q - I\right)
        - \log\tfrac{|\Sigma_q|}{|\Sigma_{\hat{q}}|}
      \Big)
    \end{split}
    \label{eq:dkl_form}
\end{equation}

where $(\mu_q,\Sigma_q)$ and $(\mu_{\hat{q}},\Sigma_{\hat{q}})$ denote the respective
means and covariances, and $\mathrm{Tr}(\cdot)$ denotes the matrix trace operator,
i.e., the sum of the diagonal entries of its argument. 
Substituting the expressions from~\eqref{eq:qxt} and~\eqref{eq:correction} gives
$\mu_q\!=\!\mu_{\hat{q}}\!=\!\sqrt{\bar{\alpha}_T}\mathbb{E}[x_0]$
and $\Sigma_q\!=\!\Sigma_{\hat{q}}\!=\!(1-\bar{\alpha}_T)I+\bar{\alpha}_T\operatorname{Cov}[x_0]$,
yielding $D_{\text{KL}}=0$.
Hence, the initialization in~\eqref{eq:correction} exactly realigns the inference distribution 
with the one observed during training, 
eliminating the bias introduced by~\eqref{eq:inference_latent}. Figure~\ref{fig:expected_bias} reports the distribution of mean image intensities: the real images have a narrow distribution of mean luminance while the standard diffusion generated images start from pure noise, carrying a higher expected mean while the corrected initialization recovers a mean closer to the real distribution.

\begin{center}
  \includegraphics[width=.7\linewidth]{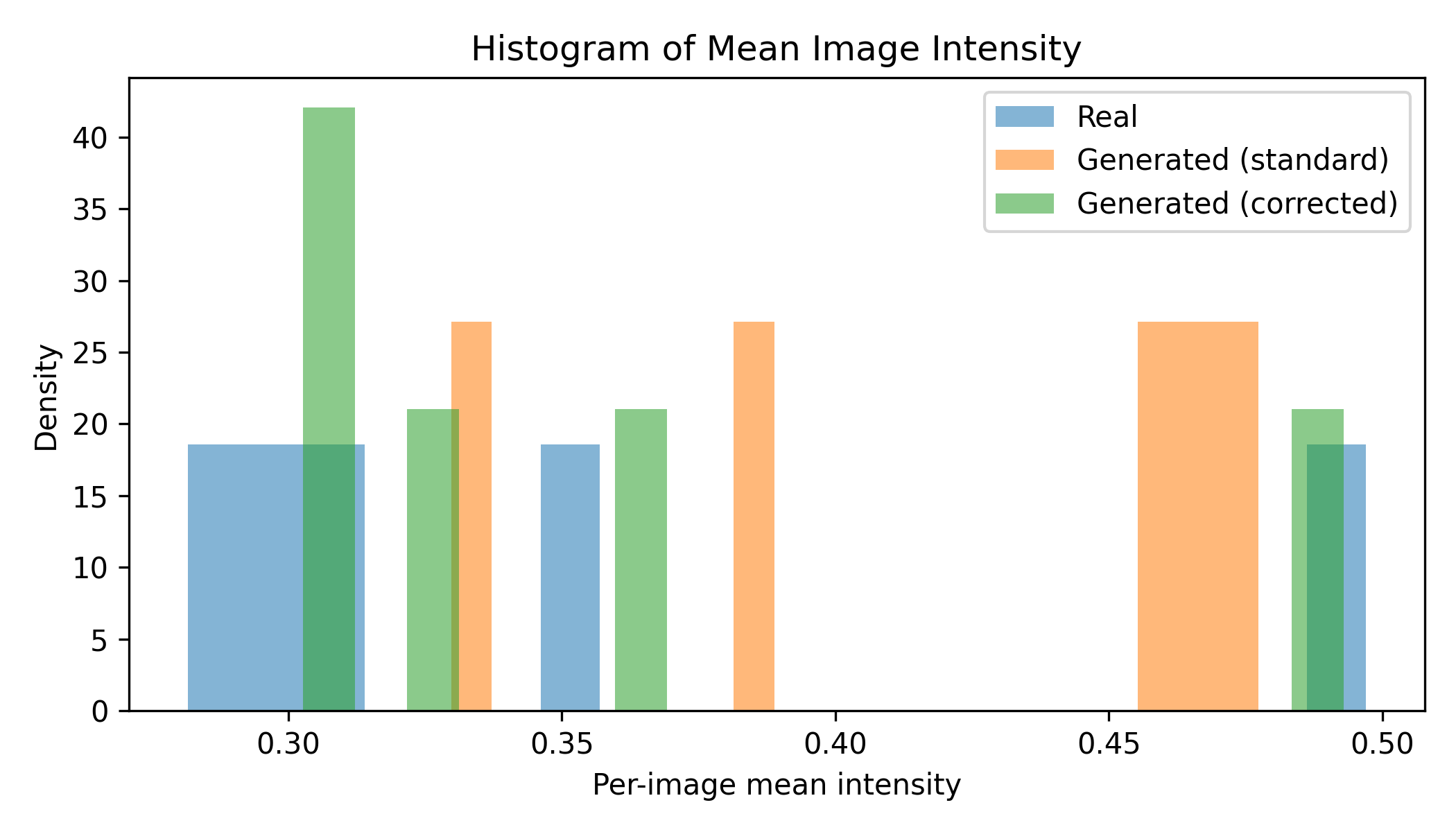}
  \captionof{figure}{\color{black}Histogram of per-image mean intensity for real images, standard generated images, and generated images produced with the corrected initialization. The three visible groups correspond to different mean-intensity bins of the evaluated samples. The plot shows that standard generation shifts the mean-intensity distribution away from real images, while the corrected initialization better aligns the generated images with the real-image intensity statistics.}
  \label{fig:expected_bias}
\end{center}

\subsection{Interpretation}  
Equation~\eqref{eq:correction} can be viewed as injecting a low–frequency prior 
into the starting latent.
The corrective term $\sqrt{\bar{\alpha}_T}\tilde{x}$ replaces the missing residual statistics
that the model expects from training, while $\varepsilon$ preserves high–frequency stochasticity.
In the limit $\tilde{q}_0=q(x_0)$, 
the inference and training distributions coincide exactly, 
and the model’s generative dynamics become statistically consistent.

\section{Proposed Method}
\label{sec:method}
The proposed Proto-LeakNet framework is organized into three functional blocks. \textit{Latent Feature Construction} extracts informative representations from the diffusion latents of Stable Diffusion 2.1~\citep{rombach2022high}. \textit{Discriminative Backbone} aggregates these multi-scale features through a temporal attention encoder. The \textit{Prototype-Based Attribution Head} interprets the learned representation using class-specific prototypes for interpretable generator attribution. Finally, once defined the pipeline we introduce the generalization methodology.

\color{black}
\subsection{Problem Definition}

This work addresses the problem of synthetic image source attribution under two distinct evaluation regimes: Closed-set attribution and Open-set rejection. Since these two settings are often conflated in the literature, we explicitly define them before introducing the proposed method.

\subsubsection{Closed-set Source Attribution}

Let $\mathcal{G}_{\mathrm{known}}=\{g_1,\ldots,g_C\}$ denote the set of generative models observed during training, where $C$ is the number of known source generators. In this scenario, given an input image $x$ produced by one generator $g_c \in \mathcal{G}_{\mathrm{known}}$, the Closed-set source attribution task consists of assigning $x$ to one of the known generators:
\begin{equation}
    f_{\mathrm{closed}}(x) \in \mathcal{G}_{\mathrm{known}}.
\end{equation}
where $f_{\mathrm{closed}}(\cdot)$ denotes the Closed-set attribution function, which maps an input image to one of the known generator classes.

In this setting, every test sample is assumed to originate from one of the generators available during training. Therefore, the model is evaluated as a multi-class classifier over the known generator set. This is the primary attribution task addressed by Proto-LeakNet. The prototype-based attribution head is trained to structure the latent representation according to generator-specific signal-leak patterns and to produce class scores for the known generators.

\subsubsection{Open-set Rejection}

In realistic forensic scenarios, a test image may be produced by a generator that was not available during training. Let $\mathcal{G}_{\mathrm{unknown}}$ denote the set of unseen generators, with
\begin{equation}
    \mathcal{G}_{\mathrm{known}} \cap \mathcal{G}_{\mathrm{unknown}} = \emptyset.
\end{equation}

In this work, we do not formulate this setting as Open-set attribution in the strict sense, because the goal is not to identify the specific unseen generator that produced the image. Instead, we evaluate whether the learned representation can distinguish samples originating from known generators from samples that fall outside the known-generator manifold. The Open-set decision is therefore formulated as a rejection problem:
\begin{equation}
    f_{\mathrm{open}}(x)=
    \begin{cases}
        \mathrm{known}, & \text{if } x \text{ is compatible with } \mathcal{G}_{\mathrm{known}}, \\
        \mathrm{unknown}, & \text{otherwise}.
    \end{cases}
\end{equation}

Accordingly, the Open-set protocol used in this paper should be interpreted as known-versus-unknown rejection rather than unseen-generator attribution. Proto-LeakNet is trained exclusively on the Closed-set generators. After training, the attribution head is not used to assign labels to unseen generators. Instead, the frozen encoder is used to extract embeddings for known and unknown samples, and a density estimator is fitted only on the embeddings of the known generators. Samples with low likelihood under this known-generator density are interpreted as lying outside the known manifold and are therefore rejected as unknown.

\subsubsection{Evaluation Protocol}

The Closed-set protocol evaluates whether Proto-LeakNet can correctly attribute an image to one of the known generators in $\mathcal{G}_{\mathrm{known}}$. In this setting, both training and test classes belong to the same generator set, and performance is measured using standard multi-class attribution metrics such as per-class AUC and Macro AUC.

The Open-set protocol evaluates whether the learned representation supports rejection of samples from unseen generators. During training, no image from $\mathcal{G}_{\mathrm{unknown}}$ is used. After training, embeddings from the known generators are used to estimate the density of the known-generator manifold. Test samples from unseen generators are then scored according to their likelihood under this density. Open-set performance is measured by comparing the score distributions of known and unknown samples using threshold-independent and threshold-based metrics such as AUC, Equal Error Rate (EER), the overlap coefficient (OVL) and  False Positive Rate when the True Positive Rate is fixed at 95\% (FPR\@95). Details about the evaluation metrics are available in Section~\ref{sec:dataset}.

This distinction is important: Proto-LeakNet performs source attribution in the Closed-set setting, while in the Open-set setting it evaluates representation-level separability and unknown-source rejection. The purpose of the Open-set experiment is therefore not to classify unseen generators by identity, but to assess whether signal-leak-aware representations learned from known generators remain sufficiently structured to separate known sources from unseen ones.
\color{black}

\subsection{Latent Feature Construction}
We extract features directly from the latent domain of Stable Diffusion~2.1 (SD2.1).  
Each image $x \in \mathbb{R}^{3\times H\times W}$ is resized to $256\times256$ and encoded through the pretrained  SD2.1's Variational Autoencoder (VAE) into a latent $z_0$ scaled by the constant $s$ as in
\begin{equation}
    \label{eq:vae_encoding}
    z_{0} = \frac{\mathcal{E}(x)}{s},
\end{equation}
where $\mathcal{E}(\cdot)$ is the VAE encoder.  
To reveal and aggregate residual diffusion traces, we reapply the forward diffusion process at discrete steps $t\in\mathcal{T}=\{0,5,10\}$, sampling
\begin{equation}
    \label{eq:forward_diffusion}
    z_{t} = \alpha_{t} z_{0} + \sigma_{t}\,\varepsilon,
    \qquad \varepsilon \sim \mathcal{N}(0,I),
\end{equation}
where $t$ denotes the diffusion timestep and $(\alpha_t,\sigma_t)$ follow the cosine schedule with
\begin{equation}
    \label{eq:cosine_identity}
    \alpha_t^2 + \sigma_t^2 = 1.
\end{equation}
Each $z_t$ is normalized by $\sigma_t$ to maintain consistent scale, producing $\{z_t/\sigma_t\}_{t \in \mathcal{T}}$.  
We adopt the early-step set $\{0,5,10\}$ because these timesteps lie in the low-noise regime where generator-specific residuals remain most informative. This choice allows us both to study how the signal-leak evolves across noise levels and to temporally aggregate its most discriminative components before they are suppressed by high-variance noise at larger $t$.

\subsection{Discriminative Backbone}
The feature tensors $\{z_t\}_{t\in\mathcal{T}}$ are encoded by a ResNet18~\citep{he2015deepresidual} backbone $\phi(\cdot;\theta)$, with the first convolution adapted to match the channel dimensionality of $z_t$, where $\theta$ denotes the trainable backbone parameters.  
For each timestep $t$, the backbone produces a latent embedding
\begin{equation}
    \label{eq:backbone_embed}
    h_t = \phi(z_t;\,\theta) \in \mathbb{R}^{D},
\end{equation}
where $D$ is the embedding dimensionality.  
A learnable attention module assigns relevance to each timestep via
\begin{equation}
    \label{eq:attn_weights}
    a_t =
    \frac{\exp(q^{\top}u_t)}{\sum_{t'\in\mathcal{T}}\exp(q^{\top}u_{t'})},
    \qquad
    u_t = W_a h_t + b_a,
\end{equation}
where $q$, $W_a$, and $b_a$ are learned parameters.  
The temporally aggregated embedding is obtained as
\begin{equation}
    \label{eq:attn_pool}
    \bar{h} = \sum_{t\in\mathcal{T}} a_t\, h_t,
\end{equation}
where $\sum_t a_t = 1$.  
The weights $\{a_t\}$ provide temporal interpretability by quantifying the contribution of each diffusion step to the final embedding $\bar{h}$.

\subsection{Prototype-Based Attribution Head}
Each class $c\!\in\!\{1,\dots,C\}$ is represented by $M$ learnable prototypes, $p_{c,m}\!\in\!\mathbb{R}^{D}$, where $m\in M$, which serve as representative points in latent space.
Empirically, four prototypes yielded compact yet well-separated latent clusters.
A feature-wise attention gating vector $w\!\in\!(0,1)^D$ is computed using a small MLP:
\begin{equation}
    \label{eq:feat_gate}
    w = \sigma(A\bar{h} + b),
\end{equation}
where $A$ and $b$ are learnable parameters.  
The attention-weighted distance between $\bar{h}$ and each prototype $p_{c,m}$ is defined as
\begin{equation}
    \label{eq:proto_dist}
    d_{c,m}(\bar{h},w) =
    \sum_{i=1}^{D} w_i \, (\bar{h}_i - p_{c,m,i})^2.
\end{equation}
Per-class scores aggregate distances using a temperature-controlled LogSumExp:
\begin{equation}
    \label{eq:lse_score}
    s_c = -\tau \log \sum_{m=1}^{M} \exp\!\Big(-\tfrac{1}{\tau}d_{c,m}(\bar{h},w)\Big),
\end{equation}
where $\tau>0$ is a learnable scalar that determines the aggregation smoothness.
Posterior probabilities are computed as
\begin{equation}
    \label{eq:softmax_logits}
    \pi_c = \frac{\exp(s_c)}{\sum_{c'} \exp(s_{c'})},
\end{equation}
where $c'$ denotes the index iterating over all classes in the denominator.  
The model is trained via cross-entropy loss
\begin{equation}
    \label{eq:ce_loss}
    \mathcal{L}_{\text{CE}} = -\frac{1}{B}\sum_{b=1}^{B}\log\pi_{y^{(b)}},
\end{equation}
where $B$ is the mini-batch size and $y^{(b)}$ are ground-truth labels.  
All parameters $\{\theta, A, b, p_{c,m}\}$ are optimized jointly with AdamW and weight decay.

\subsubsection*{Mahalanobis Scoring and Interpretability}
During evaluation, embeddings are scored using a diagonal Mahalanobis comparator fitted on training embeddings.  
For each class $c$, with empirical mean $\mu_c$ and diagonal covariance $\Sigma_c=\mathrm{diag}(\sigma^2_{c,1},\dots,\sigma^2_{c,D})$, the score is
\begin{equation}
    \label{eq:maha_score}
    s_c^{\text{maha}}(\bar{h}) =
    -\sum_{i=1}^{D}
    \frac{(\bar{h}_i - \mu_{c,i})^2}{\sigma_{c,i}^2+\epsilon},
\end{equation}
where $\sigma_{c,i}$ represents the empirical standard deviation of feature $i$ within class $c$, capturing the intra-class variance along each embedding dimension and $\epsilon$ prevents numerical instability.  
These scores provide calibrated likelihoods for attribution and Open-set evaluation.

\subsubsection*{Interpretability of Proto-LeakNet}
Proto-LeakNet is inherently interpretable.  
Each distance in Eq.~\ref{eq:proto_dist} decomposes feature-wise as
\begin{equation}
    \label{eq:explainability_distance}
    d_{c,m}(\bar{h},w) =
    \sum_{i=1}^{D}
    \underbrace{w_i(\bar{h}_i - p_{c,m,i})^2}_{r_{c,m,i}},
\end{equation}
where $r_{c,m,i}$ measures the contribution of feature $i$.  
Prototype responsibilities are obtained as
\begin{equation}
    \label{eq:proto_resp}
    \pi_{c,m} =
    \frac{\exp(-d_{c,m}/\tau)}{\sum_{m'} \exp(-d_{c,m'}/\tau)}.
\end{equation}

The most activated prototype $\arg\max_m \pi_{c,m}$ identifies the latent region that best matches the input.  
Together, the feature gates $w$, temporal weights $\{a_t\}$, and prototype responsibilities $\{\pi_{c,m}\}$ provide a three-level interpretability hierarchy, illustrated in Fig.~\ref{fig:evo_clusters} in detail with the effective distribution.

\begin{center}
  \includegraphics[width=1\linewidth]{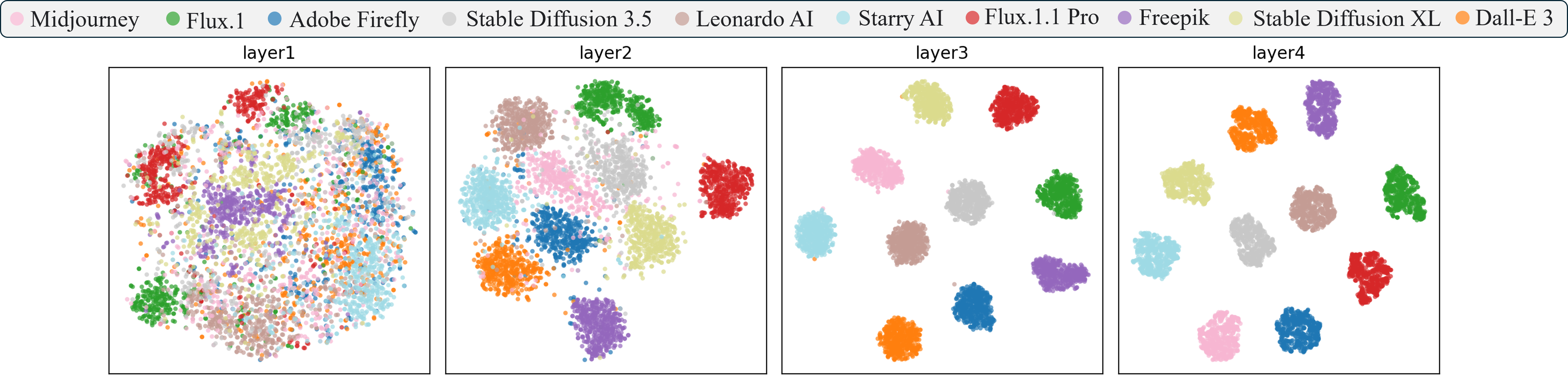}
  \captionof{figure}{Interpretability of Proto-LeakNet.
  Layer-wise evolution of the embedding space, illustrating an approximate progression of how prototypes and attention progressively refine and generalize class separation across layers.}
  \label{fig:evo_clusters}
\end{center}



\subsection{Representation-Level Generalization via Density Estimation}
\label{sec:rep_level}
While Proto-LeakNet is trained only on Closed-set generators, its latent encoder learns a structured representation that can be evaluated for generalization without requiring explicit supervision on unseen classes.  
In this setting, the goal is not to classify open samples, but to assess whether the learned latent geometry consistently separates embeddings of known generators from those of unseen ones.
After training, we discard the prototype-based classifier and use only the frozen ResNet18 backbone-based encoder to produce pooled embeddings 
$\bar{h}\!\in\!\mathbb{R}^{D}$ for both closed and open samples, forming the sets
$\mathcal{H}_{\mathrm{c}}=\{\bar{h}^{(\mathrm{c})}_{i}\}_{i=1}^{N_{\mathrm{c}}}$ and
$\mathcal{H}_{\mathrm{o}}=\{\bar{h}^{(\mathrm{o})}_{j}\}_{j=1}^{N_{\mathrm{o}}}$.  
A Gaussian kernel density estimator (KDE) is fitted on $\mathcal{H}_{\mathrm{c}}$ to model the manifold of closed embeddings:
\begin{equation}
\label{eq:kde_density}
\resizebox{0.3\columnwidth}{!}{$
p_{\text{KDE}}(h)
= \frac{1}{N_{\mathrm{c}}(2\pi\sigma^{2})^{\frac{D}{2}}}
  \sum_{i=1}^{N_{\mathrm{c}}}
  \exp\!\left(
      -\frac{\|h-\bar{h}^{(\mathrm{c})}_{i}\|^{2}}
            {2\sigma^{2}}
  \right)
$}
\end{equation}
where $\sigma$ is the kernel bandwidth and $D$ is the embedding dimensionality.  
For each sample $h$, we compute its log-likelihood score
\begin{equation}
    \label{eq:kde_score}
    s(h)=\log p_{\text{KDE}}(h),
\end{equation}
which measures how likely $h$ lies within the distribution of known generators.  
High $s(h)$ values correspond to familiar latent regions, while low scores indicate that the sample is far from any known manifold, suggesting an unseen generator.
This approach defines a form of representation-level generalization: the model is never trained on open samples or labels, yet its latent space forms structured low-density regions that naturally reject out-of-distribution inputs. Rather than predicting unseen categories, the KDE analysis evaluates whether the learned representation preserves discriminative geometry under domain shifts.
Open generators, though unlabeled, consistently occupy regions separated from the closed manifold, showing that the encoder captures signal-leak statistics that generalize beyond training sources and enable unsupervised detection of unseen generators.

\color{black}
\subsection{Computational Complexity and Efficiency}
\label{sec:complexity}

The computational cost of Proto-LeakNet is dominated by latent extraction and multi-timestep encoding. Each image is first passed once through the pretrained Stable Diffusion VAE encoder to obtain a compact latent tensor $z_0 \in \mathbb{R}^{4 \times 32 \times 32}$. The partial forward-diffusion simulation over $\mathcal{T}=\{0,5,10\}$ requires only element-wise operations and is negligible compared with the neural network forward passes.

The main cost is therefore the application of the ResNet-18 encoder to each timestep, yielding complexity
\begin{equation}
    \mathcal{O}\big(|\mathcal{T}| \cdot C_{\mathrm{ResNet}}\big),
\end{equation}
where $C_{\mathrm{ResNet}}$ is the cost of one ResNet18 forward pass on a latent tensor. Since $|\mathcal{T}|=3$, this introduces a fixed and predictable overhead, while operating on compact $4 \times 32 \times 32$ latent inputs rather than full-resolution RGB images.

The temporal attention module has complexity $\mathcal{O}(|\mathcal{T}|D)$, and the prototype head has complexity $\mathcal{O}(CMD)$, where $C$ is the number of classes, $M$ the number of prototypes per class, and $D$ the embedding dimension. These terms are small compared with the convolutional backbone. For open-set rejection, direct KDE scoring requires $\mathcal{O}(N_{\mathrm{known}}D)$ per test sample, but this step is used only after training and can be accelerated through subsampling, class-wise density estimation, or approximate nearest-neighbor search.

Thus, Proto-LeakNet introduces a linear overhead in the number of selected timesteps, while the attention and prototype components add negligible computational cost.
\color{black}

\section{Dataset and Evaluation Metrics}
\label{sec:dataset}


Our evaluation is conducted on two datasets. The primary benchmark is WILD~\citep{bongini2025wild}, a 20k-image benchmark comprising high-quality samples from a closed set of text-to-image generators and an open set of additional state-of-the-art models, with no prompt or model overlap. WILD is selected for its realistic, artifact-free images, preventing attribution bias rooted in visible flaws. The dataset is organized into three subsets: a closed set of known generators used for training and testing, a post-processed version of the closed set introducing realistic degradations, and an open set of unseen generators used exclusively for testing. In addition, we introduce a separate Partial Manipulation Dataset, described at the end of this section, designed to evaluate Proto-LeakNet under a more challenging scenario involving localized image editing.

As described above, WILD is organized into three main configurations:
\color{black}

\noindent \textbf{Closed set (10k images).} The closed set includes ten text-to-image models: Adobe Firefly~\cite{firefly}, DALL\-E 3~\cite{dalle3}, FLUX.1~\cite{flux2024}, FLUX 1.1 Pro~\cite{fluxpro}, Freepik~\cite{freepik}, Leonardo AI~\cite{leonardoai}, Midjourney~\cite{midjourney}, Stable Diffusion 3.5 Large~\cite{stable_diffusion_35_large}, Stable Diffusion XL Turbo~\cite{sdxl_turbo}, and StarryAI~\cite{starryai}. A pool of 1,000 prompts is applied uniformly across all generators. The official split (also used for our experiments) contains 5,000 training images, 2,000 validation images, and 3,000 test images, with splits defined at the prompt level. \\
\noindent \textbf{Post-Processed Closed-Set.}
 The Closed-set post-processed test images underwent 1 (Step 1), 2 (Step 2), and 3 (Step 3) random transformations selected from compression, cropping, resizing, rotation, blur, photometric changes, grayscale conversion, and super-resolution.\\
\noindent \textbf{Open set (10k images, test-only).} The open set includes ten additional generators, again with 1,000 images each: DALL\-E 1 \cite{ramesh2021zeroshot}, DeepAI \cite{deepai}, HotpotAI \cite{hotpotai}, NVIDIA Sana \cite{xie2024sana},
Stable Cascade \cite{pernias2023wuerstchen}, Stable Diffusion Attend\&Excite \cite{sd_ae}, StyleGAN \cite{karras2019stylebasedgeneratorarchitecturegenerative}, StyleGAN2 \cite{karras2020analyzingimprovingimagequality}, StyleGAN3 \cite{karras2021alias}, and Tencent Hunyuan \cite{li2024hunyuandit}.
This set includes GAN models and text-to-image diffusion/transformer models not found in the closed set. All Open-set images were used exclusively for testing.
Figure \ref{fig:dataset} shows some examples of the involved datasets.

\begin{center}
  \includegraphics[width=.9\linewidth]{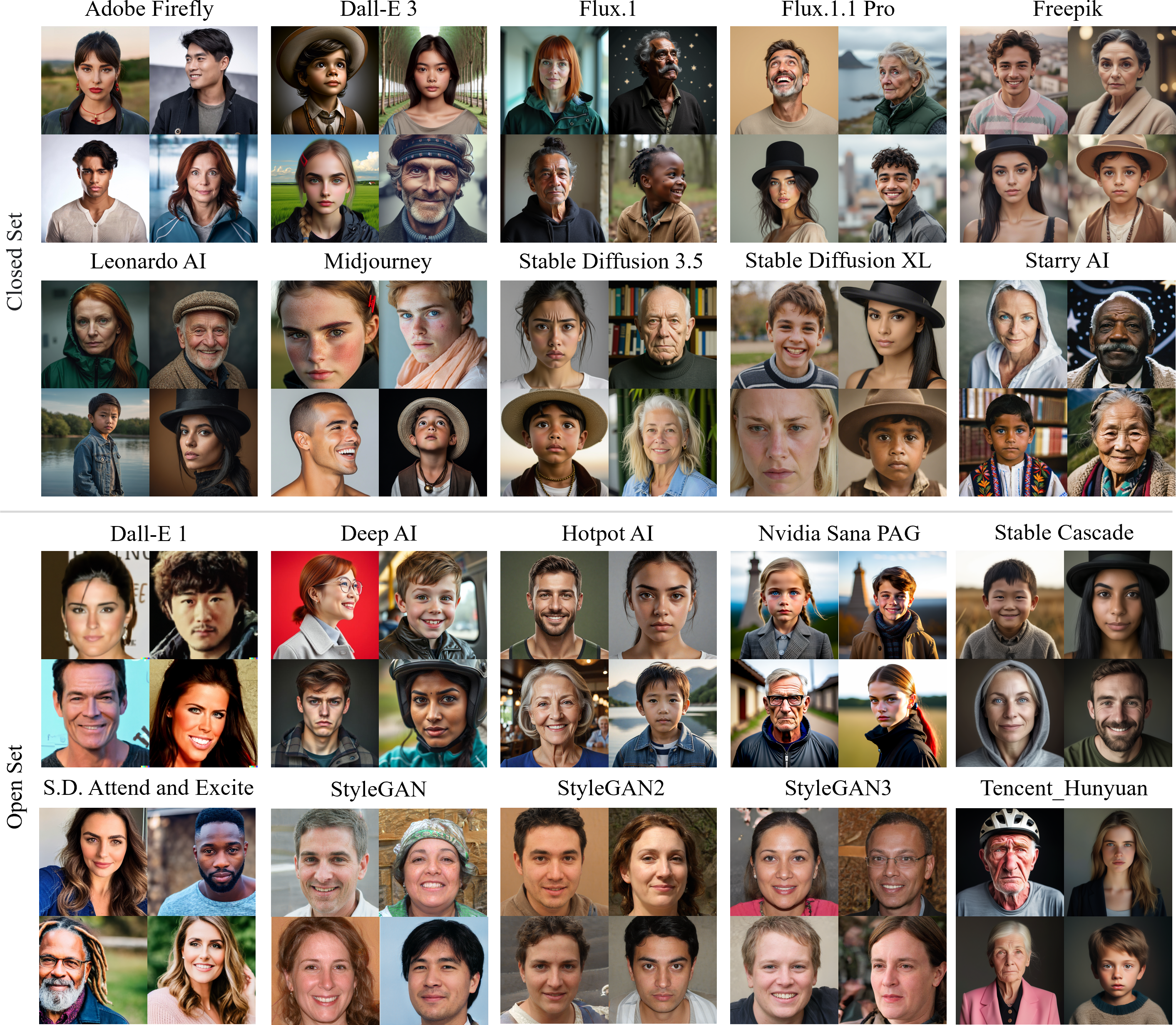}
  \captionof{figure}{Examples from the WILD dataset used in the experimental evaluation. The images show portrait samples of individuals with diverse ages, ethnicities, and genders.}
  \label{fig:dataset}
\end{center}

In addition to WILD, we introduce a separate dataset specifically designed to evaluate Proto-LeakNet under a more challenging and realistic scenario involving partial image manipulation. \\
\noindent
\textbf{Partial Manipulation Dataset.} This dataset consists of 5,000 high-resolution face portraits sourced from the FFHQ dataset~\footnote{\url{https://github.com/NVlabs/ffhq-dataset}, Last Access: March 14, 2026}, each representing a real photograph in which only a localized facial region has been altered by one of four state-of-the-art editing models: FLUX.1-Kontext~\citep{batifol2025flux}, Qwen2-VL~\citep{wu2025qwen}, GLM-Image~\footnote{\url{https://z.ai/blog/glm-image}, Last Access: March 14, 2026}, and Instruct-Pix2Pix~\citep{brooks2023instructpix2pix}. All models were used in their original pretrained configurations, without any fine-tuning. Manipulations were applied following a standardized set of 100 facial editing prompts, covering a range of modifications from structural alterations (e.g., eye size, nose shape) to accessory additions (e.g., sunglasses, earrings). The same prompt set was applied uniformly across all four models to ensure comparability. The dataset is organized in two levels of manipulation depth: Step 1 contains images manipulated once from the original, while Step 2 contains images subjected to a second sequential manipulation applied to the Step 1 output, using the same or a different model. Since the extent of each manipulation varies depending on the prompt and the model, the resulting dataset covers a broad spectrum of alteration intensities, from subtle localized edits to more extensive modifications. This variability is intentional: when the manipulation is minor, the generator-specific signal-leak traces are expected to be weaker, posing a harder attribution challenge; conversely, more extensive edits are expected to leave stronger and more discriminative latent cues. Figure \ref{fig:dataset2} shows representative examples of the manipulations produced by each editing model. The evaluation on this dataset is reported in Section \ref{sec:eval_part_dataset}.

\color{black}
\noindent
\textbf{Open-set Generalization on External Datasets and Unseen Diffusion Models.}
In addition to the datasets described above, we also consider external real-image datasets and additional unseen generative sources not included in the WILD Closed-set training split for further Open-set generalization analysis. In particular, CelebA-HQ~\citep{liu2015faceattributes,karras2018progressive}, FFHQ~\citep{karras2018ffhq}, and ImageNet~\citep{deng2009imagenet} are used to evaluate representation-level rejection of real images outside the training distribution. Specifically, the entire CelebA-HQ dataset (30,000 images), the full FFHQ dataset (70,000 images), and a subset of 50,000 images from ImageNet are considered for evaluation. Moreover, Stable Diffusion 1.5, Stable Diffusion 2.1~\cite{sd-faces}, and Z-Turbo~\cite{z-image-2025} are employed as additional unseen diffusion generators, each containing 3,500 generated images. These datasets and generators are used exclusively at test time and are never involved in training, prototype learning, density estimation, threshold selection, or calibration.
\color{black}

\begin{center}
  \includegraphics[width=.9\linewidth]{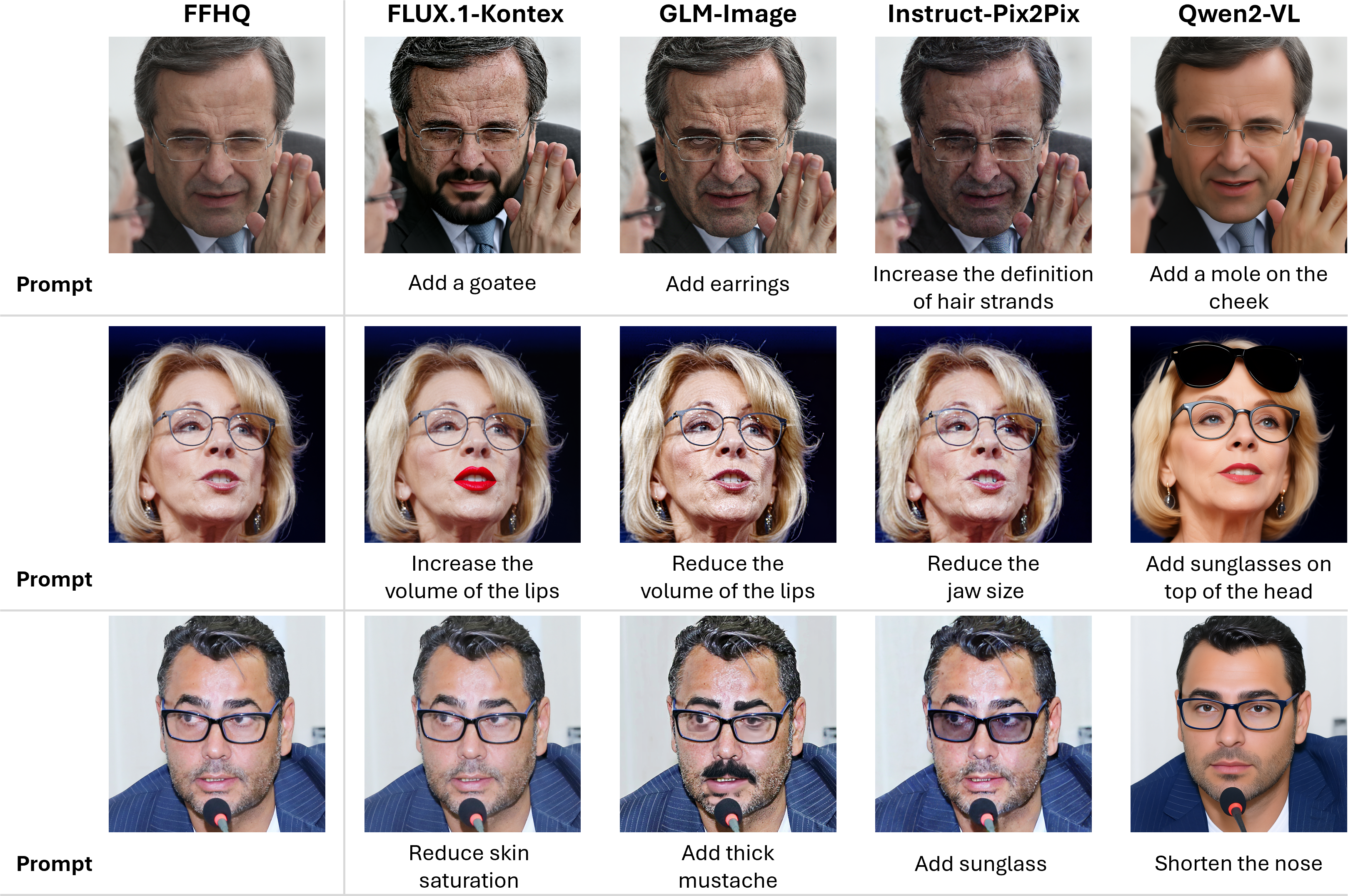}
  \captionof{figure}{Examples from the Partial Manipulation Dataset. Each row shows a real portrait from FFHQ (first column) alongside the corresponding manipulated versions produced by the four editing models: FLUX.1-Kontext, Qwen2-VL, GLM-Image, and Instruct-Pix2Pix (columns 2 to 5). The manipulations vary in extent and style across models, ranging from subtle localized edits to more visible alterations, reflecting the diversity of generator-specific traces present in the dataset.}
  \label{fig:dataset2}
\end{center}

\color{black}

\color{black}
\subsection{Evaluation Metrics}
We use different metrics for the Closed-set attribution and Open-set rejection protocols. 
For Closed-set source attribution, we report Macro AUC, which measures the discriminative performance across the known generator classes. 
For Open-set scoring, we report Equal Error Rate (EER), Overlap Coefficient (OVL), and FPR@95, which quantify the separability between known-generator samples and samples from unseen sources.

\begin{itemize}
    \item \textbf{Macro AUC:} For Closed-set source attribution, we compute the per-class area under the ROC curve (AUC) and average it over all $C$ known generator classes:
    \begin{equation}
        \label{eq:macro_auc}
        \text{Macro AUC} = \frac{1}{C}\sum_{c=1}^{C}\text{AUC}_c,
    \end{equation}
    where $\text{AUC}_c$ measures the ranking quality for class $c$ against all remaining known classes. 
    This metric evaluates threshold-independent discriminative consistency across the Closed-set generator classes.

    \item \textbf{Equal Error Rate (EER):} For Open-set rejection, EER is the operating point at which the false-acceptance rate (FAR) equals the false-rejection rate (FRR):
    \begin{equation}
        \label{eq:eer_def}
        \text{EER} = \min_{\delta}\big|\text{FAR}(\delta)-\text{FRR}(\delta)\big|,
    \end{equation}
    where $\delta$ denotes the decision threshold applied to the Open-set score. 
    Lower EER indicates better separation between known and unknown samples.

    \item \textbf{Overlap Coefficient (OVL):} The overlap coefficient quantifies the empirical intersection between the score distributions of known and unknown samples:
    \begin{equation}
        \label{eq:ovl}
        \text{OVL} = \int \min\big(P_{\text{known}}(s),\, P_{\text{unknown}}(s)\big)\, ds,
    \end{equation}
    where $P_{\text{known}}$ and $P_{\text{unknown}}$ are normalized density estimates of the Open-set scores $s$ for known and unknown samples, respectively. 
    Values close to zero indicate nearly non-overlapping score distributions and therefore stronger Open-set separability.

    \item \textbf{FPR@95:} We also report the false positive rate at 95\% true positive rate, denoted as FPR@95. 
    This metric measures the fraction of unknown samples incorrectly accepted as known when the threshold is chosen so that 95\% of known samples are correctly accepted:
    \begin{equation}
        \label{eq:fpr95}
        \text{FPR@95} = \text{FPR}(\delta_{95}),
        \quad \text{where} \quad \text{TPR}(\delta_{95}) = 0.95.
    \end{equation}
    Lower FPR@95 indicates stronger rejection of unknown samples at a high known-sample acceptance rate.
\end{itemize}
\color{black}

\begin{table}[t!]
\centering
\caption{AUC (\%) results for each generator class on the raw Closed-set. The table reports both per-class AUC and overall Macro AUC, results in bold are the best, underlined ones are second best.}
\label{tab:all_aucs_raw}
\begin{adjustbox}{width=1\linewidth}
\begin{tabular}{clcccccccccc|
>{\columncolor[HTML]{EFEFEF}}c }
\cline{3-13}
\textbf{}                         &                        & \multicolumn{10}{c|}{\textbf{Closed-set Classes}}                                                                                                                                                                                                                                                     & \cellcolor[HTML]{EFEFEF}                                     \\ \cline{1-12}
\textbf{Type}                     & \textbf{Methodologies} & \textbf{Adobe Firefly} & \textbf{Dall-E 3} & \textbf{Flux.1} & \textbf{Flux.1.1 Pro} & \textbf{Freepik} & \textbf{Leonardo AI} & \textbf{Midjourney} & \textbf{\begin{tabular}[c]{@{}c@{}}Stable \\ Diffusion 3.5\end{tabular}} & \textbf{\begin{tabular}[c]{@{}c@{}}Stable \\ Diffusion XL\end{tabular}} & \textbf{Starry AI} & \multirow{-2}{*}{\cellcolor[HTML]{EFEFEF}\textbf{Macro AUC}} \\ \hline
                                  & WILD: EfficientNet\_B4~\citep{bongini2025wild} & 95.90                   & \textbf{96.53}     & 96.59           & 98.80                 & \textbf{99.88}   & 97.28               & 94.83               & 95.55                                                                 & \textbf{99.99}                                                         & 95.69             & 97.10                                                   \\
                                  & WILD: XceptionNet~\citep{bongini2025wild}      & 94.58                   & 94.90              & 96.94           & 98.81                 & 99.16            & 94.04               & 91.41               & 91.85                                                                 & \underline{99.90}                                                     & 91.60             & 95.32                                                   \\
                                  & WILD: ResNet50~\citep{bongini2025wild}         & 94.28                   & 90.79              & 97.20           & 97.05                 & 99.34            & 98.05               & 89.02               & 96.53                                                                 & 97.57                                                                 & 94.13             & 95.40                                                   \\
                                  & FreqNet~\citep{freqnet2023}                & 96.82                   & 95.46              & 99.39           & 99.58                 & 99.43            & 97.87               & 95.44               & 97.11                                                                 & 97.90                                                                 & 96.60             & 97.56                                                   \\
                                  & SuSy~\citep{susy2024}                   & \textbf{98.10}          & 96.06              & 99.80           & 99.74                 & 99.63            & 98.56               & \underline{96.04}   & \underline{97.68}                                                     & 97.27                                                                 & \underline{97.32} & \underline{98.02}                                       \\
                                  & LatentTracer~\citep{latenttracer2024}           & 97.29                   & 95.92              & \underline{99.87} & \underline{99.80}     & 99.71            & \underline{98.57}   & 95.66               & 97.50                                                                 & 97.18                                                                 & 97.30             & 97.88                                                   \\
                                  & OCC-CLIP~\citep{occlip2024}               & 94.39                   & 93.12              & 98.54           & 97.42                 & 98.25            & 96.21               & 93.45               & 94.60                                                                 & 95.68                                                                 & 95.34             & 95.70                                                   \\
                                  & NPR~\citep{npr2024}                  & 88.34                   & 86.12              & 94.10           & 93.54                 & 93.82            & 90.53               & 86.05               & 87.91                                                                 & 89.02                                                                 & 88.71             & 89.81                                                   \\
                                  & LATTE~\citep{vasilcoiu2025latte}              & 96.39                   & 94.98              & 99.05           & 98.96                 & 98.83            & 97.70               & 94.95               & 96.88                                                                 & 97.35                                                                 & 96.48             & 97.16                                                   \\ \cline{2-13}
\multirow{-10}{*}{\textbf{RAW}}   & \textbf{Proto-LeakNet}          & \underline{97.32}       & \underline{96.17}  & \textbf{99.91}  & \textbf{99.85}        & \underline{99.74} & \textbf{98.67}      & \textbf{96.15}      & \textbf{97.79}                                                      & 98.38                                                                 & \textbf{97.35}    & \textbf{98.13}                                          \\ \hline
\end{tabular}
\end{adjustbox}
\end{table}

\section{Experimental Results}
\label{sec:experiments}

We conduct a comprehensive evaluation of Proto-LeakNet under multiple experimental settings to assess its attribution accuracy, robustness to post-processing, and generalization capability in the presence of unseen data. 
We compare our approach against the top-performing methods reported on WILD~\citep{bongini2025wild}, as well as representative state-of-the-art approaches, including FreqNet~\citep{freqnet2023}, SuSy~\citep{susy2024}, LatentTracer~\citep{latenttracer2024}, OCC-CLIP~\citep{occlip2024}, NPR~\citep{npr2024}, and LATTE~\citep{vasilcoiu2025latte}.
\color{black}Some baselines, such as NPR and FreqNet, were originally introduced for generated-image detection rather than source attribution. 
In this work, they are adapted to the multi-class WILD Closed-set protocol and used as forensic-signal baselines for generator discrimination. 
This allows us to compare Proto-LeakNet against representative forensic strategies based on pixel-level residuals, frequency-domain artifacts, patch-level inconsistencies, and latent-space representations.\color{black}
Given the conceptual proximity between Proto-LeakNet and LATTE, both operating in diffusion latent space, we adapt LATTE to the source attribution setting by extending its original binary detection objective to a multi-class formulation while preserving its architectural design. Owing to its latent-space modeling strategy, LATTE is naturally positioned to capture generator-specific characteristics, making it a relevant and competitive baseline for origin attribution.
Beyond Closed-set performance, we investigate the impact of compression and post-processing strategies on attribution accuracy, and we analyze the contribution of signal-leak exploitation within our pipeline. We further examine the interpretability of the temporal aggregation pooling module by analyzing its learned weighting behavior across diffusion steps. Finally, we conduct ablation studies to isolate and quantify the contribution of each architectural component, providing a detailed assessment of the factors driving performance gains.

\subsection{Results on Closed-set}
Proto-LeakNet is first evaluated under the Closed-set configuration to assess its ability to learn and discriminate generator-specific signal-leak patterns when all classes observed at test time are available during training. All competing methods are trained and evaluated under identical data splits and experimental conditions to ensure a fair comparison.
Table~\ref{tab:all_aucs_raw} reports the per-class AUC scores. While Proto-LeakNet does not achieve the highest score for every individual generator, it consistently ranks among the top-performing methods across all classes, resulting in the highest overall Macro~AUC. This behavior indicates improved balance across generators and stronger generalization in capturing generator-specific latent signatures.
Compared to the baselines, Proto-LeakNet exhibits enhanced latent discriminability and more structured class separation. The combination of temporal attention and prototype supervision plays a central role in this behavior: temporal attention selectively emphasizes the most informative diffusion timesteps, while prototype-based supervision encourages geometric consistency in the embedding space by attracting samples toward class centroids and increasing inter-class margins.
Figure~\ref{fig:plot_accuracies} presents the Top-1 accuracy distributions. Although SuSy~\citep{susy2024} slightly outperforms Proto-LeakNet on raw images (83.32\% vs.~82.60\%), our model achieves the highest Macro~AUC, indicating superior threshold-independent separability. This improved separability translates into a more compact and well-organized latent representation, as illustrated in Fig.~\ref{fig:evo_clusters}, where generator-specific signal-leak cues form coherent and distinguishable regions that support reliable attribution.

\begin{table*}[t!]
\centering
\caption{AUC (\%) results for each generator class under increasing post-processing levels (Steps 1–3). The last column reports Macro AUC for all models. Results in bold are the best, underlined ones are second best.}
\label{tab:all_aucs}
\begin{adjustbox}{width=1\linewidth}
\begin{tabular}{clcccccccccc|
>{\columncolor[HTML]{EFEFEF}}c }
\cline{3-13}
\textbf{}                         &                        & \multicolumn{10}{c|}{\textbf{Closed-set Classes}}                                                                                                                                                                                                                                                     & \cellcolor[HTML]{EFEFEF}                                     \\ \cline{1-12}
\textbf{Type}                     & \textbf{Methodologies} & \textbf{Adobe Firefly} & \textbf{Dall-E 3} & \textbf{Flux.1} & \textbf{Flux.1.1 Pro} & \textbf{Freepik} & \textbf{Leonardo AI} & \textbf{Midjourney} & \textbf{\begin{tabular}[c]{@{}c@{}}Stable \\ Diffusion 3.5\end{tabular}} & \textbf{\begin{tabular}[c]{@{}c@{}}Stable \\ Diffusion XL\end{tabular}} & \textbf{Starry AI} & \multirow{-2}{*}{\cellcolor[HTML]{EFEFEF}\textbf{Macro AUC}} \\ \hline
                                  & WILD: EfficientNet\_B4~\citep{bongini2025wild} & \textbf{96.46}          & \textbf{94.43}     & 94.97           & 92.43                 & 97.63            & 94.05               & \textbf{94.49}      & 94.84                                                                 & 89.84                                                                 & 94.83             & 94.40                                                   \\
                                  & WILD: XceptionNet~\citep{bongini2025wild}      & 91.79                   & \underline{94.30}  & 96.01           & 97.62                 & 98.12            & 93.06               & 90.98               & 88.92                                                                 & 89.10                                                                 & 93.71             & 93.36                                                   \\
                                  & WILD: ResNet50~\citep{bongini2025wild}         & 93.19                   & 91.62              & 97.63           & 93.81                 & \textbf{98.35}   & 96.51               & 87.66               & 94.93                                                                 & \textbf{98.01}                                                        & 92.09             & 94.38                                                   \\
                                  & FreqNet~\citep{freqnet2023}                & 94.98                   & 93.29              & 97.90           & 97.78                 & 97.62            & \textbf{96.74}      & 93.15               & 94.62                                                                 & 95.40                                                                 & 94.62             & 95.61                                                   \\
                                  & SuSy~\citep{susy2024}                   & 95.17                   & 93.61              & 98.12           & 97.98                 & 97.84            & 96.27               & 93.57               & 94.95                                                                 & 95.71                                                                 & \textbf{95.78}    & 95.90                                                   \\
                                  & LatentTracer~\citep{latenttracer2024}           & 95.33                   & 93.87              & \underline{98.29} & 98.15                 & 97.99            & 96.44               & 93.84               & \textbf{95.21}                                                        & \underline{95.98}                                                     & 95.36             & \underline{96.05}                                       \\
                                  & OCC-CLIP~\citep{occlip2024}               & 93.18                   & 91.21              & 96.74           & 96.58                 & 96.39            & 94.36               & 91.17               & 92.68                                                                 & 93.79                                                                 & 92.96             & 93.91                                                   \\
                                  & NPR~\citep{npr2024}                    & 79.73                   & 74.55              & 87.15           & 86.66                 & 87.34            & 79.12               & 75.43               & 76.00                                                                 & 80.76                                                                 & 78.05             & 80.48                                                   \\
                                  & LATTE~\citep{vasilcoiu2025latte}            & 95.21                   & 93.79              & 98.16           & \underline{98.19}     & 98.02            & 96.43               & 93.76               & 95.14                                                                 & 95.81                                                                 & 95.43             & 95.99                                                   \\ \cline{2-13}
\multirow{-10}{*}{\textbf{Step 1}} & \textbf{Proto-LeakNet}         & \underline{95.51}       & 93.72              & \textbf{98.43}  & \textbf{98.31}        & \underline{98.27} & \underline{96.54}   & \underline{93.88}   & \underline{95.15}                                                     & 95.93                                                                 & \underline{95.69} & \textbf{96.14}                                          \\ \hline
                                  & WILD: EfficientNet\_B4~\citep{bongini2025wild} & 92.58                   & 90.53              & 92.59           & 91.88                 & \underline{96.89} & 93.49               & \textbf{92.04}      & 92.62                                                                 & 84.97                                                                 & 90.60             & 91.82                                                   \\
                                  & WILD: XceptionNet~\citep{bongini2025wild}      & 92.90                   & \textbf{92.94}     & 94.66           & \textbf{97.33}        & \textbf{96.91}   & 93.21               & 91.12               & 87.41                                                                 & 89.54                                                                 & \underline{92.26} & 92.83                                                   \\
                                  & WILD: ResNet50~\citep{bongini2025wild}         & 89.61                   & 90.96              & 96.17           & 90.28                 & 95.23            & \textbf{97.55}      & 86.80               & \textbf{94.13}                                                        & \textbf{96.59}                                                        & 89.14             & 92.65                                                   \\
                                  & FreqNet~\citep{freqnet2023}                & 88.74                   & 86.58              & 94.53           & 94.70                 & 94.46            & 91.07               & 85.52               & 88.61                                                                 & 89.89                                                                 & 88.82             & 90.29                                                   \\
                                  & SuSy~\citep{susy2024}                   & 92.13                   & 89.74              & 96.19           & 96.03                 & 95.86            & 93.01               & 89.54               & 91.20                                                                 & 92.37                                                                 & 91.63             & 92.77                                                   \\
                                  & LatentTracer~\citep{latenttracer2024}           & 91.89                   & 89.49              & 95.08           & 95.79                 & 95.60            & 93.11               & 89.43               & 91.26                                                                 & 92.09                                                                 & 91.57             & 92.53                                                   \\
                                  & OCC-CLIP~\citep{occlip2024}               & 92.03                   & 89.62              & 96.12           & 95.91                 & 95.49            & 93.30               & 89.57               & 91.17                                                                 & 92.28                                                                 & 91.21             & 92.67                                                   \\
                                  & NPR~\citep{npr2024}                    & 75.71                   & 70.39              & 83.37           & 82.89                 & 83.64            & 75.22               & 69.34               & 72.15                                                                 & 75.11                                                                 & 77.01             & 76.48                                                   \\
                                  & LATTE~\citep{vasilcoiu2025latte}             & \textbf{95.73}          & 91.82              & \underline{96.99} & 96.92               & 96.87            & 94.79               & 91.71               & 93.13                                                                 & 94.03                                                                 & 92.09             & \underline{94.41}                                       \\ \cline{2-13}
\multirow{-10}{*}{\textbf{Step 2}} & \textbf{Proto-LeakNet}         & \underline{93.95}       & \underline{91.98}  & \textbf{97.17}  & \underline{97.01}     & 96.86            & \underline{95.46}   & \underline{91.96}   & \underline{93.46}                                                     & \underline{94.53}                                                     & \textbf{93.80}    & \textbf{94.62}                                          \\ \hline
                                  & WILD: EfficientNet\_B4~\citep{bongini2025wild} & 90.07                   & 88.12              & 89.65           & 88.88                 & 92.54            & 89.03               & 86.97               & 89.89                                                                 & 71.86                                                                 & 86.40             & 87.34                                                   \\
                                  & WILD: XceptionNet~\citep{bongini2025wild}      & 89.57                   & \textbf{90.88}     & 93.84           & \textbf{97.05}        & 94.73            & \textbf{93.66}      & 85.69               & 86.06                                                                 & 83.95                                                                 & \textbf{92.24}    & 90.77                                                   \\
                                  & WILD: ResNet50~\citep{bongini2025wild}         & 88.40                   & \underline{90.46}  & 91.92           & 89.79                 & 94.00            & 92.45               & 82.48               & 89.03                                                                 & \textbf{95.10}                                                        & 86.59             & 90.02                                                   \\
                                  & FreqNet~\citep{freqnet2023}                & 88.72                   & 85.51              & 93.91           & 93.74                 & 93.52            & 90.06               & 84.81               & 87.55                                                                 & 88.19                                                                 & 87.38             & 89.34                                                   \\
                                  & SuSy~\citep{susy2024}                   & 89.82                   & 86.68              & 94.95           & 94.79                 & 94.58            & 91.24               & 86.01               & 87.13                                                                 & 89.99                                                                 & 88.83             & 90.40                                                   \\
                                  & LatentTracer~\citep{latenttracer2024}           & 87.43                   & 83.57              & 93.18           & 92.85                 & 92.51            & 89.24               & 83.49               & 81.27                                                                 & 88.73                                                                 & 88.63             & 88.09                                                   \\
                                  & OCC-CLIP~\citep{occlip2024}               & \underline{91.37}       & 88.79              & \underline{95.48} & 95.23                 & \underline{94.75} & 92.58               & \textbf{90.54}      & \textbf{91.41}                                                        & 91.13                                                                 & 88.63             & \underline{91.99}                                       \\
                                  & NPR~\citep{npr2024}                    & 71.77                   & 64.29              & 78.23           & 77.34                 & 78.71            & 70.51               & 65.38               & 66.42                                                                 & 69.94                                                                 & 68.91             & 71.15                                                   \\
                                  & LATTE~\citep{vasilcoiu2025latte}             & 90.32                   & 87.68              & 94.45           & 94.20                 & 94.01            & 91.52               & 87.61               & 89.50                                                                 & 90.36                                                                 & 89.95             & 90.96                                                   \\ \cline{2-13}
\multirow{-10}{*}{\textbf{Step 3}} & \textbf{Proto-LeakNet}         & \textbf{91.62}          & 89.24              & \textbf{95.63}  & \underline{95.46}     & \textbf{95.29}   & \underline{92.84}   & \underline{89.18}   & \underline{90.35}                                                     & \underline{91.61}                                                     & \underline{91.30} & \textbf{92.25}                                          \\ \hline
\end{tabular}
\end{adjustbox}
\end{table*}

\subsection{Results on Post-Processed Closed-set}
To evaluate robustness under realistic degradations, all models, including Proto-LeakNet, are assessed on progressively post-processed Closed-set samples using checkpoints trained on the raw configuration. During evaluation, perturbations from Steps~1–3 (Section~\ref{sec:dataset}) are applied exclusively at inference time, without any additional fine-tuning.
Table~\ref{tab:all_aucs} reports the per-class AUC scores under increasing degradation levels. Although Proto-LeakNet does not achieve the highest score for every individual generator, it consistently maintains strong performance across all classes, resulting in the highest overall Macro~AUC at each degradation level. This behavior indicates improved resilience to visual distortions compared to image-domain baselines, whose performance degrades more significantly as perturbation strength increases.
The robustness of Proto-LeakNet can be attributed to its reliance on latent-domain signal-leak cues, which remain informative even when pixel-level textures and high-frequency artifacts are substantially altered. Figure~\ref{fig:plot_accuracies} further illustrates the evolution of Top-1 accuracy across degradation levels. Latent-based approaches, including OCC-CLIP, LatentTracer, LATTE, and Proto-LeakNet, consistently outperform pixel-based methods under perturbation. In particular, Proto-LeakNet achieves 75.89\%, 74.51\%, and 69.52\% Top-1 accuracy for Steps~1, 2, and~3, respectively, compared to the second-best method (LATTE), which attains 75.64\%, 74.27\%, and 68.60\%.
Overall, these results demonstrate that Proto-LeakNet preserves attribution performance under aggressive post-processing by leveraging stable latent biases rather than relying primarily on superficial image artifacts.

\begin{center}
  \includegraphics[width=.5\columnwidth]{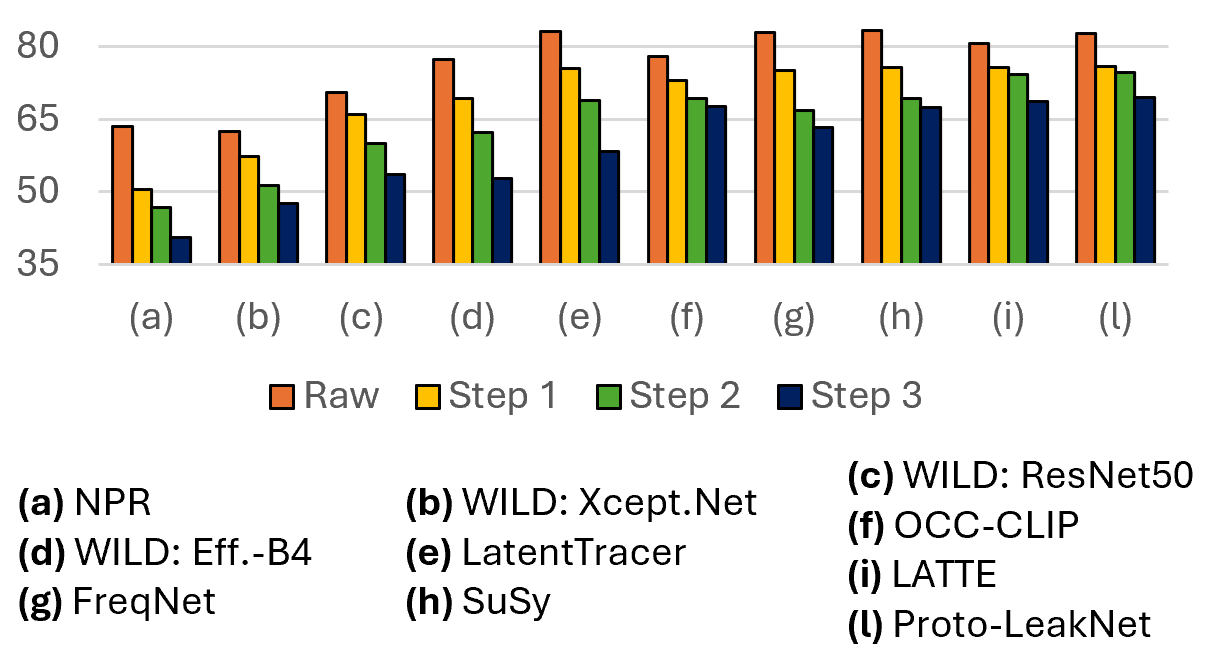}
  \captionof{figure}{Histogram of the Top-1 accuracy distributions per class for each method, from raw to step 1--3.}
  \label{fig:plot_accuracies}
\end{center}

\subsection{Extended Closed-set with Real Images}
To further assess the geometric stability of the learned embedding space, we extend the Closed-set attribution protocol by introducing an additional \emph{Real} class. 
Specifically, we augment the original 10-generator setting with 1,000 authentic images randomly sampled from FFHQ and ImageNet, resulting in an 11-class attribution task. 
The model is trained under the same conditions as the standard Closed-set configuration, without architectural modifications.
Table~\ref{tab:all_aucs_real} reports the per-class AUC values, including the newly introduced Real class, together with the overall Macro~AUC. 
Proto-LeakNet achieves a Macro~AUC of 97.35\%, demonstrating that the inclusion of authentic imagery does not degrade generator-specific discrimination. 
Importantly, the Real class is not erroneously absorbed into synthetic clusters, indicating that the learned representations preserve a clear structural distinction between authentic and generated content while maintaining inter-generator separability.
Figure~\ref{fig:tsne_closed_real} visualizes the embedding space. 
Distinct and compact clusters are observed for each generator, with the Real class forming a coherent and well-separated region. 
This behavior confirms that Proto-LeakNet does not implicitly rely on a binary real-vs-synthetic shortcut, but instead learns structured generator-specific signal-leak cues that remain geometrically organized even when authentic samples are explicitly modeled.
Overall, this extended Closed-set experiment provides additional evidence that the latent manifold learned by Proto-LeakNet is both discriminative and structurally stable, supporting reliable attribution without conflating authentic imagery with synthetic sources.

\begin{table}[t!]
\centering
\caption{AUC (\%) results for each generator class with the new Real class on the raw Closed-set. The table reports both per-class AUC and overall Macro AUC.}
\label{tab:all_aucs_real}
\begin{adjustbox}{width=1\linewidth}
\begin{tabular}{clccccccccccc|
>{\columncolor[HTML]{EFEFEF}}c }
\cline{3-13}
\textbf{}                         &                        & \multicolumn{11}{c|}{\textbf{Closed-set Classes}}                                                                                                                                                                                                                                                     & \cellcolor[HTML]{EFEFEF}                                     \\ \cline{1-12}
\textbf{Type} & \textbf{Methodologies} & \textbf{Adobe Firefly} & \textbf{Dall-E 3} & \textbf{Flux.1} & \textbf{Flux.1.1 Pro} & \textbf{Freepik} & \textbf{Leonardo AI} & \textbf{Midjourney} & \textbf{\begin{tabular}[c]{@{}c@{}}Stable \\ Diffusion 3.5\end{tabular}} & \textbf{\begin{tabular}[c]{@{}c@{}}Stable \\ Diffusion XL\end{tabular}} & \textbf{Starry AI} & \textbf{Real} & \multirow{-2}{*}{\cellcolor[HTML]{EFEFEF}\textbf{Macro AUC}} \\ \hline
\multirow{-1}{*}{\textbf{RAW}}   & \textbf{Proto-LeakNet}          & 96.58       & 97.05  & 99.82  & 99.89  & 99.81 & 96.86  & 91.94  & 95.84  & 97.50  & 96.87 & 98.66 & 97.35                                       \\ \hline
\end{tabular}
\end{adjustbox}
\end{table}

\begin{center}
  \includegraphics[width=.5\columnwidth]{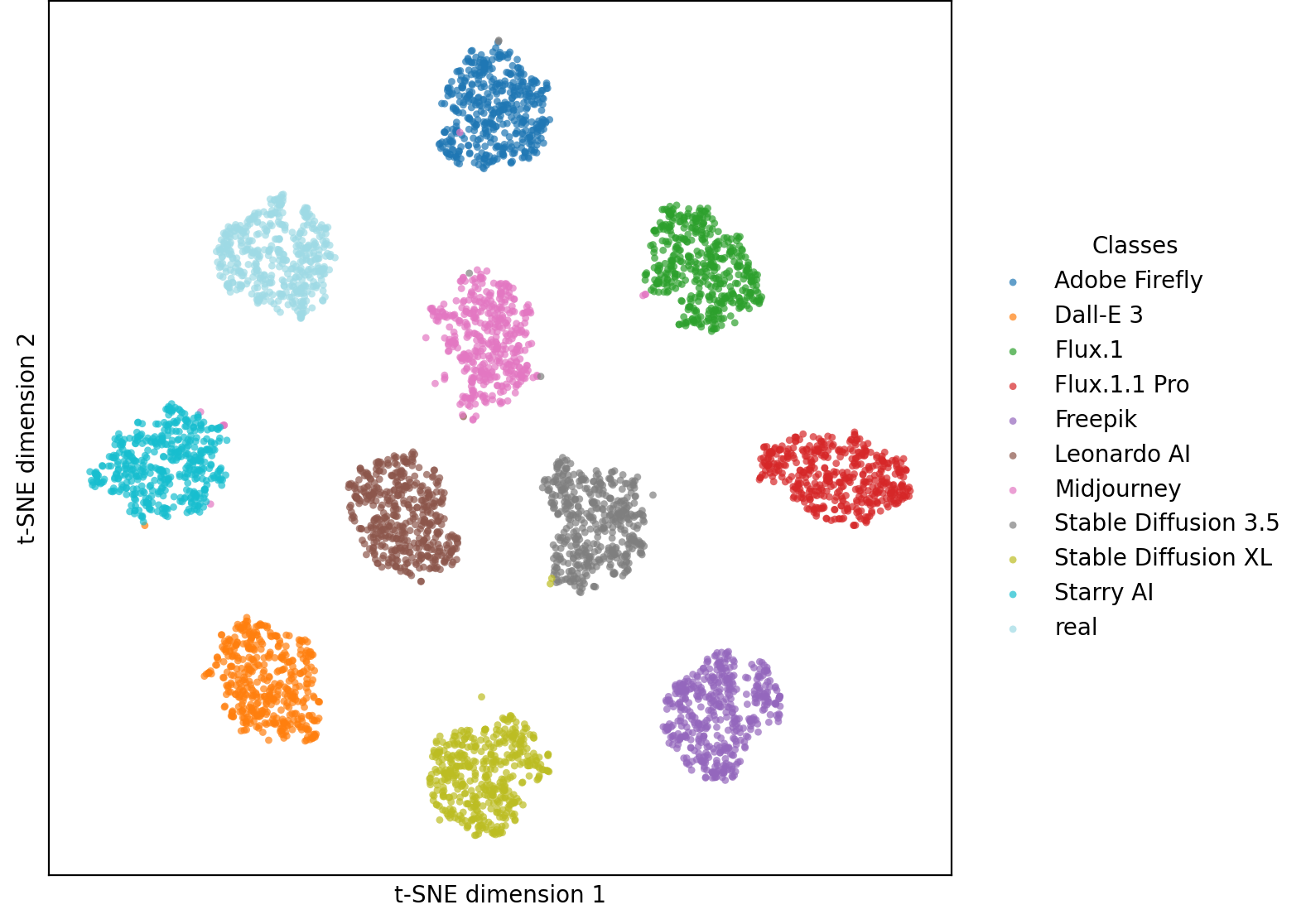}
  \captionof{figure}{Plot that shows how class-representative clusters are organized in the latent space showing a clear separation with few misclassifications.}
  \label{fig:tsne_closed_real}
\end{center}

\subsection{Feature Compression Strategies and Results}
\label{sec:compression}
To evaluate whether Proto-LeakNet’s embeddings contain redundant or weakly informative directions, we apply two post-hoc compression strategies: Principal Component Analysis (PCA) and a Denoising Variational Autoencoder (DVAE).  
Compression is performed after Temporal Attention Pooling and before the Prototype-Based Attribution Head, operating only on the fixed validation embeddings without affecting backbone training.
The motivation is twofold.  
First, prior analyses of signal-leak bias indicate that generator-specific cues often lie in a compact subspace; compressing the embeddings directly tests this hypothesis.  
Second, removing noisy or unstable dimensions may act as a regularizer, potentially improving robustness under post-processing or when encountering unseen generators.
In the following subsections, we examine PCA and DVAE independently and report their impact on Closed-set, post-processed, and Open-set rejection performance.

\subsubsection{Principal Component Analysis}

As a linear and non-parametric compression method, we employ Principal Component Analysis (PCA) as a baseline method for feature compression.  
PCA performs an orthogonal transformation of temporally pooled embeddings into a lower-dimensional subspace that maximizes retained variance while eliminating linear redundancy among features.  
Unlike learned compressors, PCA does not rely on supervision or gradient optimization and serves as a purely statistical reference for evaluating the effectiveness of non-linear latent compression.
Given a matrix of backbone embeddings $H \in \mathbb{R}^{N \times D}$, where each row $h_i$ represents the temporally aggregated feature vector of one image, PCA computes an orthogonal basis that diagonalizes the empirical covariance of the data:
\begin{align}
\tilde{H} &= H - \bar{H}, \\
C &= \tfrac{1}{N-1}\tilde{H}^\top \tilde{H}, \\
C &= U \Lambda U^\top, \\
Z &= \tilde{H} U_k,
\end{align}
where $\bar{H}$ is the mean embedding, $C$ the covariance matrix, $U_k$ the first $k$ eigenvectors associated with the largest eigenvalues in $\Lambda$, and $Z \in \mathbb{R}^{N \times k}$ the resulting compressed representations.

\subsubsection*{Dimensionality Sweep}
To analyze the trade-off between compression strength and downstream discriminative performance, we conduct a dimensionality sweep by varying the number of retained components $k$.  
Specifically, $k$ is selected from a logarithmic grid:
\[
k \in \{8, 10, 12, 16, 32, 64\}.
\]
For each configuration, the compressed embeddings are evaluated using the same attribution metrics employed throughout the study.  
This analysis exposes how the fraction of retained variance correlates with classification and Open-set recognition performance. In our experiments the sweep found its better results within 10 components.

\subsubsection*{Inference Usage}
During inference, the PCA projection matrix learned on the training embeddings is applied to new data without any further adaptation:
\begin{align}
Z_{\text{test}} = (H_{\text{test}} - \bar{H}) U_k,
\end{align}
yielding compact representations of fixed dimensionality $k$ that can be directly used by downstream attribution or classification modules.

\subsubsection{DVAE: Denoising Variational Autoencoder}
\label{sec:dvae_supp}

To explore non-linear and learnable feature compression, we employ a Denoising Variational Autoencoder (DVAE).  
Unlike PCA, which performs a fixed linear projection, the DVAE learns a stochastic mapping that compresses embeddings into a latent representation while explicitly modeling non-linear dependencies.  
By training under denoising perturbations, the DVAE enhances robustness to high-frequency distortions and residual stochasticity arising from diffusion processes or reconstruction artifacts.

Given a set of temporally pooled embeddings $H = \{h_i\}_{i=1}^N$, each $h_i \in \mathbb{R}^D$, the DVAE learns an encoder–decoder pair parameterizing the approximate posterior distribution:
\begin{align}
(\mu, \log\sigma^2) &= \mathrm{Enc}(h), \\
z &= \mu + \sigma \odot \varepsilon, \quad \varepsilon \sim \mathcal{N}(0, I), \\
\hat{h} &= \mathrm{Dec}(z),
\end{align}
where $\mu$ and $\sigma$ denote the mean and standard deviation of the approximate posterior $q(z|h)$, and $\hat{h}$ represents the reconstructed embedding.

\subsubsection*{Denoising Corruption}
Each embedding $h$ is first corrupted through a stochastic perturbation operator $g(\cdot)$ before encoding:
\begin{align}
\tilde{h} = g(h) = h + \eta - m \odot h + \mathcal{F}^{-1}\!\big(\Delta\!\operatorname{Mag} + j\,\Delta\!\operatorname{Phase}\big),
\end{align}
where $\eta \sim \mathcal{N}(0, \sigma^2 I)$ is additive Gaussian noise with variance $\sigma^2$, $m$ is a sparse Bernoulli mask and $\odot$ denotes element-wise multiplication (random feature dropout), and $\mathcal{F}(\cdot)$ denotes the discrete Fourier transform with $\mathcal{F}^{-1}(\cdot)$ its inverse.  
The terms $\Delta\!\operatorname{Mag}$ and $\Delta\!\operatorname{Phase}$ are small random perturbations applied in the frequency domain to the magnitude and phase components, respectively: if $\mathcal{F}(h)=\operatorname{Mag}(h)\,e^{j\,\operatorname{Phase}(h)}$, then the corruption modifies them as $\operatorname{Mag}(h)\!\leftarrow\!\operatorname{Mag}(h)+\Delta\!\operatorname{Mag}$ and $\operatorname{Phase}(h)\!\leftarrow\!\operatorname{Phase}(h)+\Delta\!\operatorname{Phase}$, with $j$ denoting the imaginary unit.  
This denoising process encourages the model to reconstruct clean embeddings from corrupted inputs, improving robustness to minor perturbations in the feature space.

\subsubsection*{Objective Function}
The DVAE is trained to minimize a composite loss combining reconstruction fidelity, latent regularization, and discriminative structure preservation:
\begin{equation}
\begin{split}
    \mathcal{L} 
    &= \lambda_{\text{rec}} \mathcal{L}_{\text{rec}}
     + \lambda_{\text{kl}} \mathcal{L}_{\text{kl-cap}}
\\
    &\quad + \lambda_{\text{tcons}} \mathcal{L}_{\text{tcons}}
     + \lambda_{\text{ctr}} \mathcal{L}_{\text{InfoNCE}}
     + \lambda_{\text{spec}} \mathcal{L}_{\text{spec}}.
\end{split}
\label{eq:dvae_total}
\end{equation}
where each $\lambda_{\{\cdot\}}$ term is a scalar weighting coefficient balancing the relative contribution of the corresponding loss component.
The components of the total objective are defined as follows:

\begin{itemize}
    \item \textbf{Reconstruction loss:}
    \begin{align}
        \mathcal{L}_{\text{rec}} = \| \hat{h} - h \|_1,
    \end{align}
    enforcing accurate recovery of the original embeddings $h \in \mathbb{R}^{D}$ from their reconstructions $\hat{h}$.  
    Here, $\|\cdot\|_1$ denotes the element-wise $L_1$ norm, favoring sharper reconstructions and reducing blurring effects compared to an $L_2$ objective.

    \item \textbf{KL divergence with capacity control:}
    \begin{align}
        \mathrm{KL}(q\|p) &= -\tfrac{1}{2} \sum_{j=1}^{d_z} (1 + \log\sigma_j^2 - \mu_j^2 - \sigma_j^2), \\
        \mathcal{L}_{\text{kl-cap}} &= \big|\, \mathrm{KL}(q\|p) - c(\tau) \,\big|.
    \end{align}
    This term regularizes the approximate posterior $q(z|h)=\mathcal{N}(\mu,\mathrm{diag}(\sigma^2))$ against the isotropic Gaussian prior $p(z)=\mathcal{N}(0,I)$.  
    The vector $\mu \in \mathbb{R}^{d_z}$ represents the latent mean, $\sigma \in \mathbb{R}^{d_z}$ its standard deviation, and $d_z$ the latent dimensionality.  
    The function $c(\tau)$ defines a target \emph{capacity} that increases linearly with the normalized training step $\tau \in [0,1]$, progressively allowing larger information throughput to prevent posterior collapse.

    \item \textbf{Intra-class consistency:}
    \begin{align}
        \mathcal{L}_{\text{tcons}} =
        \sum_c \sum_{i \in \mathcal{B}_c}
        \| \mu_i - m_c \|_2^2,
        \quad
        m_c = \tfrac{1}{|\mathcal{B}_c|} \sum_{i \in \mathcal{B}_c} \mu_i.
    \end{align}
    This loss promotes compactness of latent means $\mu_i$ belonging to the same class $c$.  
    Each $\mathcal{B}_c$ is the set of indices in the current mini-batch corresponding to class $c$, and $m_c$ denotes the latent centroid of that class.  
    The Euclidean norm $\|\cdot\|_2$ measures within-class dispersion in the latent space.

    \item \textbf{Contrastive regularization:}
    \begin{align}
        s_{ij} &= \frac{\langle \hat{\mu}_i, \hat{\mu}_j \rangle}{T}, \\
        \mathcal{L}_{\text{InfoNCE}} &=
        -\frac{1}{|\mathcal{P}|} \sum_{(i,j)\in\mathcal{P}}
        \log
        \frac{\exp(s_{ij})}{\sum_{k \neq i}\exp(s_{ik})},
    \end{align}
    where $\hat{\mu}_i = \mu_i / \|\mu_i\|_2$ is the normalized latent mean, $s_{ij}$ is the cosine similarity between samples $i$ and $j$, and $T$ is a temperature parameter controlling contrastive sharpness.  
    The set $\mathcal{P}$ contains positive pairs of samples sharing the same label, while all remaining samples act as negatives.  
    This loss enforces inter-class separation and helps structure the latent space for discriminative attribution.

    \item \textbf{Spectral consistency:}
    \begin{align}
        \mathcal{L}_{\text{spec}} =
        \|\, |\mathcal{F}(\hat{h})| - |\mathcal{F}(h)| \,\|_1,
    \end{align}
    aligning the amplitude spectra of reconstructed and original embeddings in the frequency domain.  
    Here, $\mathcal{F}(\cdot)$ denotes the discrete Fourier transform, and $|\cdot|$ its magnitude.  
    This constraint stabilizes the denoising process by preserving generator-specific frequency patterns crucial to attribution.
\end{itemize}

\subsubsection*{Inference and Compression}
After training, only the encoder mean $\mu(h)$ is retained as the compressed representation:
\begin{align}
Z = \mu(H) \in \mathbb{R}^{N \times d_z}.
\end{align}
Sampling is disabled during inference.  
The resulting latent embeddings are then passed to downstream attribution modules in the same manner as PCA-compressed representations.

\subsubsection{Experimental results with PCA and DVAE}

Table~\ref{tab:pca_vs_dvae} compares Proto-LeakNet with and without compression.  
The uncompressed backbone achieves the highest overall performance, confirming that direct use of high-dimensional embeddings preserves the full spectral footprint of the signal-leak encoded in the diffusion trajectory.  
Both PCA and DVAE introduce mild performance drops, reflecting the trade-off between compactness and fidelity in the latent representation.
The PCA baseline attains strong results, showing that much of the discriminative variance lies in the first few principal components.  
This is consistent with the low intrinsic dimensionality of signal-leak features, which are dominated by low-frequency correlations and thus highly compressible through linear projections.  
Since PCA preserves global variance but disregards higher-order dependencies, its performance remains close to the uncompressed setting, demonstrating that the dominant leak directions in feature space are nearly orthogonal and linearly separable.
The DVAE achieves comparable but slightly lower accuracy. 
Although the non-linear autoencoding allows modeling of complex manifolds and denoising distortions, the learned latent space emphasizes reconstruction stability over discriminative sharpness.  
In practice, the denoising objective partially regularizes away the subtle generator-specific frequency biases that the attribution model exploits, leading to smoother but less distinctive embeddings.  
This effect mirrors the analysis in the work presented by Everaert et al.~\citep{Everaert2024SignalLeak}, where suppressing high-frequency leakage reduces variance across diffusion trajectories, improving robustness but marginally weakening attribution sensitivity.
These findings confirm that most of the exploitable signal-leak information resides in low-frequency, linearly decodable components of the backbone embedding, and that explicit non-linear denoising trades off fine-grained attribution cues for stability.

\begin{table}[!ht]
  \caption{Comparative results of Proto-LeakNet evaluation without compression and with compression via PCA or DVAE.}
  \label{tab:pca_vs_dvae}
  \centering
  \begin{tabular}{@{}lcc@{}}
    \toprule
    \textbf{Compression Type} & \textbf{Top-1 Acc. (\%)} & \textbf{Macro AUC (\%)} \\
    \midrule
    PCA & 81.30 & 97.49 \\
    DVAE & 81.77 & 96.36 \\
    \midrule
    \textbf{No Compression} & \textbf{82.60} & \textbf{98.13} \\
    \bottomrule
  \end{tabular}
\end{table}

\subsection{Latent Cluster Analysis}
\label{sec:cluster_analysis}
To provide a deeper understanding of the latent geometry learned by Proto-LeakNet and expand the cluster's quality obtained, we perform a comprehensive cluster analysis over the temporally aggregated embeddings $\bar{h}\!\in\!\mathbb{R}^{D}$ obtained after the attention pooling module and organized through prototype-based supervision.  
This analysis aims to quantify how the model organizes generator-specific features in latent space, how stable these clusters remain across training, validation, and test splits, and how such structure supports Open-set generalization via density estimation.  
By examining variance statistics, signal-to-noise ratios, and silhouette distributions, we reveal that the learned embeddings form compact, well-separated manifolds that persist under domain shifts and post-processing perturbations.

\subsubsection*{Intra- and Inter-Class Variance}
To assess the compactness and separability of latent embeddings, we first compute intra-class and inter-class variance statistics.  
The intra-class variance measures the spread of samples within each generator cluster, while the inter-class variance quantifies the dispersion of class centroids in the global latent space.  
Formally, these are defined as
\begin{equation}
    \begin{split}
    \mathrm{Var}_{\text{intra}} = \frac{1}{C}\sum_{c=1}^{C} 
    \frac{1}{N_c}\sum_{i=1}^{N_c} 
    \lVert \bar{h}_i^{(c)} - \mu_c \rVert_2^2,
    \qquad \\
    \mathrm{Var}_{\text{inter}} = 
    \frac{1}{C}\sum_{c=1}^{C} 
    \lVert \mu_c - \mu_{\text{global}} \rVert_2^2,
    \end{split}
\end{equation}
where $\bar{h}_i^{(c)}$ denotes the embedding of the $i$-th image generated by model $c$, $\mu_c$ the centroid of that generator’s embeddings, and $\mu_{\text{global}}$ the mean embedding across all generators.  
These two measures capture complementary properties: intra-class variance represents the internal cohesion of a generator cluster, while inter-class variance reflects the degree of structural separation between distinct generators.

Figure~\ref{fig:variance_inter_intra} reports the inter- and intra-class variances across the training, validation, and test splits.  
Across all splits, $\mathrm{Var}_{\text{inter}}$ dominates $\mathrm{Var}_{\text{intra}}$ by more than an order of magnitude, confirming that generator embeddings occupy distinct regions of the latent space.  
The moderate rise in intra-class variance from training to test arises from natural distributional differences between the Closed-set splits, proving the robustness and unbias claims of the dataset, while the inter-class margin remains sufficiently large to sustain discriminability.
This is consistent with the intuition that signal-leak statistics encode low-frequency latent biases that are robust to pixel-level perturbations.

\begin{center}
  \includegraphics[width=.5\linewidth]{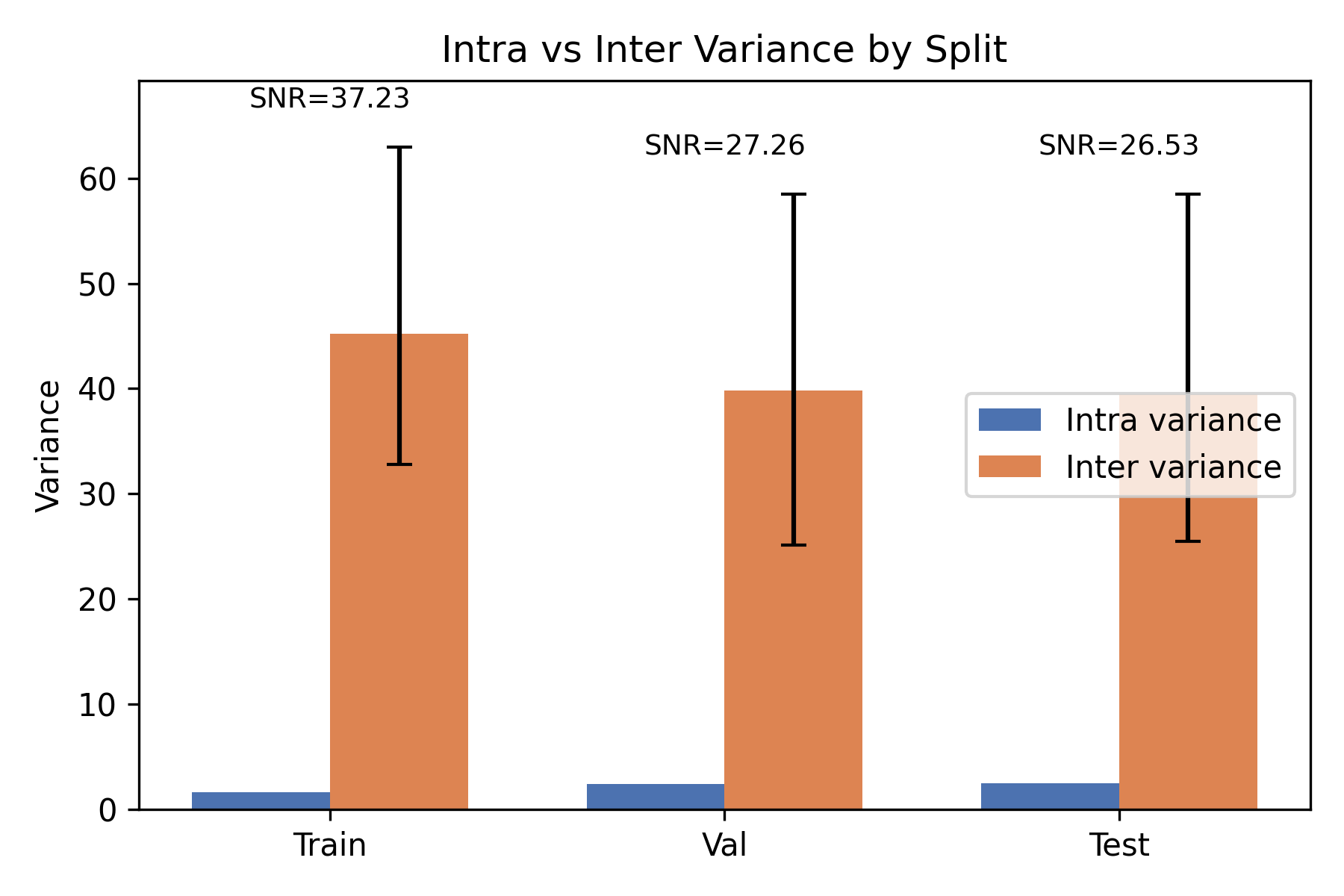}
  \captionof{figure}{Plot portraying the intra- vs. inter-class variance for each split, with orange bars representing inter-class variance and blue bars representing intra-class variance.}
  \label{fig:variance_inter_intra}
\end{center}

\subsubsection*{Signal-to-Noise Ratio (SNR)}
To summarize cluster separability into a single interpretable quantity, we compute the signal-to-noise ratio (SNR), defined as
\begin{equation}
    \mathrm{SNR} = 
    \frac{\mathrm{Var}_{\text{inter}}}{\mathrm{Var}_{\text{intra}}},
\end{equation}
which expresses how clearly class centroids stand apart from their internal variance.  
High SNR indicates compact, well-separated clusters, while low SNR implies blurred boundaries or overlapping manifolds.  
Figure~\ref{fig:snr_curve_clusters} illustrates the evolution of SNR across dataset splits.  
Proto-LeakNet achieves values of 37.2, 27.3, and 26.5 for training, validation, and test respectively, indicating that the embedding geometry remains stable and well separated across data splits.  
This confirms that the latent representations capture generator-specific cues rather than overfitting to particular prompts or samples.

\begin{center}
  \includegraphics[width=.5\linewidth]{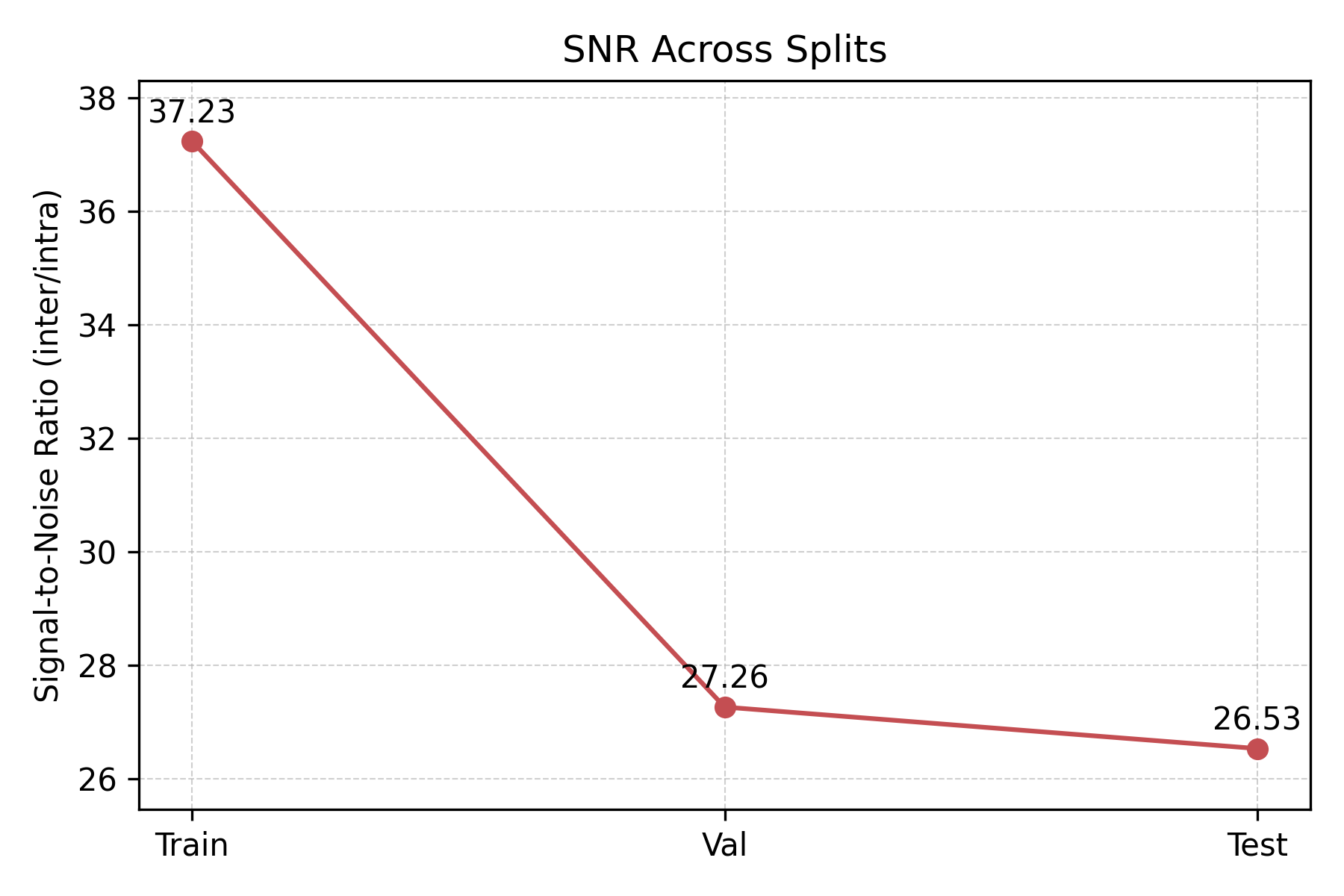}
  \captionof{figure}{Signal-to-noise ratio (SNR) behaviour across the dataset splits, showing a natural decay from training to validation and test sets.}
  \label{fig:snr_curve_clusters}
\end{center}

\subsubsection*{Per-Class Variance Distribution}
To gain finer-grained insight, we compute the per-class variance
\begin{equation}
    v_c = \frac{1}{N_c}\sum_{i=1}^{N_c}
    \lVert \bar{h}_i^{(c)} - \mu_c \rVert_2^2,
\end{equation}
where $v_c$ denotes the intra-class variance for generator $c$, $N_c$ is the number of embeddings associated with that generator, $\bar{h}_i^{(c)}$ is the latent embedding of the $i$-th sample belonging to class $c$, and $\mu_c$ represents the centroid (mean embedding) of all samples for that generator.  
The operator $\lVert \cdot \rVert_2^2$ corresponds to the squared Euclidean norm, which measures the dispersion of each embedding with respect to its class mean.
The per-class variance heatmap in Figure~\ref{fig:per_class_variance_heatmap} visualizes these values across dataset splits.  
Darker cells represent lower variance, indicating highly cohesive clusters, while lighter cells correspond to more dispersed embeddings.  

Interestingly, generators such as Flux.1.1 Pro, Freepik, and Flux.1 consistently show darker tones, meaning their latent representations are compact and internally coherent.  
This aligns with their visual consistency and constrained generation pipelines.  
Conversely, models like Midjourney and DALL\-E~3 exhibit brighter regions, reflecting greater stylistic diversity and broader latent manifolds.  
These findings highlight that generator design and prompt distribution directly influence the latent geometry: closed-source models optimized for stylistic variety tend to produce more dispersed embeddings.

\begin{center}
  \includegraphics[width=.5\linewidth]{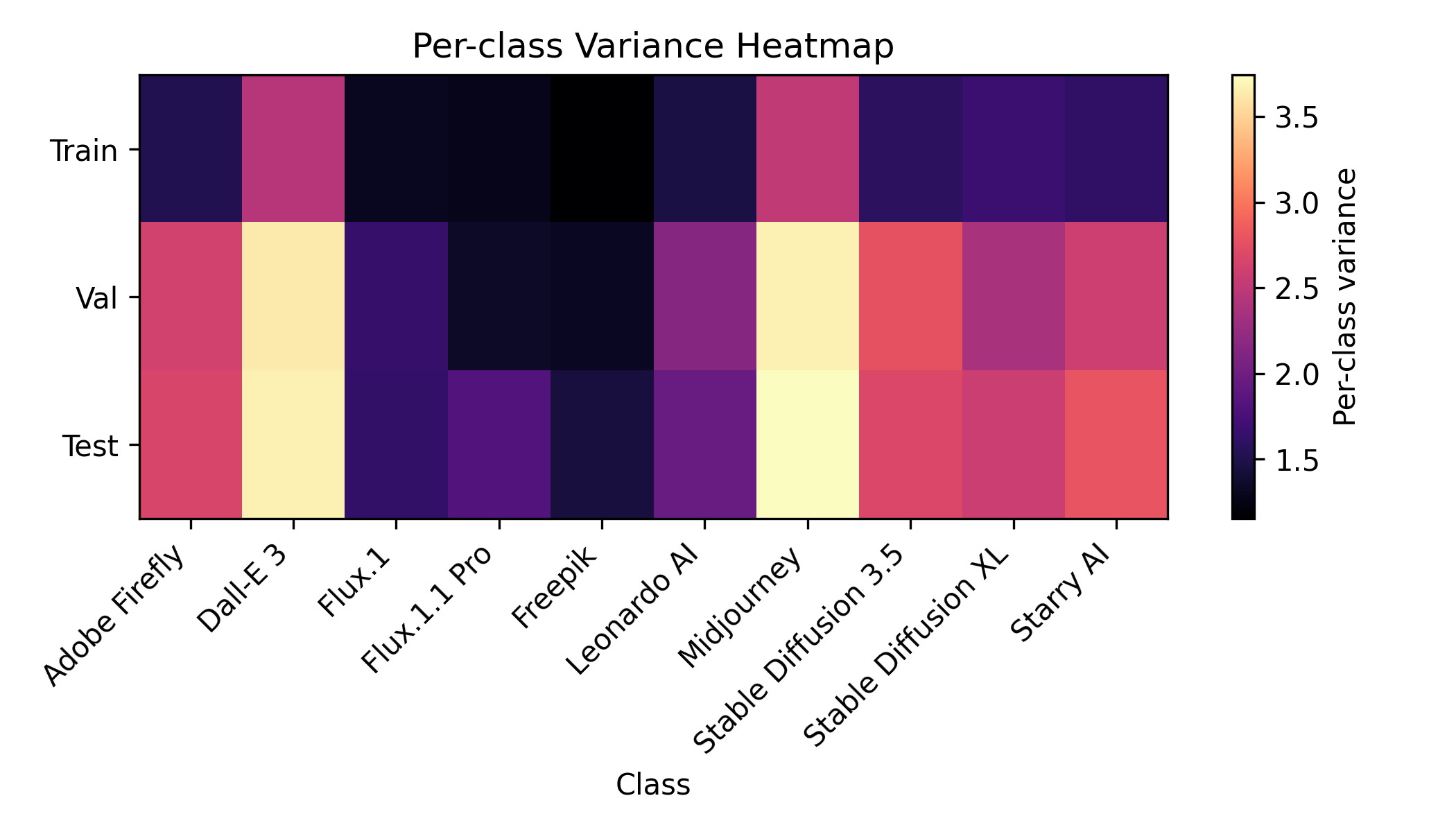}
  \captionof{figure}{Heatmap illustrating the per-class variance distribution, where darker colors represent more cohesive clusters and lighter colors indicate less compact clusters.}
  \label{fig:per_class_variance_heatmap}
\end{center}

\subsubsection*{Silhouette Coefficient Analysis}
Cluster separability is further quantified using the silhouette coefficient~$s_i$ for each sample:
\begin{equation}
    s_i = \frac{b_i - a_i}{\max(a_i, b_i)},
\end{equation}
where $a_i$ is the mean distance between sample~$i$ and all other samples of the same class (intra-class cohesion), and $b_i$ is the minimum mean distance to samples of the nearest neighboring class (inter-class separation).  
The coefficient ranges from $-1$ (misclassified or overlapping) to $+1$ (perfectly clustered).

Figure~\ref{fig:silhouette_by_class} shows the silhouette score distributions for each generator, while Figure~\ref{fig:silhouette_means} reports the per-class means.  
Most distributions are skewed toward positive values, confirming the presence of dense, well-separated clusters.  
Flux.1.1 Pro and Freepik achieve the highest average silhouettes ($s_{\text{mean}}>0.5$), consistent with their low variance and compact geometry.  
On the other hand, Midjourney and DALL\-E~3 exhibit broader, flatter distributions, indicating that their embeddings partially overlap with neighboring clusters.  
This again corresponds to their high stylistic variability and weaker latent regularity.  
Overall, the silhouette analysis confirms that Proto-LeakNet organizes latent embeddings into generator-specific manifolds that are both cohesive and interpretable.

\begin{center}
  \includegraphics[width=1\linewidth]{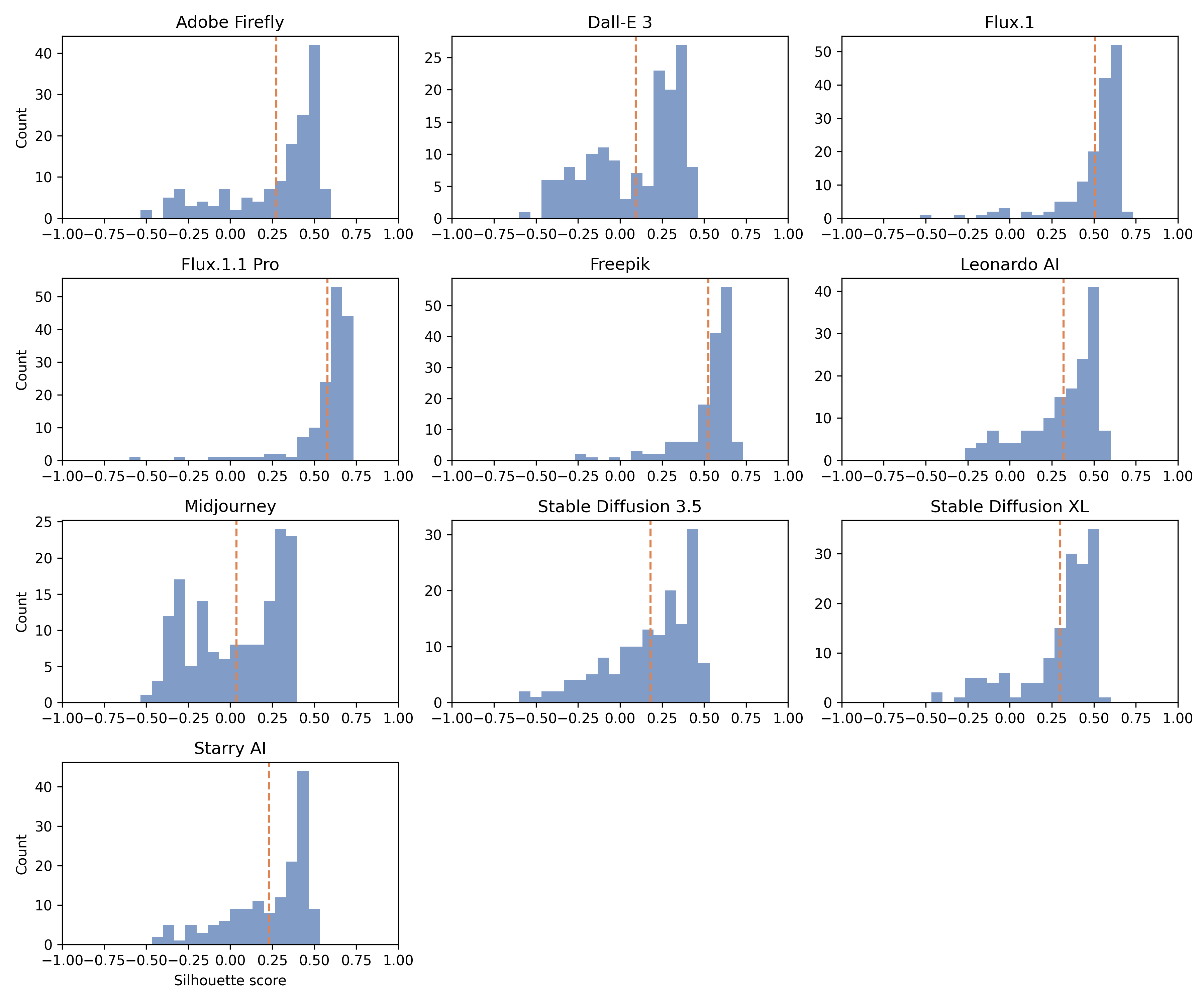}
  \captionof{figure}{Silhouette score distributions for each class. Positive values indicate well-separated and cohesive clusters, whereas negative values suggest poor clusterization and overlapping regions.}
  \label{fig:silhouette_by_class}
\end{center}

\begin{center}
  \includegraphics[width=.5\linewidth]{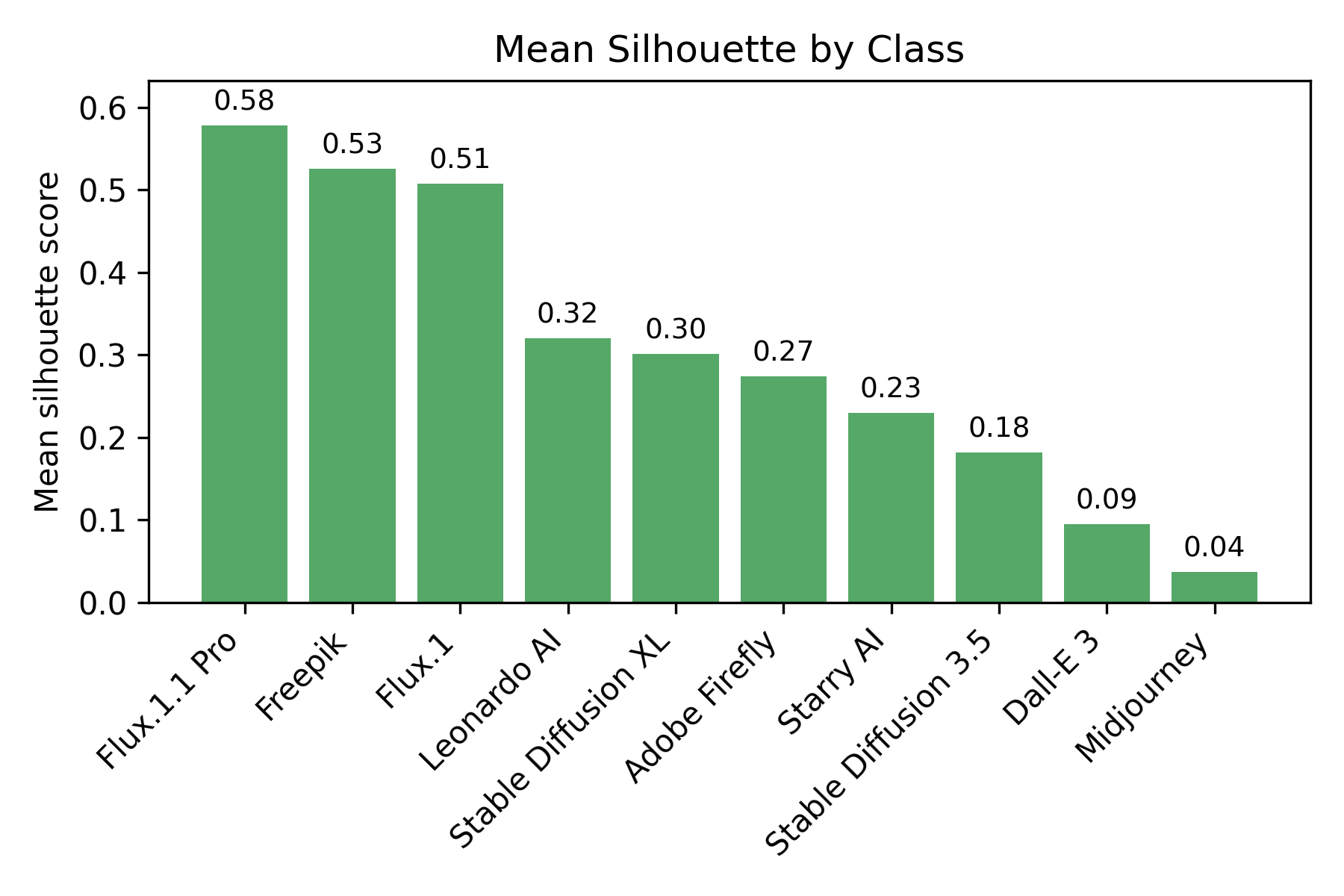}
  \captionof{figure}{Histogram of the per-class mean silhouette scores.}
  \label{fig:silhouette_means}
\end{center}

\subsubsection*{Relation to Density-Based Open-set Evaluation}
The clustering properties described above directly explain the strong Open-set behavior observed before. 
Because the Closed-set clusters are compact and non-overlapping (high SNR, positive silhouettes), their density estimates exhibit clear low-density boundaries.  
As a result, open samples from unseen generators fall into low-probability regions of $p_{\text{KDE}}$, naturally yielding low log-likelihood scores $s(h) = \log p_{\text{KDE}}(h)$ without any need for explicit open-class supervision.
Empirically, this manifests as consistent separability between closed and open distributions, as shown by high Macro~AUC and low Overlap Coefficient (OVL) values.  
The KDE-based scoring thus acts as an unsupervised detector of out-of-distribution generators, relying entirely on the structured geometry produced by Proto-LeakNet.  
This finding demonstrates that the model not only learns discriminative prototypes but also forms a representation space with meaningful statistical boundaries, where unseen generators can be rejected by density alone.

\subsubsection*{Discussion}
The combined variance, SNR, and silhouette analyses provide a quantitative confirmation of Proto-LeakNet’s ability to capture generator-specific biases as structured, low-dimensional manifolds in the latent domain.  
Across all metrics, the latent geometry remains remarkably stable: intra-class variance stays low, inter-class separation high, and silhouette scores predominantly positive across splits.  
The moderate SNR drop from train to test indicates realistic domain adaptation without loss of discriminative power.  
Generators that are visually consistent, or governed by constrained architectural priors, yield compact and highly cohesive clusters; those optimized for diversity exhibit broader latent spread but remain separable.
Crucially, the strong correlation between cluster compactness and Open-set KDE behavior confirms that Proto-LeakNet’s representation-level generalization stems from intrinsic geometric structure rather than explicit supervision.  
The encoder’s latent manifolds exhibit statistically meaningful density gaps that correspond to unseen generator domains, effectively turning signal-leak bias into a measurable and discriminative fingerprint.  
This latent geometry not only enables reliable attribution under closed conditions but also provides a principled foundation for unsupervised Open-set detection.

\color{black}
\subsection{Open-set Rejection and Representation-Level Generalization}
\label{sec:6.6}
This section evaluates whether the representation learned by Proto-LeakNet can support Open-set rejection. 
The goal is not to attribute an unseen sample to its true unknown generator identity. 
Instead, we test whether samples produced by unseen generators, or authentic real images, can be distinguished from samples produced by the known generators used during training. 
Therefore, this experiment can be interpreted as a known-versus-unknown detector operating on the learned embedding space.
After Closed-set training, the ResNet18 encoder is frozen and used to extract embeddings for all samples. 
No retraining, fine-tuning, or adaptation is performed. 
A kernel density estimator (KDE) is fitted only on embeddings from the Closed-set generators. 
This KDE models the density of the known-generator manifold. 
At test time, embeddings from unseen generators or real images are projected into the same frozen space and scored according to their likelihood under this known-generator density. 
High likelihood indicates compatibility with the known-generator manifold, while low likelihood indicates that the sample lies outside this manifold and should be rejected as unknown.
We then use this Open-set rejection protocol to analyze how temporal attention affects the geometry of the learned representation. 
Specifically, we compare three configurations:

\begin{itemize}
    \item \textbf{No Attention:} temporal attention is disabled, and embeddings are obtained by simple aggregation of diffusion-step features.

    \item \textbf{Attention on Both Domains:} temporal attention is applied uniformly to both closed-set samples used to fit the KDE and Open-set samples evaluated at test time. 
    This corresponds to the standard inference setting in which all inputs are processed in the same way.

    \item \textbf{Attention on Closed Only:} temporal attention is applied only when constructing the Closed-set KDE reference manifold, while Open-set samples are embedded without attention. 
    This configuration is not intended as a deployable inference protocol, because at runtime the model does not know whether an input is known or unknown. 
    Instead, it is a diagnostic setting designed to isolate how attention shapes the known-generator manifold used for density estimation.
\end{itemize}
Figure~\ref{fig:modes}(a--c) and Table~\ref{tab:open_attention_modes} summarize the results. 
Without attention, the score distributions of known and unknown samples partially overlap, reducing Open-set separability. 
When attention is applied to both Closed-set and Open-set samples, unseen samples are projected closer to the known-generator manifold, which increases overlap and weakens rejection performance. 
In contrast, when attention is used only to construct the Closed-set KDE reference manifold, the known generators form a more compact density region, while unseen generators remain in low-density areas. 
This yields stronger separation between known and unknown samples.
The asymmetric ``Attention on Closed Only'' setting should therefore be interpreted as a representation-level diagnostic, not as the operational Open-set protocol used at deployment. 
It estimates how well the learned attention mechanism can organize the known-generator manifold when used to define the KDE reference density. 
In practice, a deployable system should process all incoming samples uniformly, as in the ``Attention on Both Domains'' configuration, unless an additional unsupervised mechanism is introduced to decide when attention should be selectively applied. 
We leave this selective-attention strategy as future work.

\subsubsection{Further Studies on Open-set Rejection}
To further evaluate whether the learned representation generalizes beyond synthetic generators, we apply the same KDE-based rejection protocol to authentic real-image datasets: CelebA-HQ~\citep{liu2015faceattributes,karras2018progressive}, FFHQ~\citep{karras2018ffhq}, and ImageNet~\citep{deng2009imagenet}. 
These real images are treated strictly as Open-set samples and are never observed during training, density fitting, or calibration. 
Their embeddings are projected into the frozen Proto-LeakNet space and scored against the KDE fitted on Closed-set synthetic generators.

Across the evaluated real-image datasets, real-image embeddings fall outside the high-density regions associated with the known synthetic generators. 
Figures~\ref{fig:modes}(d),~\ref{fig:imagenet_kde}, and~\ref{fig:ffhq_kde} illustrate this separation, while the quantitative results show negligible overlap between authentic imagery and the known synthetic-generator manifold.

\color{black}
Furthermore, we extend our evaluation under the same settings to three datasets generated by out-of-distribution diffusion models: Stable Diffusion 1.5, Stable Diffusion 2.1~\cite{sd-faces}, and Z-Turbo~\citep{z-image-2025}. Figure~\ref{fig:kde_open} 
demonstrate that Proto-LeakNet perfectly disentangles these unseen images without any overlap, highlighting the strong generalization and robustness of its learned embeddings.
It is worth emphasizing an interesting aspect related to the additional Stable Diffusion variants included in the Open-set evaluation. The WILD Closed-set split already contains diffusion-based generators such as Stable Diffusion XL Turbo and Stable Diffusion 3.5 Large. Consequently, evaluating unseen models such as Stable Diffusion 1.5 and Stable Diffusion 2.1 provides a more challenging Open-set setting, since these generators share similar diffusion-based generation principles. Despite these similarities, Proto-LeakNet consistently rejects them as unseen generators. This behaviour suggests that the learned representation is not simply capturing generic diffusion-related characteristics, but is instead able to model more specific signal-leak patterns associated with different generation pipelines.
\color{black}

Finally, we apply the same analysis to the Partial Manipulation Dataset described in Section~\ref{sec:dataset}, excluding the FFHQ real class and treating the manipulated images as Open-set samples. 
As shown in Figure~\ref{fig:step1_kde}, the score distributions remain separated, indicating that Proto-LeakNet can also distinguish known full-image synthetic generators from localized manipulation sources not observed during training.

Overall, this section shows that Proto-LeakNet learns a representation that supports known-versus-unknown rejection in addition to Closed-set attribution. 
The Closed-set classifier attributes images to known generators, while the KDE-based analysis evaluates whether samples from unseen generators or real images fall outside the learned known-generator manifold. 
This distinction clarifies that the Open-set component is a rejection mechanism, not an attribution mechanism for unseen generator identities.
\color{black}

\begin{center}
  \includegraphics[width=.7\columnwidth]{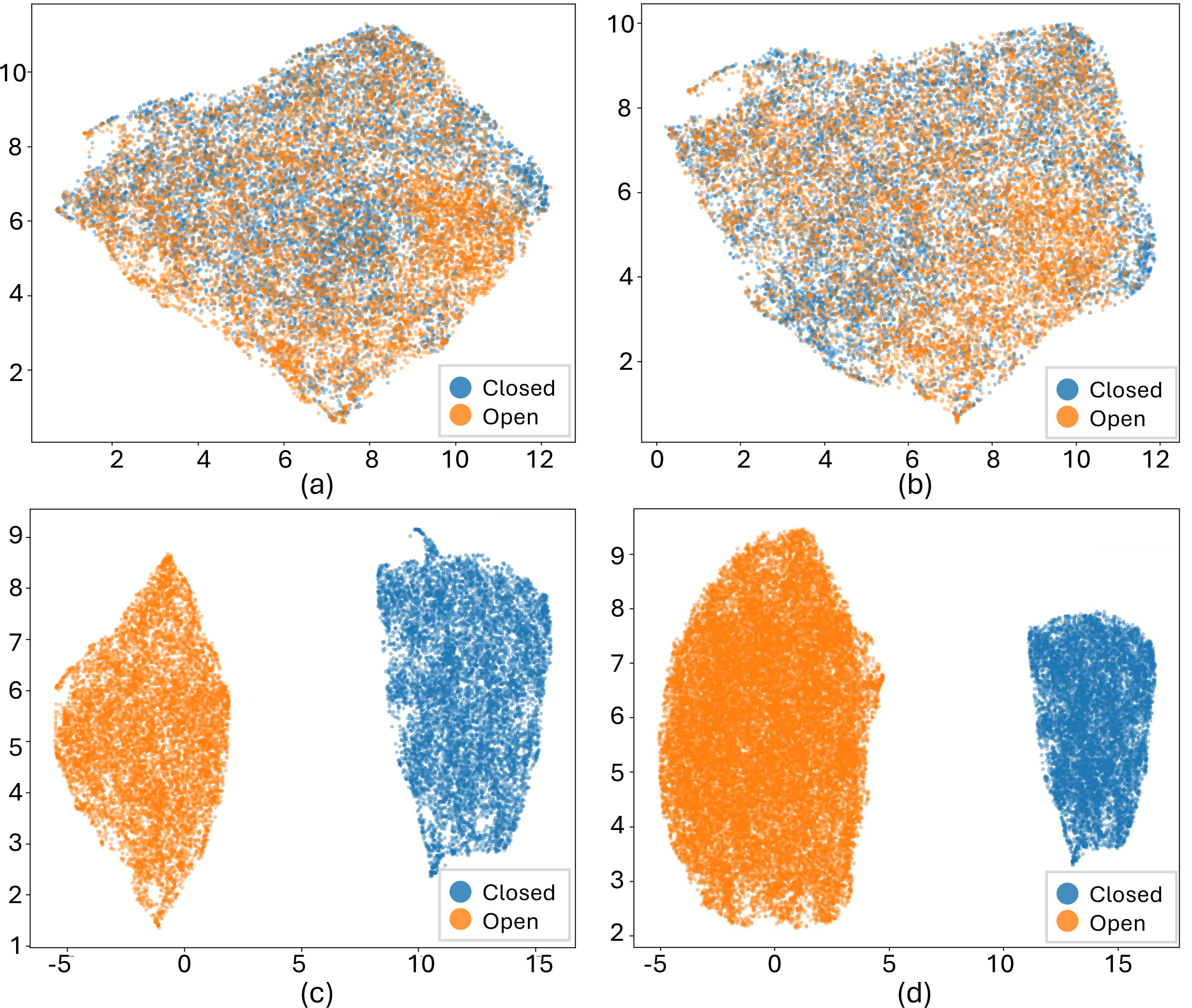}
  \captionof{figure}{Impact of attention configurations on latent-space separation.
  (a) Disabling attention for both domains leads to overlapping closed and open clusters.
  (b) Enabling attention for both aligns open embeddings with the closed distribution.
  (c) Attention on Closed Only, produces well-separated clusters.
  (d) Applying the same setup as (c) with real CelebA-HQ images, separability is preserved.}
  \label{fig:modes}
\end{center}

\begin{table}[t!]
\centering
\caption{Open-set evaluation under different attention configurations.  
  We report AUROC, Equal Error Rate (EER), Overlap Coefficient (OVL) and FPR@95.  
  Lower EER and OVL indicate better separation between closed and open domains.}
  \label{tab:open_attention_modes}
\begin{tabular}{ccccc}
\hline
\textbf{Configuration} & \textbf{AUC (\%)} & \textbf{EER}  & \textbf{OVL} & \textbf{FPR@95} \\ \hline
Attention Off               & 57.24             & 0.44          & 0.89    & 0.91      \\
Attention On                & 56.62             & 0.45          & 0.90    & 0.92       \\
Attention on Closed Only                   & \textbf{100.00}   & \textbf{0.00} & \textbf{0.00} & \textbf{0.00}\\ 
\hline
\end{tabular}
\end{table}

\begin{center}
  \includegraphics[width=.5\columnwidth]{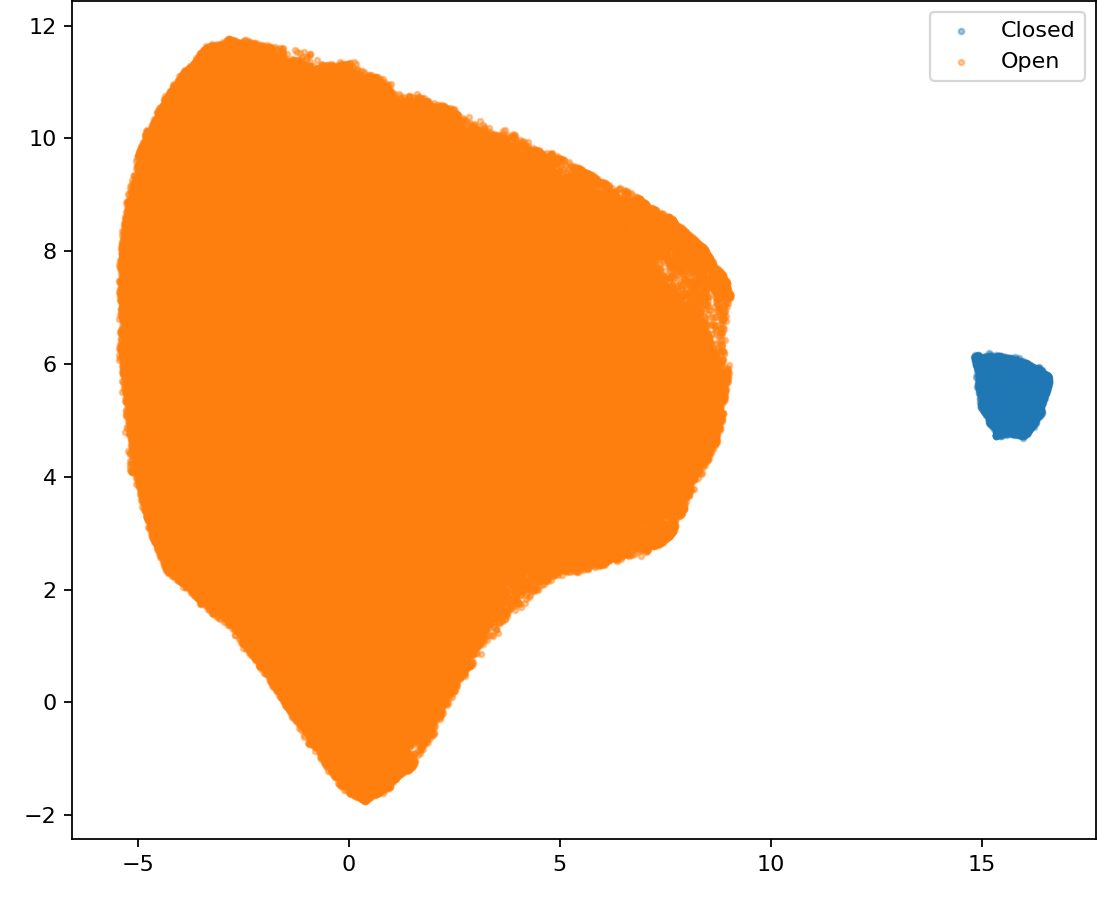}
  \captionof{figure}{Attention on Closed Only, produces well-separated clusters with real ImageNet images. AUC: 100.0\%, EER: 0.00 and OVL: 0.00}
  \label{fig:imagenet_kde}
\end{center}

\begin{center}
  \includegraphics[width=.5\columnwidth]{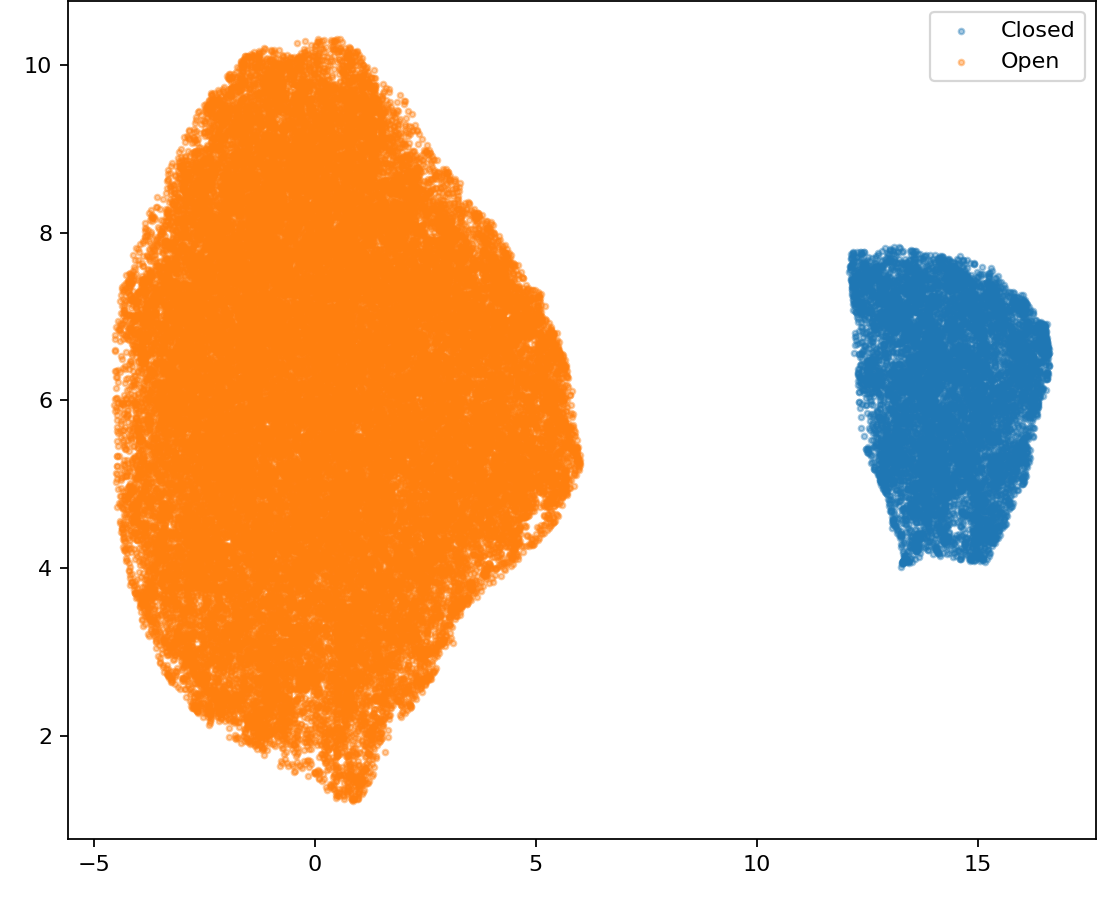}
  \captionof{figure}{Attention on Closed Only, produces well-separated clusters with real FFHQ images. AUC: 100.0\%, EER: 0.00 and OVL: 0.00}
  \label{fig:ffhq_kde}
\end{center}

\begin{center}
  \includegraphics[width=.5\columnwidth]{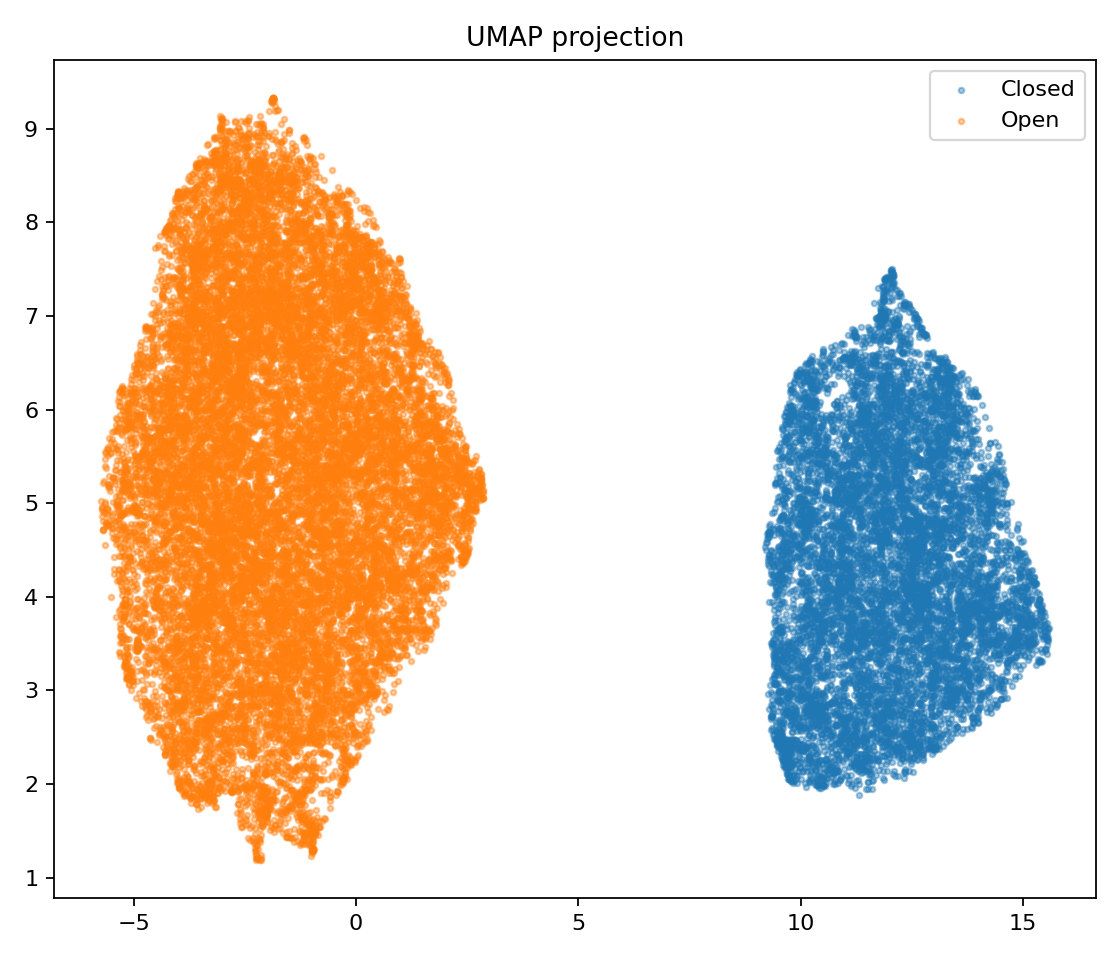}
  \captionof{figure}{Attention on Closed Only, produces well-separated clusters with Partial manipulation dataset. AUC: 100.0\%, EER: 0.00 and OVL: 0.00}
  \label{fig:step1_kde}
\end{center}




\begin{center}
\begin{tabular}{ccc}
\includegraphics[width=0.30\textwidth]{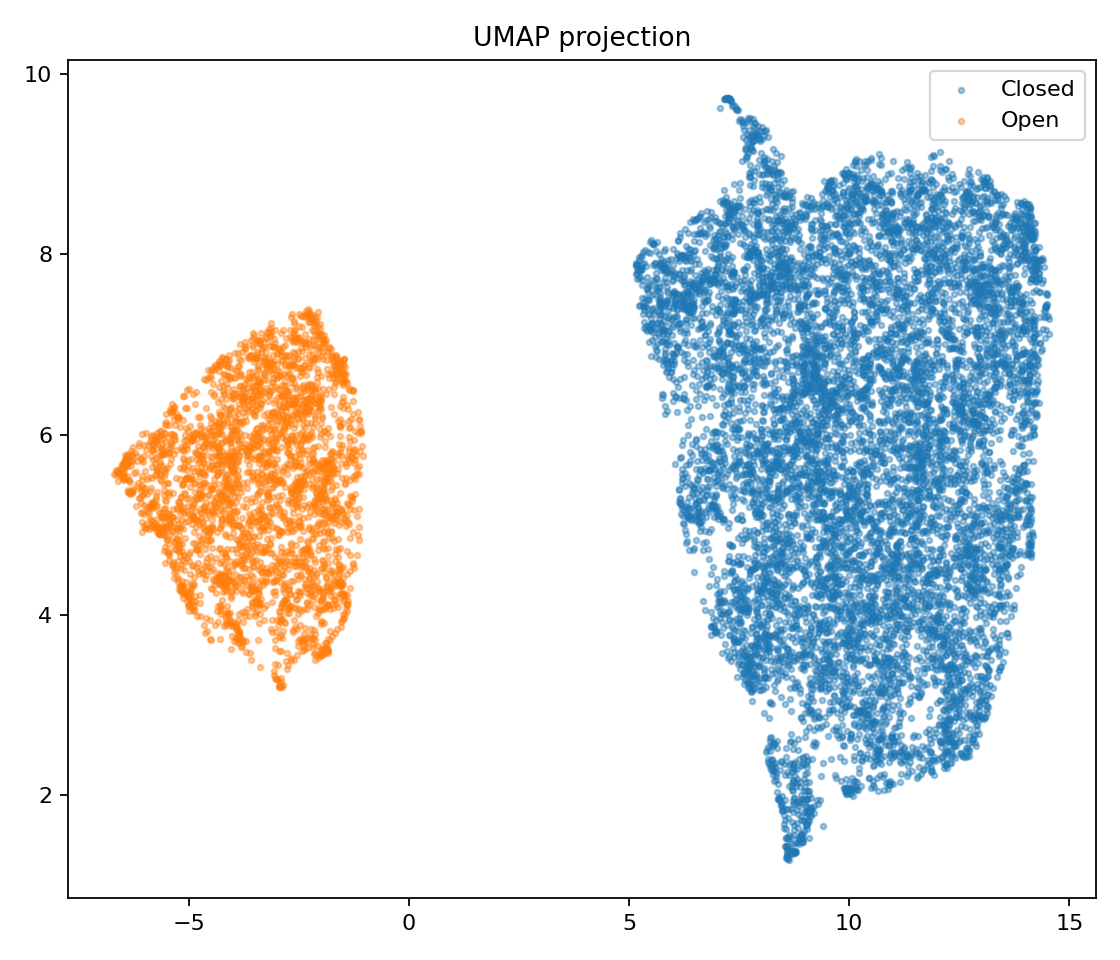} &
\includegraphics[width=0.30\textwidth]{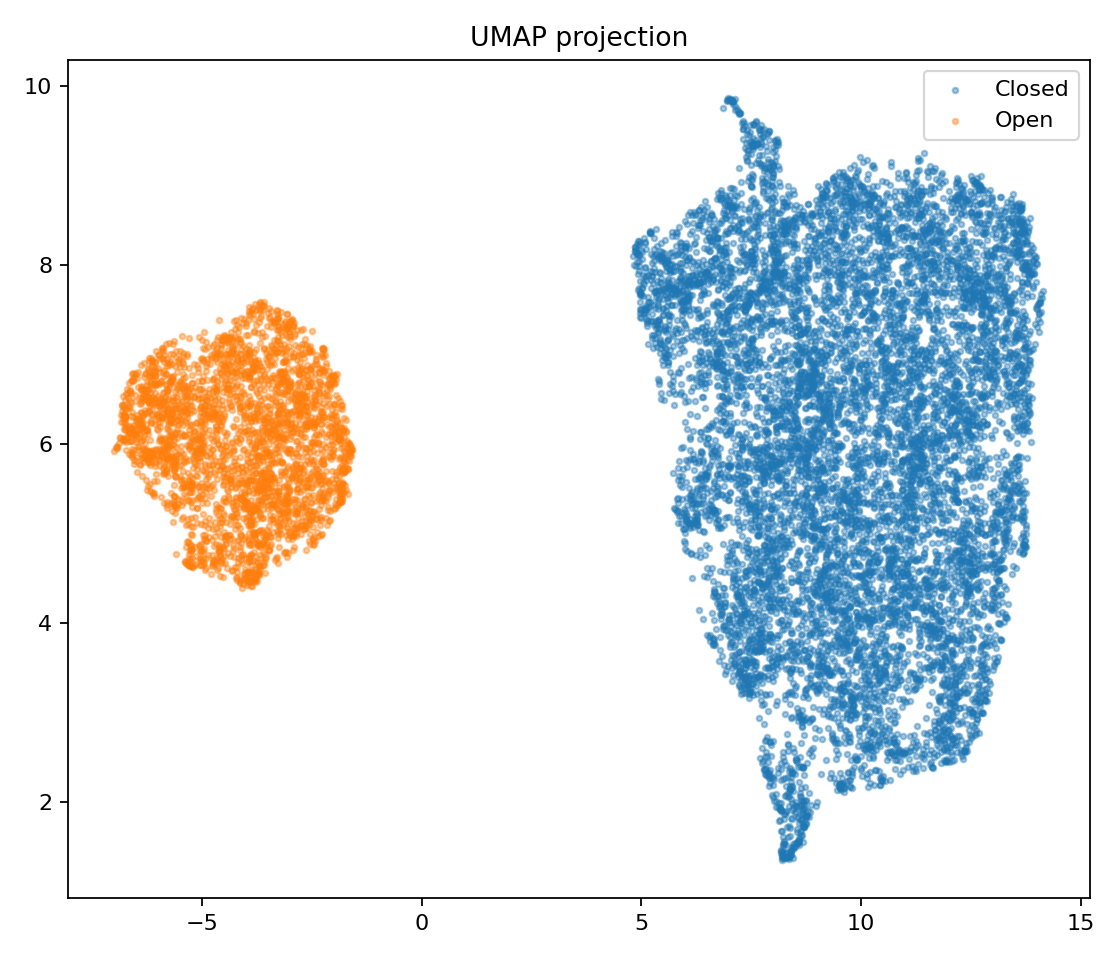} &
\includegraphics[width=0.30\textwidth]{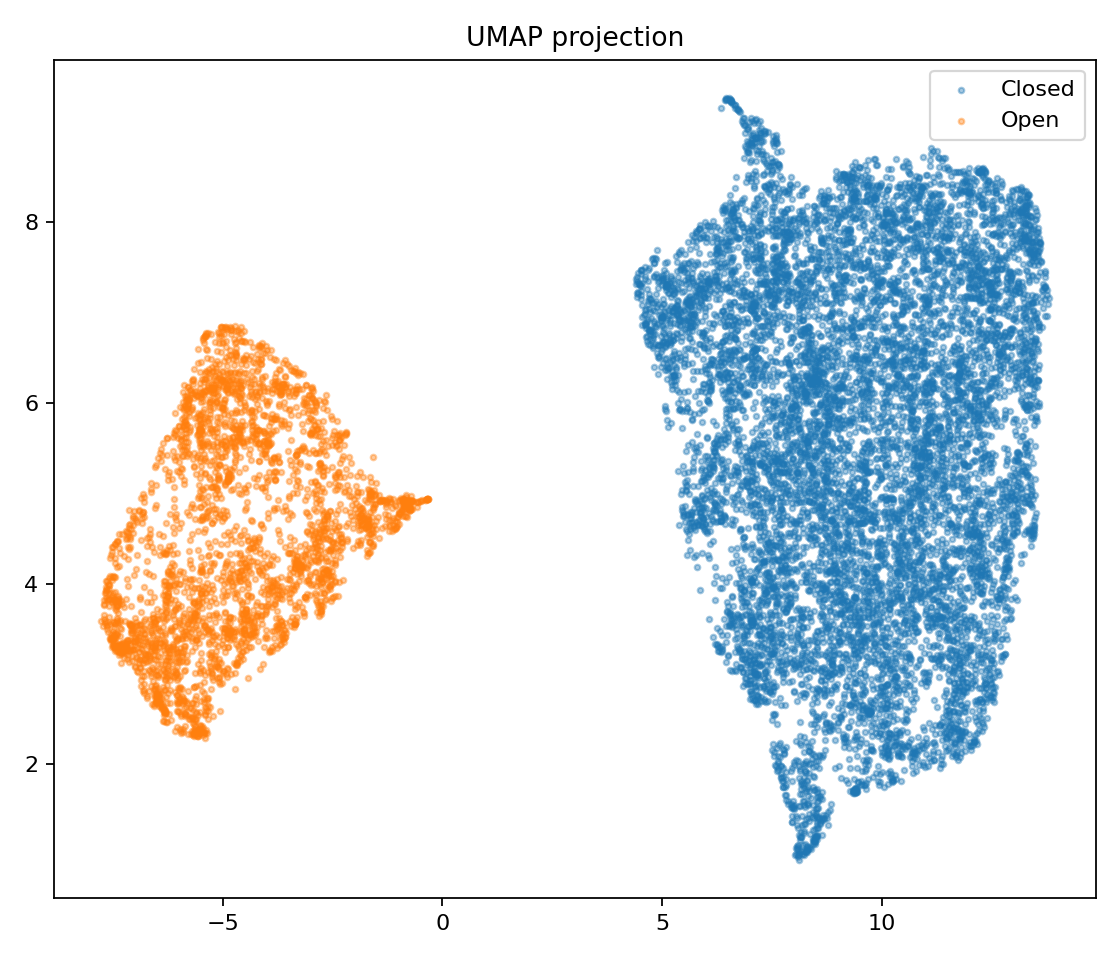} \\[2mm]

\parbox{0.30\textwidth}{\centering (a) Stable Diffusion 1.5 dataset} &
\parbox{0.30\textwidth}{\centering (b) Stable Diffusion 2.1 dataset} &
\parbox{0.30\textwidth}{\centering (c) Z-Turbo dataset}
\end{tabular}

\captionof{figure}{\color{black}Attention on Closed Only produces well-separated clusters across the Stable Diffusion 1.5, Stable Diffusion 2.1, and Z-Turbo datasets. AUC: 100.0\%, EER: 0.00, OVL: 0.00 for all datasets.}
\label{fig:kde_open}
\end{center}

\color{black}

\subsubsection{Comparison with Open-set Baselines}
To contextualize the Open-set behavior of Proto-LeakNet, we compare it with representative state-of-the-art methods under the WILD Open-set split. 
This comparison should be interpreted with care, because the considered approaches do not all follow the same Open-set formulation. 
Some methods are primarily designed as Closed-set attribution or forensic classification models and require an additional confidence-, calibration-, or distance-based mechanism to operate in an Open-set setting. 
Other approaches, such as OWDFA-CAL~\citep{zheng2026open}, LatentTracer~\citep{latenttracer2024} and OCC-CLIP~\citep{occlip2024}, define their own open-world or Open-set protocols, which may involve different assumptions, adaptation strategies, or calibration procedures.
In contrast, Proto-LeakNet follows a strictly post-training Open-set rejection protocol. 
The model is trained only on the WILD Closed-set generators. 
After training, the encoder is frozen and the density estimator is fitted exclusively on embeddings from the known Closed-set classes. 
No image from the WILD Open-set split (composed by 10 different generative engines - Section~\ref{sec:dataset}) is used during training, prototype learning, density estimation, threshold selection, or calibration. 
The Open-set samples are used only at test time to evaluate whether their embeddings fall outside the known-generator manifold.

Table~\ref{tab:open_sota} highlights the effectiveness of Proto-LeakNet in the Open-set rejection scenario. Since the compared approaches adopt partially different Open-set formulations, the comparison should be interpreted as a positioning with respect to the same known-versus-unknown separation task rather than as a fully homogeneous benchmark.

The proposed method achieves perfect separation, obtaining 100.00\% AUC together with 0.00 EER, 0.00 OVL, and 0.00 FPR@95. These results indicate that the latent representations learned by Proto-LeakNet form a highly structured embedding space, where known and unseen generators occupy clearly separable regions. This behaviour suggests that the proposed signal-leak-aware representation generalizes well beyond the generators observed during training.

Among the compared methods, the WILD baselines based on XceptionNet, EfficientNet\_B4, and ResNet50 achieve very strong Open-set performance, with AUC values above 99\%. However, these approaches still exhibit non-zero EER, OVL, and FPR@95 values, indicating that a residual overlap between known and unseen generators is still present. Although these image-domain models are able to capture discriminative generator-specific cues, they mainly rely on visual artifacts and spatial patterns extracted directly from the image content. As a consequence, their representations can still be influenced by semantic similarities, image content, and dataset-specific characteristics shared across different generators. Moreover, these architectures do not explicitly model the latent generative process responsible for producing diffusion traces, making their representations more sensitive to distribution shifts introduced by unseen generators. This results in a less robust separation between known and unknown samples compared to Proto-LeakNet.

OCC-CLIP~\citep{occlip2024} obtains the lowest performance among the compared methods in our Open-set rejection protocol, with an AUC of 81.92\%. OCC-CLIP addresses model origin attribution as a few-shot one-class classification problem, where a small set of images from a target generator is contrasted against non-target images sampled from open-domain datasets. Although OCC-CLIP is effective in its intended few-shot attribution setting, it does not explicitly model diffusion-specific signal-leak traces or their temporal evolution. This may limit its ability to separate known and unseen generators in our Open-set rejection protocol, especially when unseen generators share similar visual or semantic properties with the known ones. The relatively high OVL and FPR@95 further suggest that the learned representation produces a larger overlap between known and unknown samples than Proto-LeakNet.

OWDFA-CAL~\citep{zheng2026open} achieves good Open-set performance, obtaining 91.10\% AUC. The method improves Open-World Deepfake Attribution through confidence-aware pseudo-label learning and asymmetric calibration between known and novel classes. Nevertheless, the reported EER, OVL, and FPR@95 values show that the separation between known and unseen generators remains imperfect. The proposed confidence regularization strategy effectively reduces the bias toward known classes and improves the discovery of novel categories. However, the method mainly relies on confidence-aware consistency regularization and prototype-based learning. While effective for improving the attribution of novel categories, the reported Open-set metrics indicate that a residual overlap between known and unseen generators is still present, leading to a less discriminative separation compared with Proto-LeakNet.

LatentTracer~\citep{latenttracer2024} achieves strong Open-set performance, obtaining 98.97\% AUC with relatively low EER and OVL values. These results confirm that operating in the latent space provides more discriminative information for source attribution compared to purely image-domain approaches. However, the method still does not reach the performance achieved by Proto-LeakNet. In particular, the non-zero EER, OVL, and FPR@95 values indicate that a residual overlap between known and unseen generators is still present in the learned representation space. LatentTracer mainly relies on latent inversion and reconstruction analysis, without explicitly modeling how diffusion-related traces evolve across multiple timesteps of the generation process. As a result, the extracted representations may remain less robust when dealing with unseen generators that share similar latent characteristics with the known ones.

Overall, the comparison highlights how existing approaches still struggle to achieve a fully separable representation between known and unseen generators under Open-set conditions. Image-domain methods mainly rely on visual artifacts, while confidence-based approaches depend on feature-level clustering and pseudo-label regularization. Latent-space approaches provide stronger discriminative capabilities, but still exhibit residual overlap between known and unseen generators. In contrast, Proto-LeakNet benefits from the joint use of multi-timestep latent residuals, temporal attention, and prototype-based supervision, leading to a more compact and discriminative representation space.

\color{black}

\begin{table}[!htp]
\centering
\caption{\color{black}Open-set rejection comparison on the WILD Open-set split. 
Higher AUC is better, while lower EER, OVL, and FPR@95 indicate stronger Open-set rejection.}
  \label{tab:open_sota}
\begin{tabular}{ccccc}
\hline
\color{black}
\textbf{Method} & \color{black}\textbf{AUC (\%)} & \color{black}\textbf{EER}  & \color{black}\textbf{OVL} & \color{black}\textbf{FPR@95} \\ \hline
\color{black}WILD: EfficientNet\_B4   & \color{black}99.92    &\color{black} 0.13         & \color{black}0.27    &\color{black} 0.43      \\
\color{black}WILD: XceptionNet      &\color{black} \underline{99.99}    & \color{black}\underline{0.08}         & \color{black}\underline{0.17}    & \color{black}\underline{0.15}       \\
\color{black}WILD: ResNet50         & \color{black}99.93    & \color{black}0.15         & \color{black}0.29    &\color{black} 0.46       \\
\color{black}OWDFA-CAL~\citep{zheng2026open} & \color{black}91.10 &\color{black} 0.15 & \color{black}0.30 & \color{black}0.45 \\
\color{black}LatentTracer~\citep{latenttracer2024} & \color{black}89.31 & \color{black}0.16 &\color{black} 0.34 & \color{black}0.48 \\
\color{black}OCC-CLIP~\citep{occlip2024} & \color{black} 81.92 & \color{black} 0.25 & \color{black} 0.48 & \color{black} 0.55 \\
\color{black}\textbf{Ours}     & \color{black}\textbf{100.00}   & \color{black}\textbf{0.00} & \color{black}\textbf{0.00} &\color{black} \textbf{0.00}\\ 
\hline
\end{tabular}
\end{table}
\color{black}

\subsection{Temporal Interpretability of the Aggregation Module}
\label{sec:temp_analysis}
Temporal aggregation determines how Proto-LeakNet synthesizes information across the diffusion trajectory. Understanding which timesteps contribute most is essential for explaining how signal-leak cues evolve under increasing noise. The following analysis examines attention weights, contribution shares, and per-step energy distributions, highlighting a consistent and interpretable temporal structure. The main paper discusses the motivation for using multiple diffusion timesteps. Here we quantify how each latent contributes to the final embedding and show that the temporal module captures structured and complementary signal-leak information.

\subsubsection*{Attention weight distribution across timesteps}

Figure~\ref{fig:temporal-weights} shows the temporal attention weights $a_t$ for the three diffusion steps $\{0, 5, 10\}$, visualized for all test samples. The distribution is clearly non-uniform. The earliest latent $z_{0}$ receives the highest average weight, while $z_{5}$ and $z_{10}$ contribute with lower but stable weights across samples.

This pattern indicates that the model does not collapse to a single timestep and that all latent views are actively exploited. The global decrease of attention with increasing $t$ aligns with the expected behaviour of signal-leak cues, which are sharper at early noise levels and progressively shaped by stochastic refinement.

\begin{center}
  \includegraphics[width=0.5\linewidth]{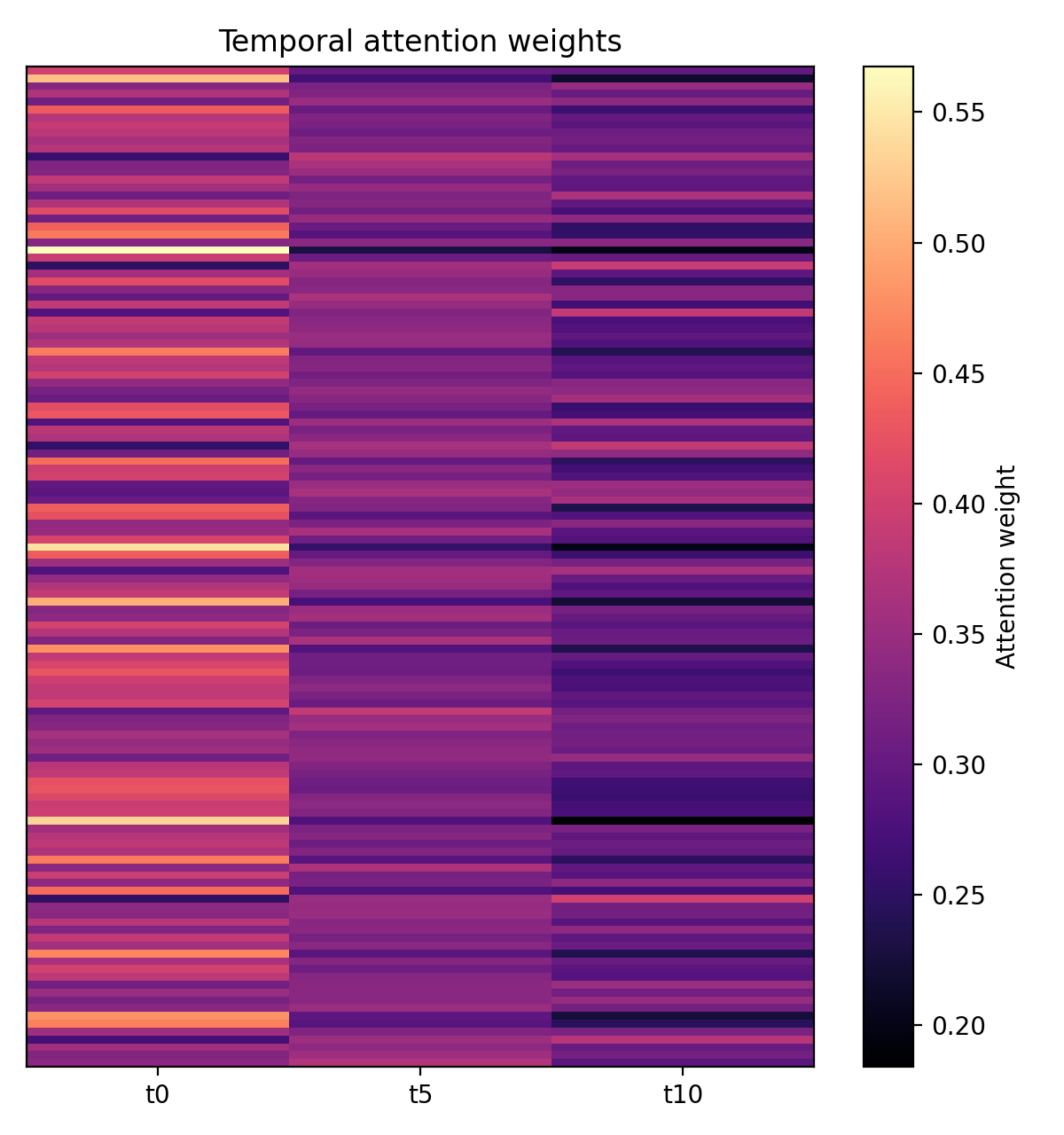}
  \captionof{figure}{Temporal attention weights across all test samples. Each row corresponds to an input image and each column to one of the diffusion timesteps. The attention distribution exhibits a structured pattern, with $t_0$ consistently receiving the highest weight, while later steps provide secondary yet stable contributions.}
  \label{fig:temporal-weights}
\end{center}

\subsubsection*{Attention weights vs.\ contribution shares}

To further analyze temporal behaviour, Figure~\ref{fig:temporal-contribution-bar} compares the mean attention weights with the effective contribution share induced by the attention-weighted projections. The two quantities closely track each other, demonstrating that the aggregation module behaves in a predictable and transparent manner.

Although $t_0$ contributes the most, both $t_5$ and $t_{10}$ retain substantial influence, with contribution shares between 25\% and 30\%. This confirms that the model leverages meaningful information from later steps rather than relying solely on the deterministic $t_0$ latent.

\begin{center}
  \includegraphics[width=.5\linewidth]{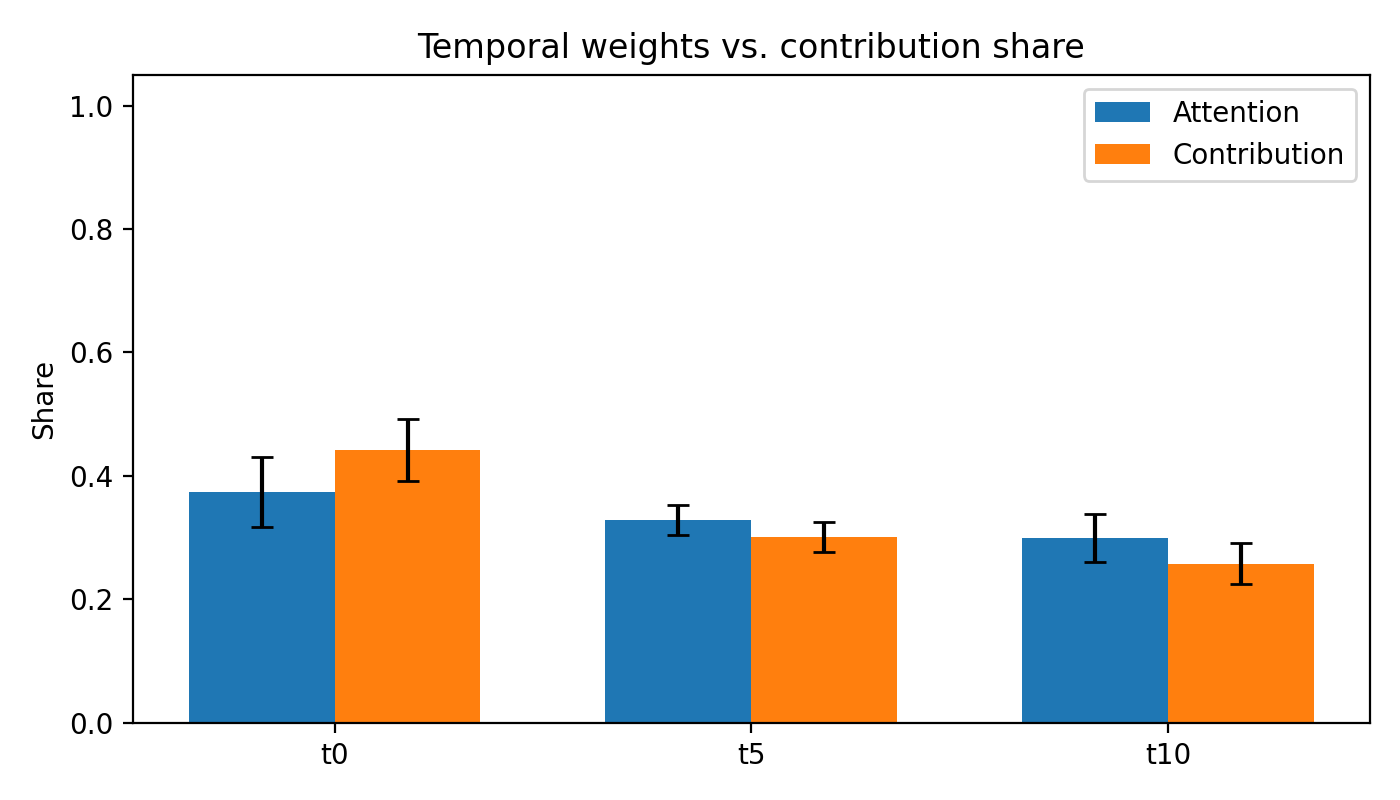}
  \captionof{figure}{Mean temporal attention weights versus normalized contribution shares for $t_0$, $t_5$, and $t_{10}$. Error bars denote standard deviations across samples. The close alignment between the two curves indicates that the aggregation module consistently transforms timestep importance into contribution strength.}
  \label{fig:temporal-contribution-bar}
\end{center}

\subsubsection*{Per-step contribution heatmap}

Figure~\ref{fig:temporal-energy} reports the normalized contribution share for each timestep across the entire test set. This visualization shows that the contribution pattern remains consistent at the dataset scale. The early timestep $t_0$ dominates, but $t_5$ and $t_{10}$ retain a stable contribution fraction, confirming that the temporal aggregation module does not degenerate into a single-step classifier.

\begin{center}
  \includegraphics[width=.7\linewidth]{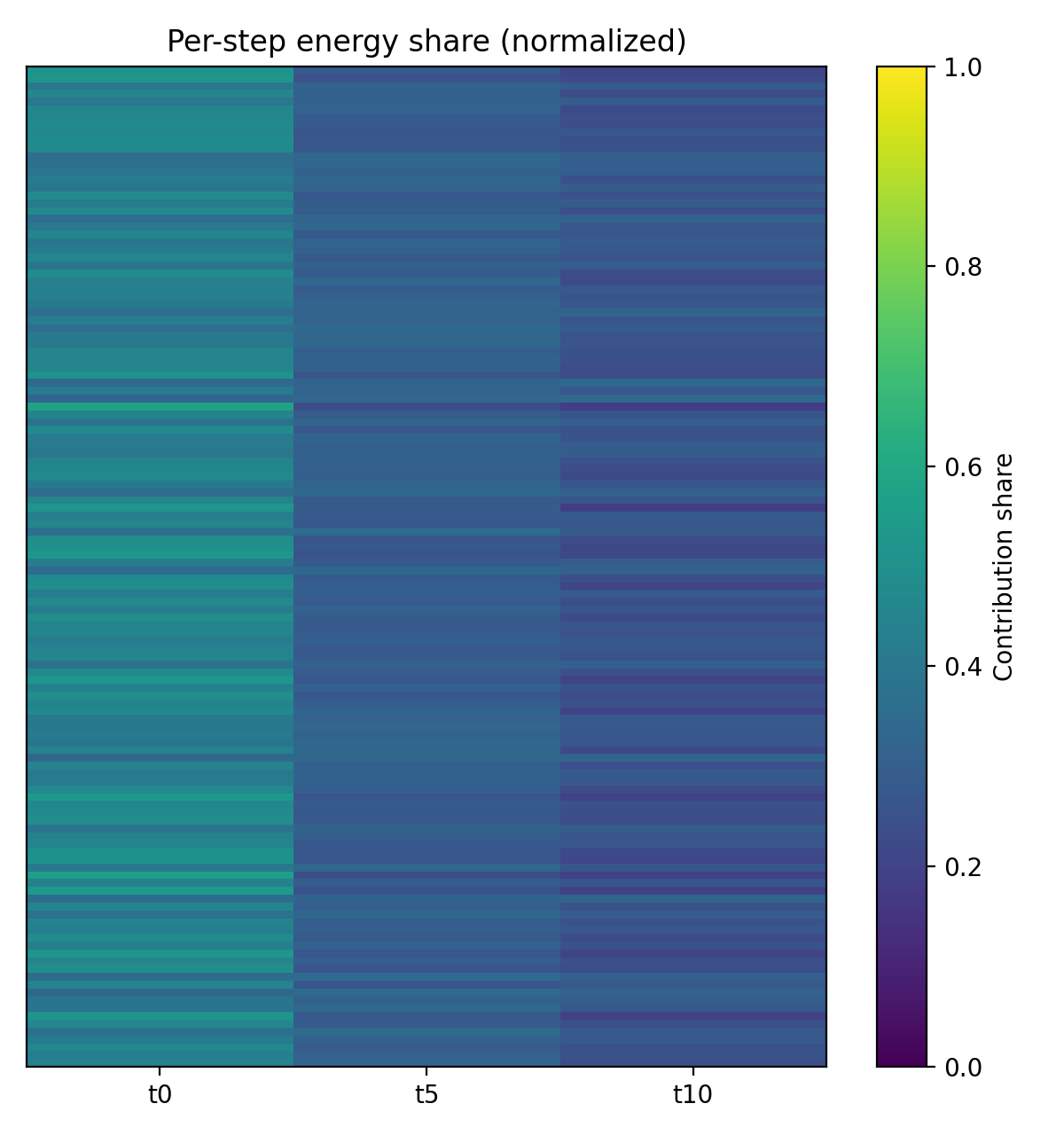}
  \captionof{figure}{Normalized contribution share for each diffusion timestep across all test samples. The structure mirrors the attention pattern: $t_0$ dominates, while $t_5$ and $t_{10}$ consistently provide secondary contributions. This confirms that the aggregation module leverages complementary information across multiple timesteps.}
  \label{fig:temporal-energy}
\end{center}

\subsubsection*{Interpretation of temporal behaviour}

The combined evidence from Figures~\ref{fig:temporal-weights}, \ref{fig:temporal-contribution-bar}, and \ref{fig:temporal-energy} provides a clear interpretability narrative:

\begin{itemize}
    \item The timestep $t_0$ contains the sharpest signal-leak residuals produced by the VAE encoder, which makes it the most discriminative source of evidence.
    \item The timestep $t_5$ introduces moderate stochastic variation that amplifies latent discrepancies not fully visible at $t_0$.
    \item The timestep $t_{10}$ retains weaker but still useful cues that help stabilize the representation and reduce variance across inputs.
\end{itemize}

The aggregation module therefore captures complementary information across the diffusion trajectory, rather than relying on a single latent view.

\subsubsection*{Why temporal aggregation cannot be ablated}

Temporal aggregation is structurally required by the architecture. Each image produces multiple latent vectors $\{z_{0}, z_{5}, z_{_{10}}\}$, but the classifier head predicts a single generator label. Without a temporal aggregation operator $A(\cdot)$, the model would produce three unrelated projections $Wz_{_0}$, $Wz_{_5}$, and $Wz_{_{10}}$ with no supervision signal to determine how they should be combined. This produces an under-determined learning objective that cannot be optimized.

The only meaningful ablation equivalent to removing aggregation is to use a single timestep. As shown in Table~\ref{tab:steps_ablation}, this variant yields consistently lower performance, indicating that additional timesteps provide useful complementary information. This confirms that temporal aggregation is not only structurally required but also empirically beneficial.

\begin{table}[!htp]
    \centering
    \caption{Ablation on the individual $t$ diffusion steps.  
    We report Top-1 Accuracy and Macro AUC on the Closed-set configuration.}
    \label{tab:steps_ablation}
    \begin{tabular}{ccc}
    \hline
    \textbf{Prototypes ($M$)} & \textbf{Top-1 Acc. (\%)} & \textbf{Macro AUC (\%)} \\ \hline
    $t=0$                     & 82.46                    & 97.83                   \\
    $t=5$                     & 82.40                   & 97.50                   \\
    $t=10$                     & 82.51                    & 97.55                   \\ \hline
    \textbf{$t=\{0,5,10\}$}            & \textbf{82.60}           & \textbf{98.13}          \\ \hline
    \end{tabular}
\end{table}

\subsection{Ablation Study}
To systematically assess the contribution of each architectural component in Proto-LeakNet, we conduct controlled ablation experiments targeting its core modules. 
All ablations are performed under identical training and evaluation conditions to isolate the effect of each design choice.
First, we analyze the role of prototype supervision and feature-wise attention in structuring the embedding space (Table~\ref{tab:ablation}). 
Removing prototype supervision significantly degrades both Top-1 Accuracy and Macro~AUC, indicating that explicit centroid attraction and inter-class margin enforcement are essential for maintaining geometric consistency across generator-specific clusters. 
Eliminating feature-wise attention, on the other hand, reduces the model’s ability to emphasize the most informative latent dimensions across diffusion steps, resulting in less compact embeddings and decreased separability. 
The combination of prototypes and attention consistently yields the strongest performance, confirming that structured supervision and selective feature aggregation jointly enhance discriminative capacity while preserving representation stability.
To evaluate whether signal-leak exploitation depends on a specific diffusion backbone, we replace Stable Diffusion~2.1 with Stable Diffusion~XL (SDXL) while keeping the remainder of the architecture unchanged. 
The resulting performance remains nearly identical (Top-1 Acc. 82.55\% and Macro~AUC 98.09\%), suggesting that the observed signal-leak behavior is not model-specific but rather intrinsic to diffusion-based generative processes.

\subsubsection{Encoder Architecture}
We next investigate the impact of the encoder backbone (Table~\ref{tab:backbone_ablation}). 
ResNet18 is compared against deeper convolutional architectures (ResNet50, ResNet101), EfficientNet-B4, and the transformer-based ViT-B16.
Increasing model capacity does not improve performance; in fact, larger backbones consistently reduce both Closed-set accuracy and Macro~AUC. 
This suggests that deeper or transformer-based models may overfit local latent variations, thereby weakening the global geometric structure required for generator-level attribution. 
In contrast, ResNet18 provides the best balance between compactness and discriminative power. 
Its moderate capacity appears sufficient to capture signal-leak patterns while maintaining stable generalization across perturbations and unseen distributions.

\subsubsection{Changing Prototype Number}
We further analyze the influence of the number of prototypes per class $M$ (Table~\ref{tab:prototypes_ablation}). 
Using only two prototypes restricts intra-class flexibility, leading to reduced cluster compactness and lower attribution accuracy. 
Conversely, increasing the number of prototypes to six introduces redundancy and partial overlap, slightly degrading separability. 
Setting $M=4$ achieves the best trade-off between intra-class expressiveness and inter-class separation, resulting in compact and well-structured clusters together with the highest Closed-set performance.
Temporal attention pooling, responsible for aggregating multi-step diffusion features, is not removed entirely in ablation, as doing so would eliminate temporal modeling by design. 
Instead, its influence is analyzed through the attention-disabled configuration reported in Table~\ref{tab:ablation}, which confirms that selective diffusion-step weighting materially contributes to stable and discriminative embeddings.
Overall, these ablation results validate the architectural design of Proto-LeakNet and demonstrate that each component contributes meaningfully to robust, structured latent-space attribution.

\begin{table}[!htp]
\centering
    \caption{Ablation study on Proto-LeakNet components. 
    We report Top-1 Accuracy and Macro AUC to quantify the effect of removing prototypes and attention mechanisms.}
    \label{tab:ablation}
    \begin{tabular}{ccc}
    \hline
    \textbf{Experiment}           & \textbf{Top-1 Acc (\%)} & \textbf{Macro AUC (\%)} \\ \hline
    No prototypes                 & 72.63                   & 93.09                   \\
    No attention                  & 81.23                   & 97.49                   \\
    No prototypes \& no attention & 79.80                   & 96.67                   \\ \hline
    \textbf{Full Proto-LeakNet}   & \textbf{82.60}          & \textbf{98.13}          \\ \hline
    \end{tabular}
\end{table}

\begin{table}[t!]
\centering
    \caption{Ablation on different backbone architectures.  
    We report Top-1 Accuracy and Macro AUC on the Closed-set configuration to evaluate the impact of the feature extractor on attribution performance.}
     \label{tab:backbone_ablation}
    \begin{tabular}{ccc}
    \hline
    \textbf{Backbone} & \textbf{Top-1 Acc. (\%)} & \textbf{Macro AUC (\%)} \\ \hline
    EfficientNet\_B4  & 65.80                    & 91.82                   \\
    ViT-B16           & 72.07                    & 94.53                   \\
    ResNet50          & 81.83                    & 97.39                   \\
    ResNet101         & 76.53                    & 95.36                   \\ \hline
    \textbf{ResNet18} & \textbf{82.60}           & \textbf{98.13}          \\ \hline
    \end{tabular}
\end{table}

\begin{table}[t!]
    \centering
    \caption{Ablation on the number of prototypes per class ($M$).  
    We report Top-1 Accuracy and Macro AUC on the Closed-set configuration.}
    \label{tab:prototypes_ablation}
    \begin{tabular}{ccc}
    \hline
    \textbf{Prototypes ($M$)} & \textbf{Top-1 Acc. (\%)} & \textbf{Macro AUC (\%)} \\ \hline
    $M=2$                     & 81.17                    & 97.41                   \\
    $M=6$                     & 81.33                    & 97.58                   \\ \hline
    \textbf{$M=4$}            & \textbf{82.60}           & \textbf{98.13}          \\ \hline
    \end{tabular}
\end{table}

\subsubsection{Open-set Density Modeling and Score Justification}
\label{sec:open_set_model}
Open-set recognition in Proto-LeakNet is performed by scoring how far a test embedding lies from the Closed-set manifold. This section compares three standard post-hoc OOD scoring functions: Mahalanobis distance, Energy score, and kernel density estimation (KDE) as defined in Eq. (15) in the main paper. We show that all three scores yield an almost identical geometric separation on our learned embedding, and we justify our choice of KDE as the primary score reported in the main paper.

\subsubsection*{Mahalanobis distance}

The Mahalanobis score assumes that the Closed-set embeddings follow a Gaussian distribution with mean $\mu$ and covariance matrix $\Sigma$. Given a test embedding $z \in \mathbb{R}^d$, the score is defined as
\begin{equation}
    s_{\mathrm{Mahalanobis}}(z) 
    = (z - \mu)^\top \Sigma^{-1} (z - \mu).
\end{equation}
Here, $\mu$ is the empirical mean of the training embeddings and $\Sigma$ is the estimated covariance matrix. This score measures the squared distance of $z$ from the center of the Closed-set distribution under a Gaussian metric. Its effectiveness depends on how well the class manifold can be approximated by a single Gaussian.

\subsubsection*{Energy score}

The Energy score is a log-partition-based statistic derived from classifier logits~\cite{liu2020energy}.  
Given a vector of class logits $f(x) \in \mathbb{R}^C$ for a sample $x$, the score is
\begin{equation}
    s_{\mathrm{Energy}}(x)
    = - \log \left( \sum_{c=1}^{C} e^{f_c(x)} \right).
\end{equation}
Here, $f_c(x)$ is the pre-softmax activation for class $c$ produced by the prototype head.  
High-energy (less negative) values typically correspond to inputs lying away from the training distribution.  
Unlike Mahalanobis distance, this score depends on classifier outputs rather than geometric density.

\subsubsection*{Geometric comparison of scores}

Figures~\ref{fig:umap-energy}, \ref{fig:umap-kde} and \ref{fig:umap-maha} visualize the UMAP projection of the learned embedding colored by each ranking score. All three scores produce nearly identical separations between closed and open generators. This confirms that Open-set performance is driven primarily by the \emph{geometry} of the embedding produced by Proto-LeakNet, rather than by the specific choice of scoring function.

\begin{center}
  \includegraphics[width=.5\linewidth]{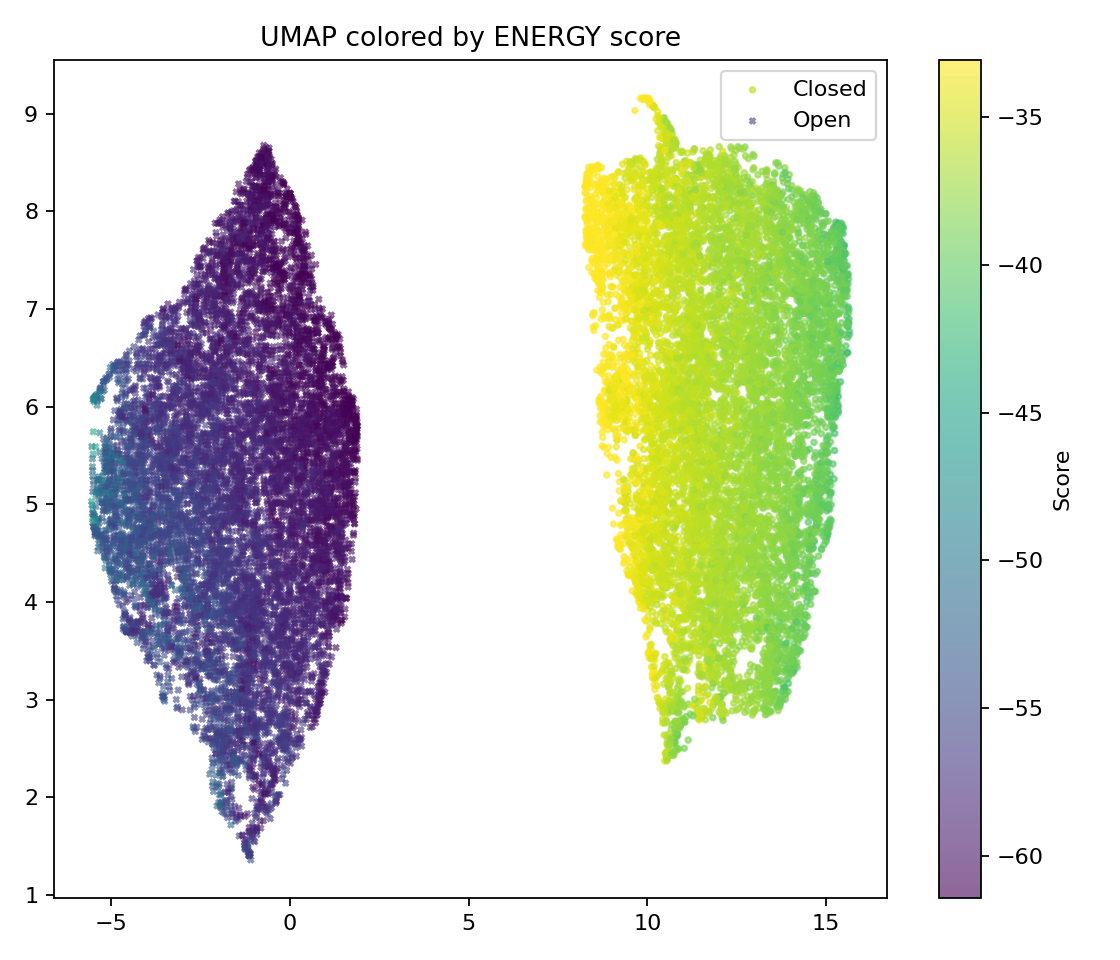}
  \captionof{figure}{UMAP projection colored by the Energy score. Closed-set and Open-set samples form two well-separated structures.}
  \label{fig:umap-energy}
\end{center}

\begin{center}
  \includegraphics[width=.5\linewidth]{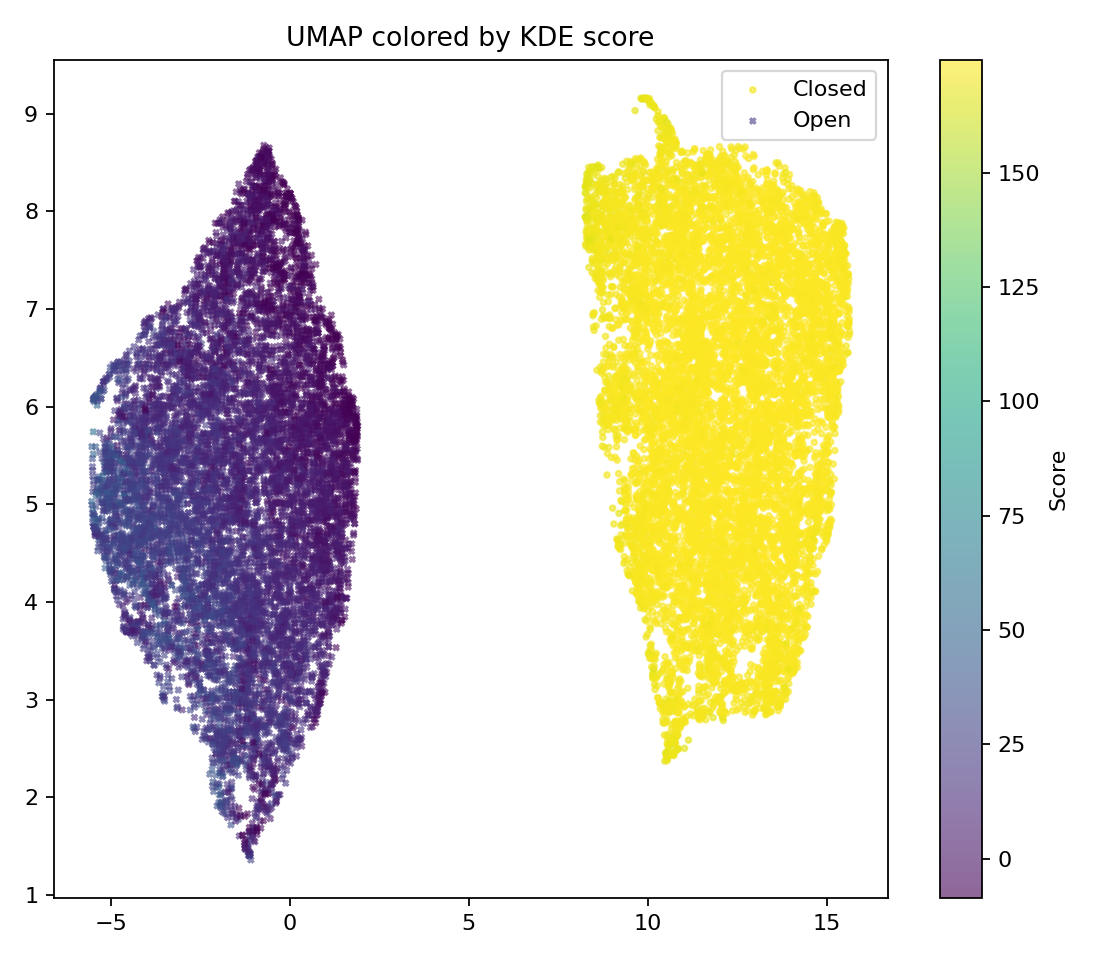}
  \captionof{figure}{UMAP projection colored by the KDE score. The separation closely mirrors that observed with the Energy score.}
  \label{fig:umap-kde}
\end{center}

\begin{center}
  \includegraphics[width=.5\linewidth]{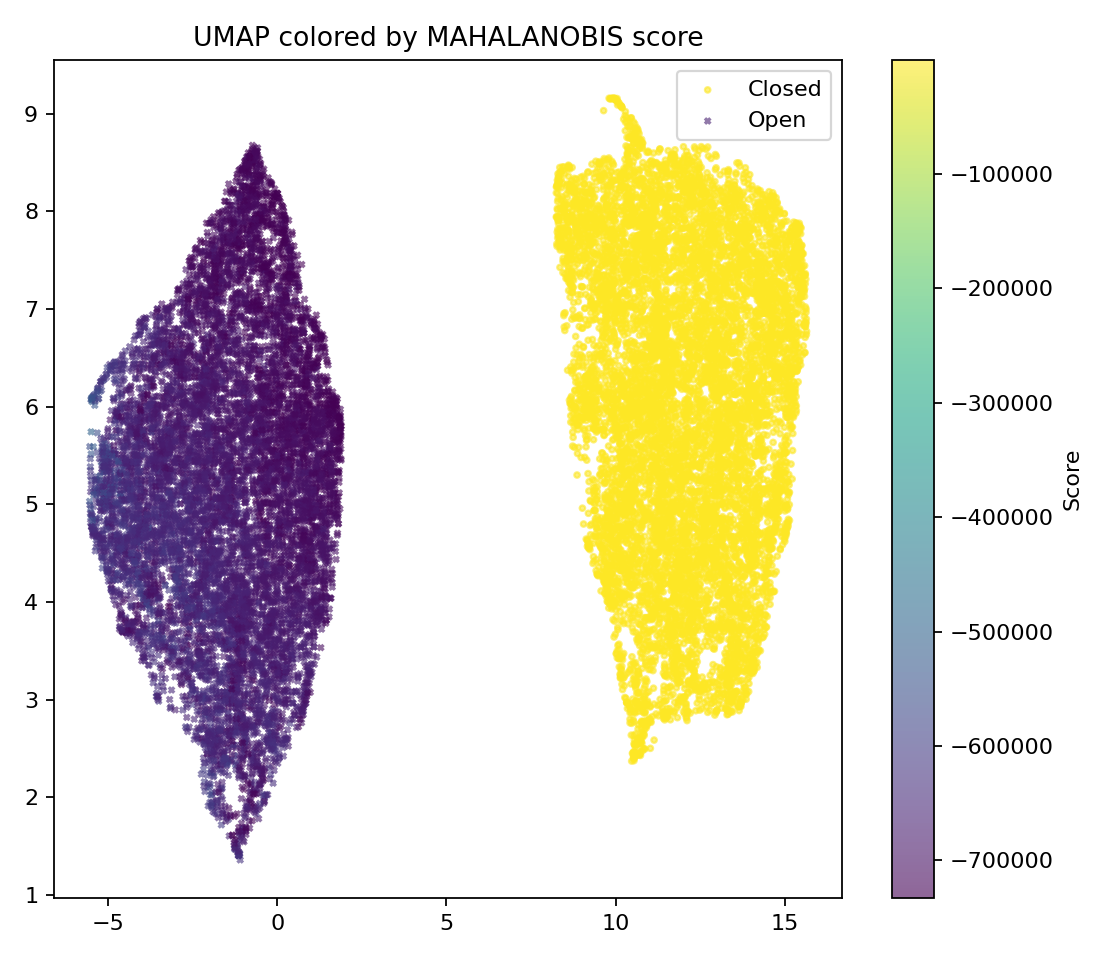}
  \captionof{figure}{UMAP projection colored by the Mahalanobis distance. The geometric separation between closed and open domains is consistently preserved.}
  \label{fig:umap-maha}
\end{center}

\subsubsection*{Why KDE is preferred}
Although all scores produce nearly identical separation on our embedding, KDE offers several conceptual and practical advantages that make it a more appropriate choice for attribution-oriented Open-set analysis. KDE naturally captures the multi-modal structure of the Closed-set latent space, which is crucial in attribution settings where different generators induce distinct modes; by contrast, Mahalanobis distance relies on a single Gaussian approximation and cannot express multi-cluster manifolds. KDE additionally provides an explicit probability density estimate $p(z)$, enabling overlap-based metrics such as the OVL used in the main paper, whereas neither Mahalanobis distance nor the Energy score yields a normalized density. The Energy score also depends on classifier logits and therefore couples Open-set behaviour with calibration of the prototype head, while KDE operates directly on the embedding geometry and remains stable across classifier architectures. Moreover, Mahalanobis distance requires estimating and inverting a global covariance matrix that may be ill-conditioned in high-dimensional anisotropic spaces, a limitation that KDE avoids through local kernel smoothing. KDE is also more robust under distributional shifts or the introduction of additional generators, as it does not rely on re-estimating global parametric structures. For these reasons, despite the similar empirical ranking produced by all three scores, KDE provides a density-based, multi-modal, and classifier-agnostic formulation that aligns naturally with the structure of Proto-LeakNet embeddings and with the Open-set metrics employed in our evaluation.

\section{Evaluation on Partial Manipulation Dataset}
\label{sec:eval_part_dataset}

To further evaluate the generalization ability of Proto-LeakNet under a more challenging and realistic manipulation scenario, we conduct additional experiments on the Partial Manipulation Dataset described in Section~\ref{sec:dataset}. 
Unlike the previous benchmarks used throughout the paper, which consist of fully generated images, this dataset introduces a fundamentally different attribution setting: real images are lightly modified by generative editing models, leaving large portions of the original content intact. 
Consequently, the generator-specific traces become weaker and more localized, making attribution significantly more difficult.

The dataset contains five classes: four corresponding to images edited by different models (FLUX.1-Kontext, Qwen2-VL, GLM-Image, and Instruct-Pix2Pix) and one class containing the original unmodified FFHQ images. 
Each class contains 5,000 samples, resulting in a balanced evaluation set of 25,000 images. 
In this configuration, the task requires the model to discriminate not only between different manipulation pipelines but also between manipulated and fully real images. 
This differs from standard generator attribution benchmarks where every sample is synthetic and the generative signal is typically present across the entire image.

Results for this experiment are reported in Table~\ref{tab:all_aucs_partial_man}, where Proto-LeakNet achieves a Macro AUC of \textbf{80.21\%}. 
Considering the difficulty of the task, this result indicates that the signal-leak cues captured by Proto-LeakNet remain partially detectable even when the generative model affects only a localized region of the image. 
The performance gap compared to the fully synthetic benchmarks evaluated earlier in the paper is expected, as the edited images still retain substantial real-image content that can obscure or dilute the generator-specific latent traces.

To better visualize the structure of the learned embedding space, Fig.~\ref{fig:step1_closed_man} shows a two-dimensionalprojection of the latent representations extracted by Proto-LeakNet. 
Despite the partial nature of the manipulations, a noticeable degree of clustering emerges among several classes. 
In particular, the embeddings corresponding to certain editing models form partially distinct regions of the latent space, indicating that the model is able to capture manipulation-specific patterns even when the generative process affects only part of the image. 
At the same time, some overlap between classes is visible, which is consistent with the nature of the dataset: because many manipulations are subtle and localized, the resulting signal-leak traces are inherently weaker and less spatially pervasive than those produced by fully generative pipelines.

The visualization therefore reflects the expected difficulty of the task. 
Rather than producing perfectly separated clusters as observed in the Closed-set generator attribution experiments, the embeddings exhibit a mixture of compact regions and partially overlapping areas. 
This behavior suggests that Proto-LeakNet still extracts meaningful manipulation-dependent cues, but that the underlying signal becomes less dominant when only a limited portion of the image is synthesized.

Although the dataset also provides a second manipulation stage (Step~2), where images undergo an additional sequential edit, our evaluation focuses on the Step~1 configuration. 
This choice is motivated by the goal of isolating the attribution signal introduced by the first editing operation. 
When multiple editing stages are applied sequentially, the resulting image may contain traces from several generative processes simultaneously, potentially producing mixed or partially overwritten signals that make attribution ambiguous even at the semantic level. 
Evaluating Step~1 therefore provides a clearer and more controlled benchmark for assessing whether Proto-LeakNet can detect generator-specific traces in partially manipulated images.

Overall, these results demonstrate that Proto-LeakNet retains a measurable degree of attribution capability even when the generative signal is sparse and localized. 
This experiment therefore complements the previous evaluations by showing that signal-leak cues are not limited to fully synthetic images but can also emerge in partially edited real photographs, albeit with reduced separability due to the weaker presence of generative artifacts.

\begin{table}[t!]
\centering
\caption{AUC (\%) results for each model class with the Partial Manipulation Dataset on Closed-set setting. The table reports both per-class AUC and overall Macro AUC.}
\label{tab:all_aucs_partial_man}
\begin{adjustbox}{width=.8\linewidth}
\begin{tabular}{clcccccc|
>{\columncolor[HTML]{EFEFEF}}c }
\cline{3-7}
\textbf{}                         &                        & \multicolumn{6}{c|}{\textbf{Closed-set Classes}}                                                                                                                                                                                                                                                     \\ \cline{1-7}
\textbf{Type} & \textbf{Methodologies} & \textbf{FFHQ} & \textbf{FLUX} & \textbf{GLM} & \textbf{Pix2Pix} & \textbf{QWEN} & \multirow{-2}{*}{\cellcolor[HTML]{EFEFEF}\textbf{Macro AUC}} \\ \hline
\multirow{-1}{*}{\textbf{RAW}}   & \textbf{Proto-LeakNet}          & 78.60  & 71.33  & 88.15  & 84.37  & 78.57 & 80.21  \\ \hline
\end{tabular}
\end{adjustbox}
\end{table}

\begin{center}
  \includegraphics[width=.5\columnwidth]{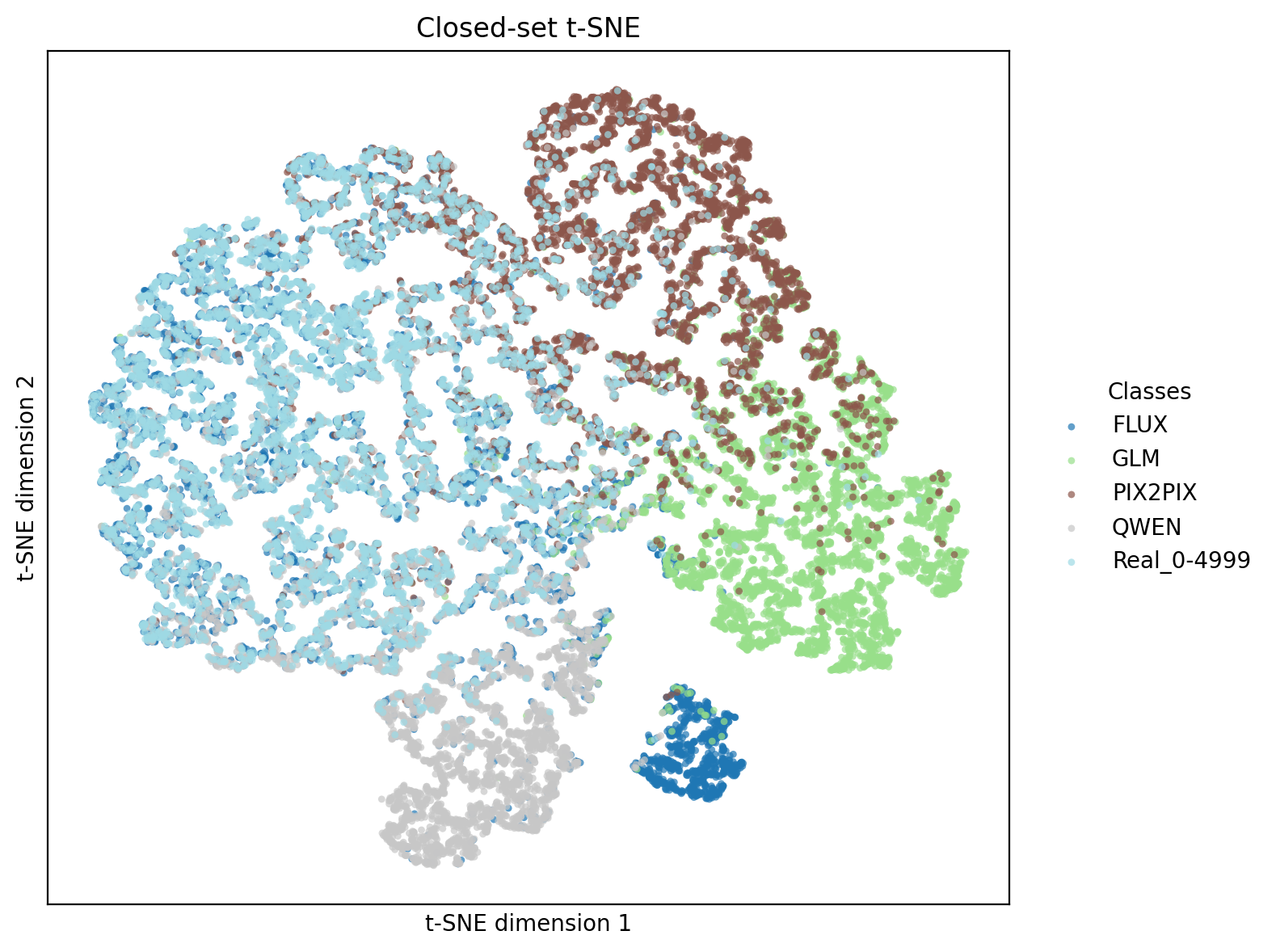}
  \captionof{figure}{Plot portraying how class representative-clusters are organized in the latent space denoting an hint to separating with moderate misclassifications on the Step 1 of the Partial Manipulation Dataset.}
  \label{fig:step1_closed_man}
\end{center}

\section{Discussion and Limitations}
\label{sec:discussion}

The experimental results demonstrate that signal-leak bias can be leveraged as a stable and discriminative cue for diffusion-based source attribution. 
Across Closed-set, degraded, and Open-set scenarios, Proto-LeakNet consistently learns structured latent representations that preserve generator-specific separation while maintaining robustness to perturbations. 
These findings suggest that diffusion models encode persistent structural signatures in their latent trajectories, which can be systematically modeled for forensic attribution. 
In the following, we position our approach within the broader attribution literature, discuss its methodological implications, and outline current limitations and directions for future research.

\subsection{Positioning within the Attribution Literature}
This work situates itself at the intersection of deepfake source attribution and diffusion-model analysis. 
While prior attribution methods primarily operate in pixel space~\citep{freqnet2023, susy2024, npr2024} or rely on global embeddings extracted from pretrained models~\citep{occlip2024}, Proto-LeakNet leverages structured latent residuals derived from the diffusion process itself. 
In contrast to reverse-engineering approaches such as LatentTracer~\citep{latenttracer2024}, which attempt to invert images into candidate generators, our method directly models the geometry of diffusion latents and their temporal evolution.
More broadly, this study extends recent observations on signal-leak bias in diffusion models by demonstrating that these residual traces are not merely generation artifacts but stable, discriminative cues that can be structured into a coherent attribution manifold. 
To our knowledge, this is the first framework that explicitly integrates multi-step diffusion residuals with prototype supervision and temporal attention to perform generator-level attribution.

\subsection{Implications of Signal-Leak Modeling}
The empirical results suggest that signal-leak patterns persist across diffusion steps and remain detectable even under substantial post-processing. 
This observation supports the hypothesis that diffusion-based generators encode structural biases in their latent trajectories, rather than relying solely on superficial pixel-level artifacts.
Furthermore, the representation-level analysis indicates that the learned embedding space maintains clear geometric margins between known generators, unseen generators, and real imagery. 
Separability is not tied to a specific density estimation method, as consistent trends are observed across KDE, Mahalanobis distance, and Energy-based scoring. 
This reinforces the interpretation that Open-set discrimination arises from the learned representation itself rather than from a particular scoring mechanism.

\subsection{Generalization and Robustness}
Across all experimental settings, Proto-LeakNet maintains stable attribution performance under progressive degradation, including compression and compound perturbations. 
Compared to pixel-domain baselines, latent-domain modeling appears less sensitive to high-frequency distortions, suggesting that signal-leak cues reside in deeper structural representations.
The extended Closed-set experiment including authentic images further demonstrates that the embedding space preserves separation between synthetic generators and real content without collapsing into a binary detector. 
This supports the claim that Proto-LeakNet captures generator-specific structure rather than relying on generic real-vs-fake cues.

\subsection{Limitations}
Despite its robustness, Proto-LeakNet presents several limitations.
First, the primary evaluation focuses on face-centric datasets. 
Although additional experiments with ImageNet and FFHQ confirm representation-level separability for real imagery, further validation across object-centric, scene-centric, and multimodal datasets is required to establish broader generalization.
Second, the framework relies on latent representations derived from Stable Diffusion’s VAE. 
While experiments with SDXL indicate similar behavior, it remains an open question whether equivalent signal-leak separability emerges in fundamentally different generative paradigms, such as autoregressive transformers or non-diffusion-based architectures.
Third, our feature construction step requires mapping images into a diffusion VAE latent space in order to apply a controlled forward noising procedure that exposes signal-leak residuals. 
Notably, this does not assume access to the unknown generator, nor does it rely on latent inversion or reverse diffusion. 
However, it introduces a dependence on the latent encoder used for projection (e.g., the Stable Diffusion VAE) and on the stability of this mapping under domain shifts and heavy transformations. 
Exploring alternative latent encoders and quantifying sensitivity to VAE-domain mismatch constitute important directions for future work.

\section{Conclusions and Future Works}
\label{sec:conclusions}
In this work, we introduced Proto-LeakNet, a signal-leak-aware framework for generative source attribution that operates directly in diffusion latent space. 
Rather than relying on pixel-level artifacts or inversion-based heuristics, our approach models structured residual cues emerging from the diffusion process itself. 
By combining multi-step latent aggregation, temporal attention, and prototype-based supervision, Proto-LeakNet learns a geometrically organized embedding space that supports both discriminative attribution and representation-level generalization.
Across Closed-set, degraded, and Open-set scenarios, the experimental results demonstrate that signal-leak bias constitutes a stable forensic signal. 
The model preserves attribution accuracy under strong post-processing and maintains clear separability between known generators, unseen generators, and authentic imagery. 
Importantly, separability is not tied to a specific density estimation method, suggesting that the observed margins arise from intrinsic properties of the learned representation rather than from a particular scoring mechanism.
Beyond classification performance, Proto-LeakNet contributes a structured interpretation of generator-specific latent behavior. 
Prototype supervision enforces cluster geometry at the class level, while temporal attention highlights informative diffusion steps, providing process-level transparency. 
Although these explanations remain structural rather than semantic, they represent a step toward interpretable generative attribution grounded in latent statistics.
Future work will investigate whether signal-leak patterns extend beyond face-centric domains to object- and scene-centric imagery, as well as multimodal and cross-domain settings. 
Exploring alternative generative paradigms, including autoregressive and GAN-based architectures such as StyleGAN, will further clarify whether signal-leak traces represent a diffusion-specific phenomenon or a broader property of deep generative models. 
Additionally, theoretical characterization of signal-leak dynamics may help bridge empirical observation and formal analysis. 
\color{black}
Furthermore, we plan to extend the proposed framework to the video domain, with particular focus on fully synthetic AI-generated videos produced by recent text-to-video and image-to-video models. This scenario is becoming increasingly challenging for authenticity analysis, as modern generative systems are able to produce highly realistic and temporally coherent content that is often difficult to distinguish from real footage. In this direction, future evaluations will include recent benchmarks specifically designed for synthetic video forensics, such as the SynthForensics dataset~\citep{leotta2026synthforensics}, which introduces a large-scale benchmark of fully synthetic videos generated by multiple state-of-the-art video generation models.
\color{black}
By framing signal-leak bias as a structured and measurable fingerprint, this work lays the groundwork for robust and interpretable attribution of generative media, contributing toward more reliable forensic analysis in the evolving landscape of synthetic content.

\clearpage


\end{document}